\documentclass[pdflatex,sn-mathphys-num]{sn-jnl}

\usepackage{graphicx}%
\usepackage{multirow}%
\usepackage{amsmath,amssymb,amsfonts}%
\usepackage{amsthm}%
\usepackage{mathrsfs}%
\usepackage[title]{appendix}%
\usepackage{xcolor}%
\usepackage{textcomp}%
\usepackage{manyfoot}%
\usepackage{booktabs}%
\usepackage{algorithm}%
\usepackage{algorithmicx}%
\usepackage{algpseudocode}%
\usepackage{listings}%
\usepackage{tikz}
\usetikzlibrary{positioning}
\usetikzlibrary{arrows.meta}
\usetikzlibrary{shapes.geometric}

\theoremstyle{thmstyleone}%
\theoremstyle{thmstyletwo}%

\theoremstyle{thmstylethree}%

\begin{document}

\title[Article Title]{Evaluation of Convolutional Neural Network For Image Classification with Agricultural and Urban Datasets}


\author*[1]{\fnm{Shamik} \sur{Shafkat Avro}}\email{shamikshafkat-2019117793@cs.du.ac.bd}
\author*[1]{\fnm{Nazira} \sur{Jesmin Lina}}\email{nazirajesmin-2019117838@cs.du.ac.bd}
\author*[1]{\fnm{Shahanaz} \sur{Sharmin}}\email{shahanaz-2019417790@cs.du.ac.bd}
\equalcont{These authors contributed equally to this work.}

\affil*[1]{\orgdiv{Department of Computer Science and Engineering}, \orgname{University of Dhaka}}


\abstract
{This paper presents the development and evaluation of a custom Convolutional Neural Network (CustomCNN) created to study how architectural design choices affect multi-domain image classification tasks. The network uses residual connections, Squeeze-and-Excitation attention mechanisms, progressive channel scaling, and Kaiming initialization to improve its ability to represent data and speed up training. The model is trained and tested on five publicly available datasets: unauthorized vehicle detection, footpath encroachment detection, polygon-annotated road damage and manhole detection, MangoImageBD and PaddyVarietyBD. A comparison with popular CNN architectures shows that the CustomCNN delivers competitive performance while remaining efficient in computation. The results underscore the importance of thoughtful architectural design for real-world Smart City and agricultural imaging applications.}

\keywords{Convolutional Neural Network (CNN), Transfer Learning, Deep Learning, Image Classification, ResNet50, VGG16}



\maketitle

\section{Introduction}\label{sec1}

Convolutional Neural Networks (CNNs)\cite{deshpande2016beginner} have become essential in modern computer vision. They show impressive results in tasks like image classification, object detection, and segmentation. Creating a custom CNN allows us to investigate how different parts of the architecture, like convolutional depth, filter sizes, activation functions, normalization layers, and pooling methods, affect model performance and generalization. The CustomCNN for this paper is a modern hybrid design that combines the benefits of residual learning, Squeeze-and-Excitation (SE)\cite{hu2018squeeze} attention, and deep feature hierarchies. This model is built from scratch using PyTorch. Every part of the architecture is based on established principles of CNN design.

\vspace{5pt}
The network starts with a lightweight 3×3 convolutional stem. It then has four deeper stages built with ResidualSEBlocks. Each block includes two 3×3 convolutions with batch normalization and ReLU activation \cite{he2018relu}. There is also a residual skip pathway to help with gradient flow, and a Squeeze-and-Excitation module that applies channel-wise attention to highlight important feature maps. As the network progresses, the spatial resolution decreases through carefully planned strided convolutions, while the channel width follows a 1×, 2×, 4× pattern to boost representational capacity.
\vspace{5pt}
To improve generalization, the architecture includes Dropout2d in each block, along with a 0.5 dropout layer in the fully connected head. A global average pooling layer takes the place of traditional large fully connected layers. This change reduces the number of parameters and improves robustness across various datasets. The final classifier is a compact two-layer MLP \cite{griniasty1992two}. It maps the learned 4C-dimensional feature vector through a 128-unit hidden layer to the number of target classes.For stable and efficient training, the model uses Kaiming (He) initialization for all convolutional and linear layers. It is optimized with an Adam optimizer that employs selective weight decay to avoid penalizing batch normalization and bias parameters.
\vspace{5pt}
In summary, the CustomCNN shows a thoughtful balance of depth, width, attention, and regularization. It is designed to work well across the five datasets used in this work while remaining computationally efficient and clear in its design choices.

\section{Background}
\subsection{Overview of CNN}
Convolutional Neural Networks (CNNs) are a type of deep learning model mainly used for image classification and computer vision tasks. CNNs are designed to automatically learn important features from images, such as edges, shapes, and textures. Unlike traditional machine learning methods, CNNs do not require manual feature extraction.

\vspace{5pt}
A typical CNN consists of convolutional layers, pooling layers, and fully connected layers. Convolutional layers extract features from the input image, pooling layers reduce the spatial size of the feature maps, and fully connected layers perform the final classification. CNNs are widely used because they can achieve high accuracy while handling large and complex image datasets.
\subsection{ResNet}
ResNet \cite{wang2020rsnet}, short for Residual Network, is a deep CNN architecture introduced to solve the problem of training very deep neural networks. As networks become deeper, they can suffer from vanishing gradients, which makes training difficult and reduces performance.ResNet addresses this problem by using residual connections, also called skip connections. These connections allow the input of a layer to be added directly to its output, making it easier for the network to learn. Because of this design, ResNet models can be very deep while still training effectively. ResNet has shown strong performance in many image classification tasks and is commonly used as a baseline in CNN comparisons.
\subsection{VGG16}
VGG16 \cite{kaur2019automated} is a deep CNN architecture known for its simple and uniform design. It consists of 16 layers with learnable parameters, mainly using small 3×3 convolution filters stacked on top of each other. This consistent structure makes VGG16 easy to understand and implement. Although VGG16 has a large number of parameters and requires more computational resources, it has been very successful in image classification tasks. Due to its straightforward architecture, VGG16 is often used as a reference model when comparing CNN performance. However, its large size can lead to longer training times and higher memory usage.

\subsection{Transfer Learning in CNN}
Transfer learning \cite{torrey2010transfer, pan2020transfer} is a technique where a CNN trained on a large dataset is reused for a new but related task. In image classification, models are often pretrained on large datasets such as ImageNet. These pretrained models have already learned general image features like edges and textures. In transfer learning, the pretrained layers are either frozen or fine-tuned, while new layers are added for the specific task. This approach helps improve performance, especially when the new dataset is small. Transfer learning also reduces training time and helps prevent overfitting. Because of these advantages, transfer learning is widely used in practical CNN applications.

\section{Methodology}
\subsection{Custom CNN Architecture}

In this work, we propose a novel convolutional neural network (CNN) architecture that integrates residual learning, channel-wise attention, and scalable width, while maintaining a compact parameter budget suitable for training across multiple datasets. The architecture is constructed modularly using four stages of Residual Squeeze-and-Excitation (SE) blocks, preceded by a lightweight convolutional stem and followed by a fully connected classification head. 

\subsubsection{Overall Architecture Design}

The proposed model follows a hierarchical feature extraction pipeline consisting of the following sequential components:

\begin{itemize}
    \item \textbf{Stem:} A single $3 \times 3$ convolution with batch normalization and ReLU activation maps the input RGB image into a base feature space of $C$ channels.
    \item \textbf{Stage 1:} A ResidualSEBlock that maintains spatial resolution while refining the base channel features.
    \item \textbf{Stage 2:} A ResidualSEBlock with stride $2$ to downsample spatial dimensions and double the number of channels to $2C$.
    \item \textbf{Stage 3:} A second downsampling ResidualSEBlock that further expands feature dimensionality to $4C$.
    \item \textbf{Stage 4:} A final downsampling block operating at the same channel dimensionality ($4C$), enabling deeper feature abstraction.
    \item \textbf{Global Average Pooling:} The resulting tensor is reduced to a $4C$-dimensional vector.
    \item \textbf{Fully Connected Head:} A two-layer multilayer perceptron (MLP) \cite{griniasty1992two} projects the pooled features into a 128-dimensional latent space, followed by a final linear layer producing logits for the target classes.
\end{itemize}

All convolutional layers use He (Kaiming) initialization and are trained using an Adam optimizer with carefully structured weight decay to prevent over-regularization of batch normalization and bias parameters.

\subsection{Design Rationale and Architectural Choices}

The proposed CustomCNN architecture integrates residual learning, channel-wise attention, progressive feature scaling, and efficient downsampling \cite{lin2006adaptive} strategies to provide a compact yet expressive model suitable for diverse image classification datasets. This section outlines the rationale behind each architectural component and design decision.

\subsubsection{Residual Connections for Stable Deep Learning}

Residual shortcuts are incorporated to improve gradient propagation and mitigate both vanishing gradients and optimization degradation. By enabling each block to learn a residual function $F(x)$ added to an identity mapping, the network allows deeper stacks of layers without sacrificing convergence stability. This principle, inspired by ResNet~\cite{wang2020rsnet}, ensures that increasing network depth enhances representational capacity rather than hindering it.

\subsubsection{Squeeze-and-Excitation Attention Mechanism}

Each block integrates a Squeeze-and-Excitation (SE) module to adaptively recalibrate channel-wise feature activations. The SE block computes global descriptors via average pooling, followed by a bottleneck MLP with reduction ratio $r = 16$. This mechanism enables the network to emphasize semantically informative channels while suppressing weak or noisy features. SE attention is lightweight and parameter-efficient, making it especially valuable when training on datasets of different scales and characteristics.

\subsubsection{Choice of Convolutional Kernel Size}

The architecture employs $3\times 3$ convolutions throughout, consistent with established designs such as VGG and ResNet. This kernel size provides a strong balance between expressive receptive fields and computational cost. Stacking multiple $3\times 3$ layers increases nonlinear representational capacity while preserving spatial locality, allowing the network to model fine-grained textures as well as broader spatial patterns.

\subsubsection{Progressive Channel Expansion}

Feature dimensionality scales progressively across stages, following the sequence:
\[
C \rightarrow 2C \rightarrow 4C \rightarrow 4C.
\]
This progression allocates more channels to deeper layers, which encode increasingly abstract patterns and benefit from higher representational capacity. Maintaining constant width in the final stage avoids unnecessary parameter increase while still allowing deeper processing at rich feature scales.

\subsubsection{Padding Strategy}

All $3 \times 3$ convolutions use a padding value of $1$ to preserve spatial resolution within each block. This ensures compatibility with residual additions, which require input and output tensors of identical dimensions. The padding choice follows standard CNN design principles and maintains spatial consistency until intentional downsampling via strided convolution.

\subsubsection{Downsampling with Strided Convolutions}

Instead of max pooling, downsampling is performed using strided convolutions (stride $2$ in Stages 2--4). This allows the model to learn optimal downsampling kernels and reduces reliance on hand-crafted pooling operations. Strided convolutions preserve more spatial context and enable smoother transitions between feature resolutions while simultaneously increasing channel dimensionality.

\subsubsection{Activation Function: ReLU}

ReLU activation was selected for its simplicity, computational efficiency, and ability to mitigate vanishing gradients. Compared to sigmoid or tanh, ReLU avoids saturation and maintains stable gradient flow. More complex activations such as GELU or Swish offer marginal improvements at higher computational cost; given the multi-dataset training requirement, ReLU provides the most practical performance-to-efficiency ratio.

\subsubsection{Global Average Pooling over Other Pooling Methods}

Global Average Pooling (GAP) replaces both flattening and spatial pooling layers in the classifier head. GAP reduces each feature map to a single scalar, drastically reducing parameters and limiting overfitting. Unlike max pooling-which retains only the largest activation-GAP aggregates all spatial information, resulting in more stable feature summarization. It also aligns with modern classification architectures (e.g., ResNet, MobileNet \cite{pan2020new}), where GAP enhances interpretability and parameter efficiency.

\subsubsection{Classification Head and Softmax Output}

The fully connected head includes a 128-dimensional hidden layer, providing compact but expressive feature transformation before classification. A Softmax activation is applied at the output layer to convert logits into a normalized probability distribution over mutually exclusive classes. Softmax, combined with cross-entropy loss, provides smooth gradients and well-behaved optimization dynamics, making it the standard choice for multi-class image classification.

\subsubsection{Initialization and Optimization Strategy}

Kaiming initialization ensures stable activation variance through ReLU layers, improving early-stage training stability. The Adam optimizer is used due to its adaptive learning rate mechanism, which helps maintain robust training across heterogeneous datasets. Weight decay is selectively applied only to convolutional and linear weights, avoiding over-regularization of batch normalization and bias parameters.

\subsection{Transfer Learning Setup}
Transfer learning was used to evaluate how pretrained CNN models perform compared to training models from scratch. In this study, transfer learning was applied using ResNet50 and VGG16 models pretrained on the ImageNet dataset. ImageNet contains a large and diverse collection of images, allowing the models to learn general visual features that can be reused for the road condition classification task.

\vspace{5pt}

For the transfer learning setup, the pretrained backbone networks were frozen, and only the final classification layers were trained. This ensures that the models retain the general features learned from ImageNet while adapting only the final layers to the target dataset. In the ResNet50 transfer learning model, the original fully connected layer was replaced with a new classification head consisting of a dropout layer with a dropout rate of 0.5. During training, all layers except the final fully connected layer were frozen. As a result, only the final FC layer was trained, while the remaining backbone layers remained unchanged.

\vspace{5pt}
For the VGG16 transfer learning model, the pretrained convolutional feature extractor was frozen. The final layer of the classifier was replaced with a new linear layer to match the number of output classes. Only this final classifier layer was trained during the transfer learning process. A learning rate of 0.001 was used for the transfer learning models. This higher learning rate was chosen because only a small number of parameters were being updated, allowing the models to adapt more quickly to the new classification task. Dropout with a rate of 0.5 in the classification head was used to reduce overfitting. This transfer learning setup reduces training time, lowers the risk of overfitting, and provides a fair comparison with models trained from scratch.
\subsection{Training Configuration}
All experiments were conducted using Google Colab. Training was performed on a NVIDIA T4 GPU when available, which provided hardware acceleration for faster model training. If a GPU was not available, the code automatically fell back to CPU execution; however, all reported experiments were run using the GPU.

\vspace{5pt}
The input images were resized to 224 × 224 pixels to match the input size required by the pretrained CNN models. The dataset was split into training and validation sets using an 80/20 ratio, with stratified sampling to maintain class balance across splits. The validation set was used for model evaluation and comparison.
\vspace{5pt}
Data augmentation was applied to the training set to improve generalization. This included random horizontal flipping, random rotation, and color jittering. Both training and validation images were normalized using ImageNet mean and standard deviation values to ensure compatibility with pretrained models.

\vspace{5pt}
All models were trained using a batch size of 32 for 20 epochs. The cross-entropy loss function was used for binary classification. Optimization was performed using the Adam optimizer with weight decay to improve generalization. For models trained from scratch and the custom CNN, a learning rate of 0.0001 was used. For transfer learning models, a higher learning rate of 0.001 was applied since only the classification layers were trainable.

\vspace{5pt}
During training, model performance was monitored using training and validation accuracy and loss. The best-performing model for each architecture was saved based on the highest validation accuracy. After training, the saved best model was used for evaluation and comparison across all metrics. This training configuration ensured a consistent and fair comparison across all models while maintaining efficient training and stable convergence.

\subsection{Hyperparameter Tuning}
Hyperparameter tuning was performed to study the effect of learning rate and dropout on model performance. The tuning process focused on a small set of controlled experiments to ensure a fair comparison while keeping the training setup consistent across models. For the custom CNN, the learning rate and dropout probability were varied. The custom CNN includes dropout inside the residual blocks and in the classification head, allowing regularization strength to be adjusted. Two different dropout settings were tested: a head dropout of 0.5 to reduce overfitting and a head dropout of 0.0 to observe performance without dropout regularization. In both cases, a learning rate of 0.0001 was used, which is suitable for training models from scratch and provides stable convergence.

\vspace{5pt}
In addition, a higher learning rate of 0.001 was evaluated in combination with a head dropout of 0.5. This configuration was tested to analyze whether faster parameter updates, together with stronger regularization, could improve training efficiency and validation performance. Overall, the following hyperparameter combinations were evaluated:

\begin{itemize}
    \item Learning rate 0.0001 with head dropout 0.5
    \item Learning rate 0.0001 with head dropout 0.0
    \item Learning rate 0.001 with head dropout 0.5

\end{itemize}

All other training parameters, including batch size, number of epochs, optimizer, and data preprocessing, were kept constant across experiments. This controlled tuning strategy allowed the impact of learning rate and dropout on model performance to be clearly analyzed without introducing additional confounding factors.

\subsection{Evaluation Metrics}
To fairly compare the performance of all models, multiple evaluation metrics were used. These metrics evaluate classification accuracy, class-wise performance, probabilistic behavior, and computational efficiency.

\subsubsection{Accuracy}
Accuracy measures the percentage of correctly classified images out of the total number of images. It was calculated on the validation set and used as the primary metric for model comparison. During training, the model checkpoint with the highest validation accuracy was saved and later used for evaluation.

\subsubsection{Precision, Recall and F1-Score}
Precision, recall, and F1-score \cite{yacouby2020probabilistic} were computed to provide a more detailed evaluation of classification performance. Precision indicates how many predicted positive samples were correctly classified, while recall measures how many actual positive samples were correctly identified. The F1-score represents the harmonic mean of precision and recall. Weighted averaging was used to account for class imbalance.

\subsubsection{ROC Curve and ROC-AUC}
Receiver Operating Characteristic (ROC) \cite{fan2006understanding} curves were plotted using predicted class probabilities. The Area Under the ROC Curve (ROC-AUC) was calculated to measure the model’s ability to distinguish between the two classes across different classification thresholds. A higher ROC-AUC value indicates better class separability.

\subsubsection{Model Complexity and Training Time}
In addition to classification performance, model efficiency was also evaluated. For each model, the total number of parameters, number of trainable parameters, and approximate model size in megabytes were recorded. Training time was also measured to compare computational cost across models.

\section{Experimental Results}\label{sec3}

This section presents a comprehensive empirical evaluation of the proposed deep learning framework across five diverse computer–vision datasets developed for this study. These datasets span multiple real-world application domains in the Bangladeshi context, including unauthorized vehicle detection, footpath encroachment monitoring, Road Damage and Manhole Detection, mango cultivar classification and paddy variety identification. For each dataset, we report both dataset-level and class-wise performance metrics, including overall accuracy, precision, recall, F1-score, per-class accuracy, ROC–AUC, and PR–AUC. In addition, we analyze the model’s learning dynamics using training and validation curves, and further assess class-specific behavior through detailed confusion matrices. Together, these experiments offer a holistic understanding of model robustness, convergence stability, and discriminative capability across tasks of varying complexity, data imbalance, and inter-class visual similarity.

\subsection{Binary Classification: Auto-RickshawImageBD Dataset}

This section presents a comprehensive evaluation of the different CNN
models on Dataset~1 \cite{sukanto2025detectingunauthorizedvehiclesusing}, which contains two classes: \textit{autorickshaw} and 
\textit{non-autorickshaw}. The results include overall dataset-level metrics, model comparison and training behavior, and 
ROC--PR curve assessment. All reported values are derived directly from the 
model outputs obtained during experimentation.

\subsubsection{Overall Performance}

Table~\ref{tab:overall-performance} summarizes the overall performance of all evaluated models on the Autorickshaw dataset. The table presents classification performance along with computational characteristics, including training time, model size, and parameter counts. Across all models, transfer learning–based architectures achieve the highest accuracy values, with both ResNet50 (TL) and VGG16 (TL) obtaining an accuracy of 77.53\%. Models trained from scratch show slightly lower performance across all evaluation metrics.The Custom CNN models achieve comparable performance across different hyperparameter configurations, with accuracy values ranging between 73.41\% and 74.91\%. These models have significantly smaller model sizes and fewer parameters compared to ResNet50 and VGG16, while maintaining competitive classification performance. In terms of model complexity, ResNet50 and VGG16 contain substantially more total parameters than the Custom CNN. However, when transfer learning is applied, only a small fraction of parameters are trainable, as reflected in the reduced number of trainable parameters for the transfer learning configurations. Overall, the table highlights clear differences in classification performance, model complexity, and computational cost across the evaluated architectures for the Autorickshaw classification task.

\newpage
\begin{table}[ht]
\centering
\caption{Overall model performance comparison across all Models for the Unauthorised vehicle Dataset}
\label{tab:overall-performance}
\footnotesize
\begin{tabular}{@{}p{2.2cm}cccccccc@{}}
\toprule
Model & Acc. & Prec. & Recall & F1 & Time (s) & Total Par. & Train. Par. & Size (MB) \\
\midrule
Custom CNN (lr: 0.0001, do: 0.0) & 74.91 & 0.7362 & 0.7491 & 0.7363 & 3751.84 & 641304 & 641304 & 2.45 \\
Custom CNN (lr: 0.001, do: 0.5) & 74.91 & 0.7362 & 0.7491 & 0.7363 & 3761.96 & 641304 & 641304 & 2.45 \\
Custom CNN (lr: 0.0001, do: 0.5) & 73.41 & 0.7166 & 0.7341 & 0.7101 & 22103.51 & 641304 & 641304 & 2.45 \\
ResNet50 (Scratch) & 74.53 & 0.7389 & 0.7453 & 0.7082 & 4018.28 & 23512130 & 23512130 & 89.69 \\
ResNet50 (TL) & 77.53 & 0.7676 & 0.7753 & 0.769 & 3757.17 & 23512130 & 4098 & 89.69 \\
VGG16 (Scratch) & 76.03 & 0.7489 & 0.7603 & 0.7429 & 3901.49 & 134268738 & 134268738 & 512.19 \\
VGG16 (TL) & 77.53 & 0.7719 & 0.7753 & 0.7524 & 3812.42 & 134268738 & 8194 & 512.19 \\
\bottomrule
\end{tabular}
\end{table}


\subsubsection{Model Comparison and Accuracy Curve}

\begin{figure}[ht]
    \centering
    \begin{minipage}{0.48\textwidth}
        \centering
        \includegraphics[width=\textwidth]{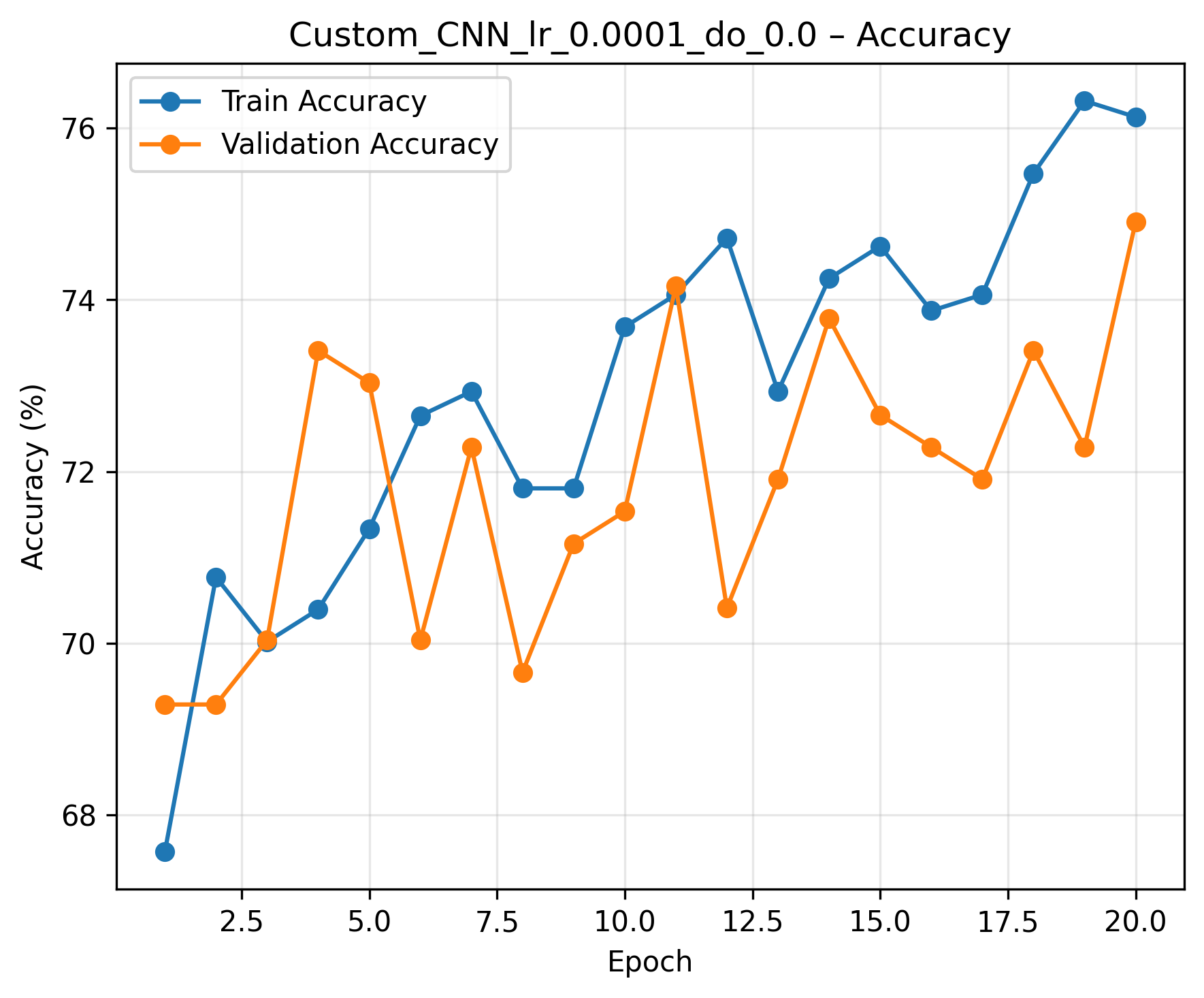}
    \end{minipage}
    \hfill
    \begin{minipage}{0.48\textwidth}
        \centering
        \includegraphics[width=\textwidth]{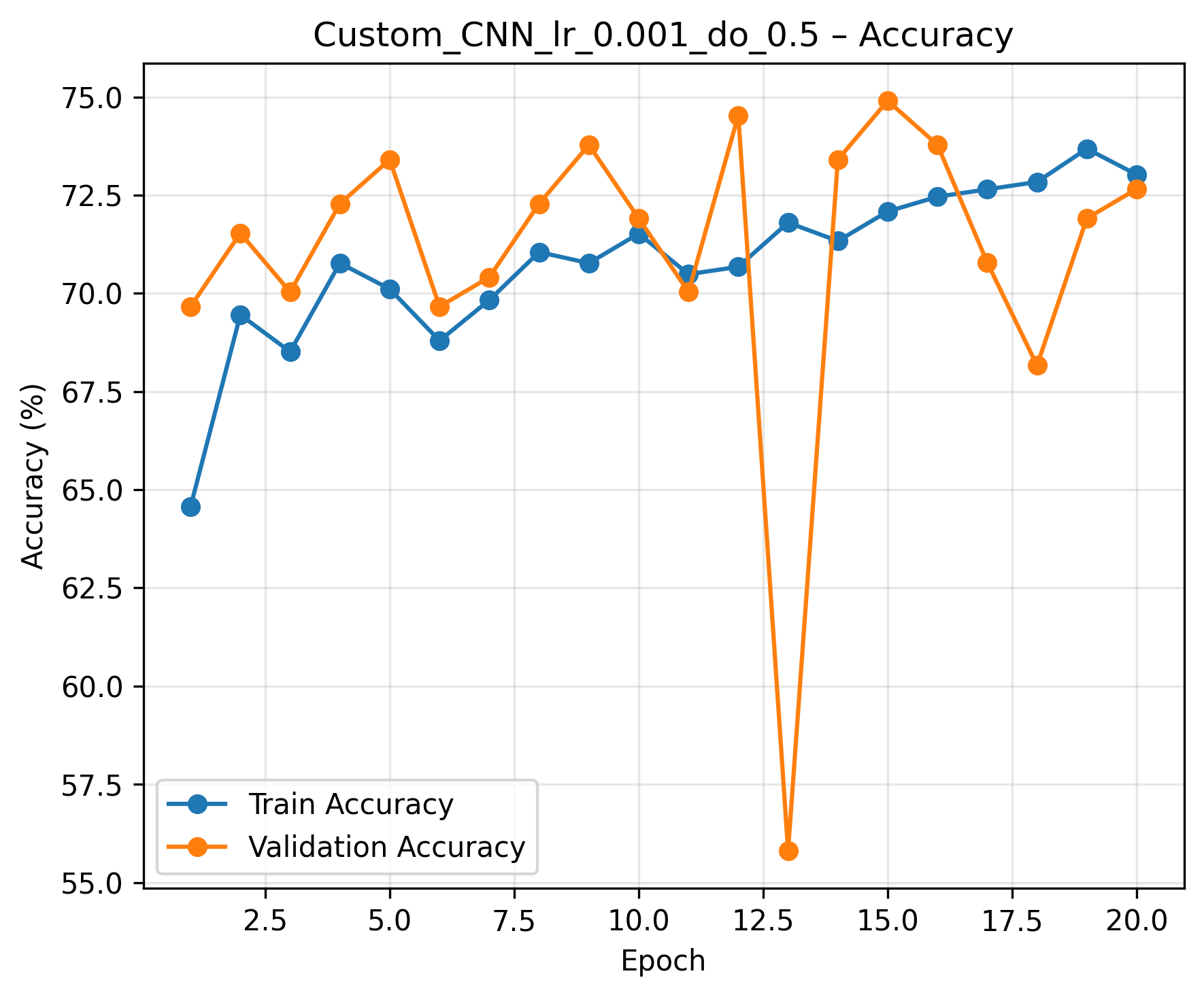}
    \end{minipage}
    \caption{Training vs Validation Accuracy for Custom CNN showing (a) Learning Rate: 0.0001, Dropout: 0.0 and (b) Learning Rate: 0.001, Dropout: 0.5.}
    \label{fig:auto_custom-cnn-1-2}
\end{figure}

\begin{figure}[ht]
    \centering
    \begin{minipage}{0.48\textwidth}
        \centering
        \includegraphics[width=\textwidth]{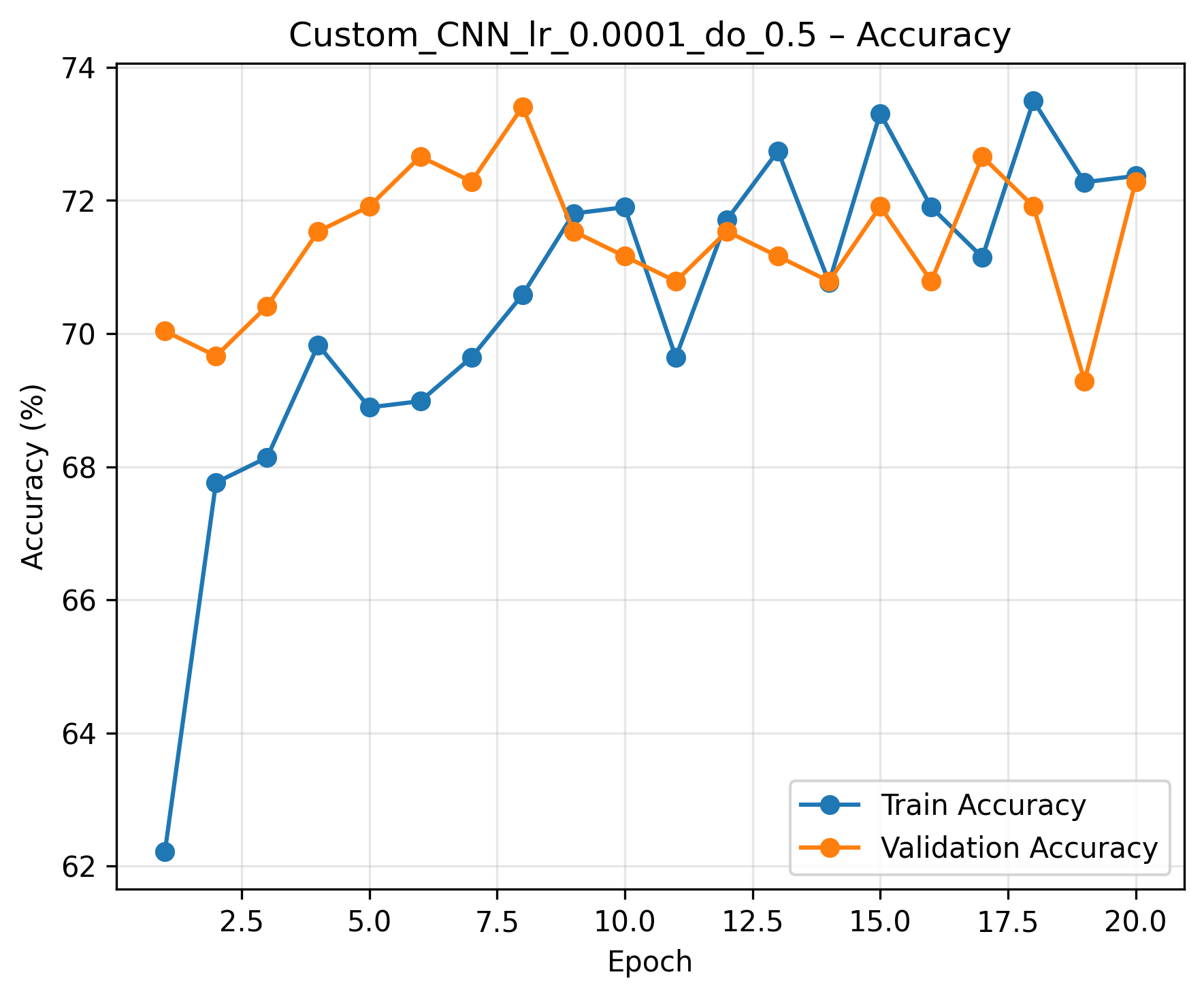}
    \end{minipage}
    \hfill
    \begin{minipage}{0.48\textwidth}
        \centering
        \includegraphics[width=\textwidth]{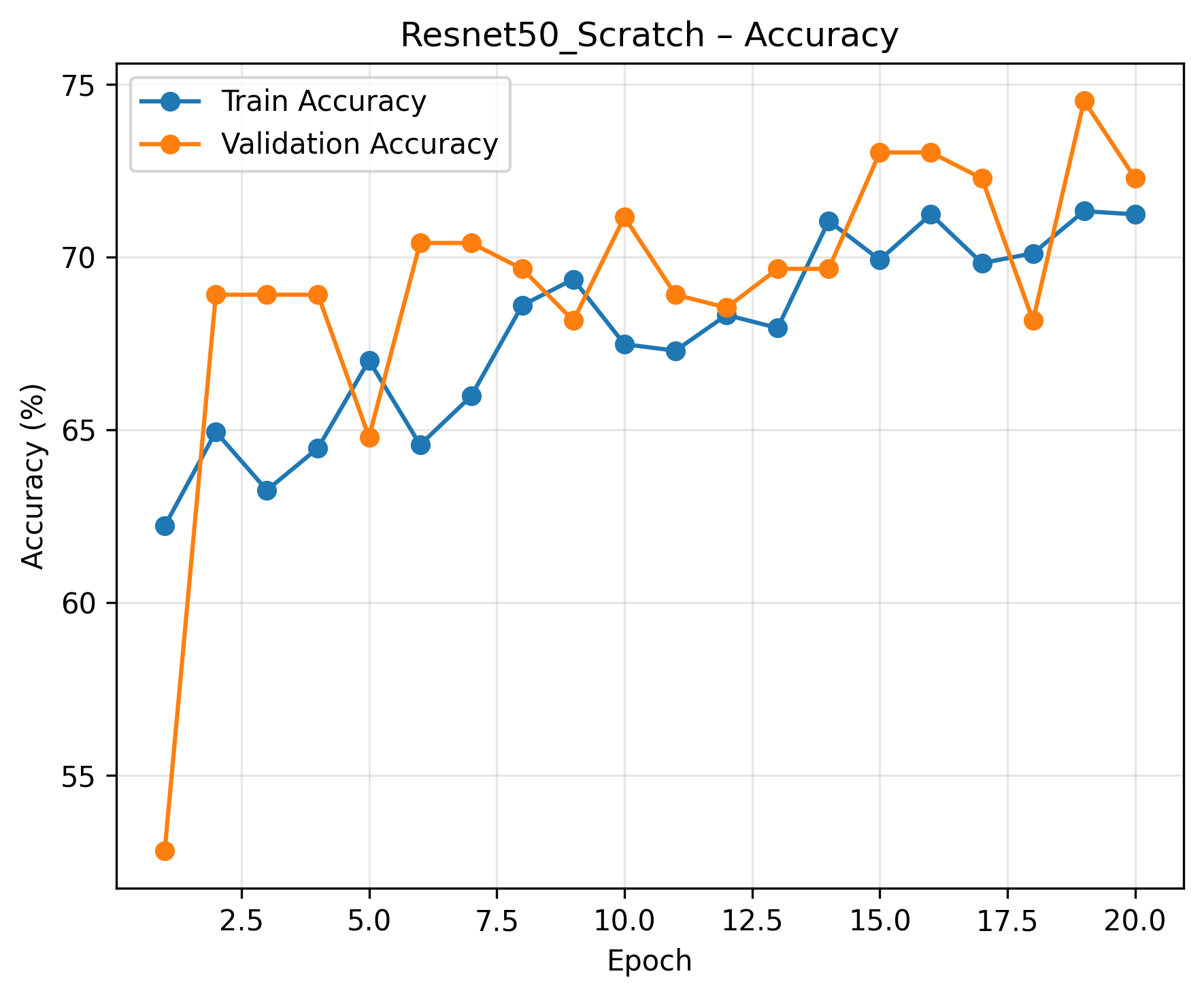}
    \end{minipage}
    \caption{Training vs Validation Accuracy showing (a) Custom CNN (Learning Rate: 0.0001, Dropout: 0.5) and (b) ResNet50 from scratch.}
    \label{fig:auto_custom-cnn-3-resnet-scratch}
\end{figure}

\begin{figure}[ht]
    \centering
    \begin{minipage}{0.48\textwidth}
        \centering
        \includegraphics[width=\textwidth]{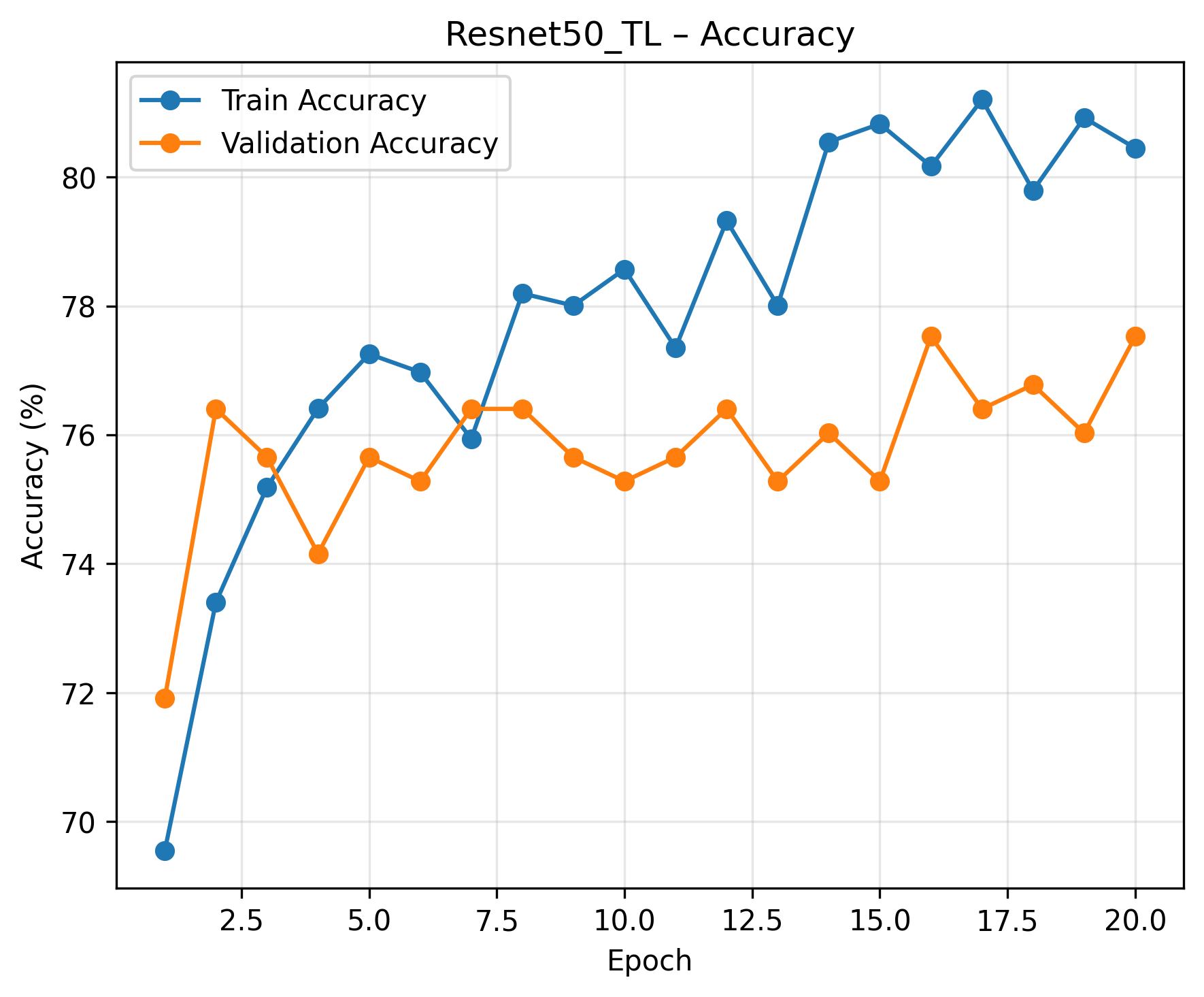}
    \end{minipage}
    \hfill
    \begin{minipage}{0.48\textwidth}
        \centering
        \includegraphics[width=\textwidth]{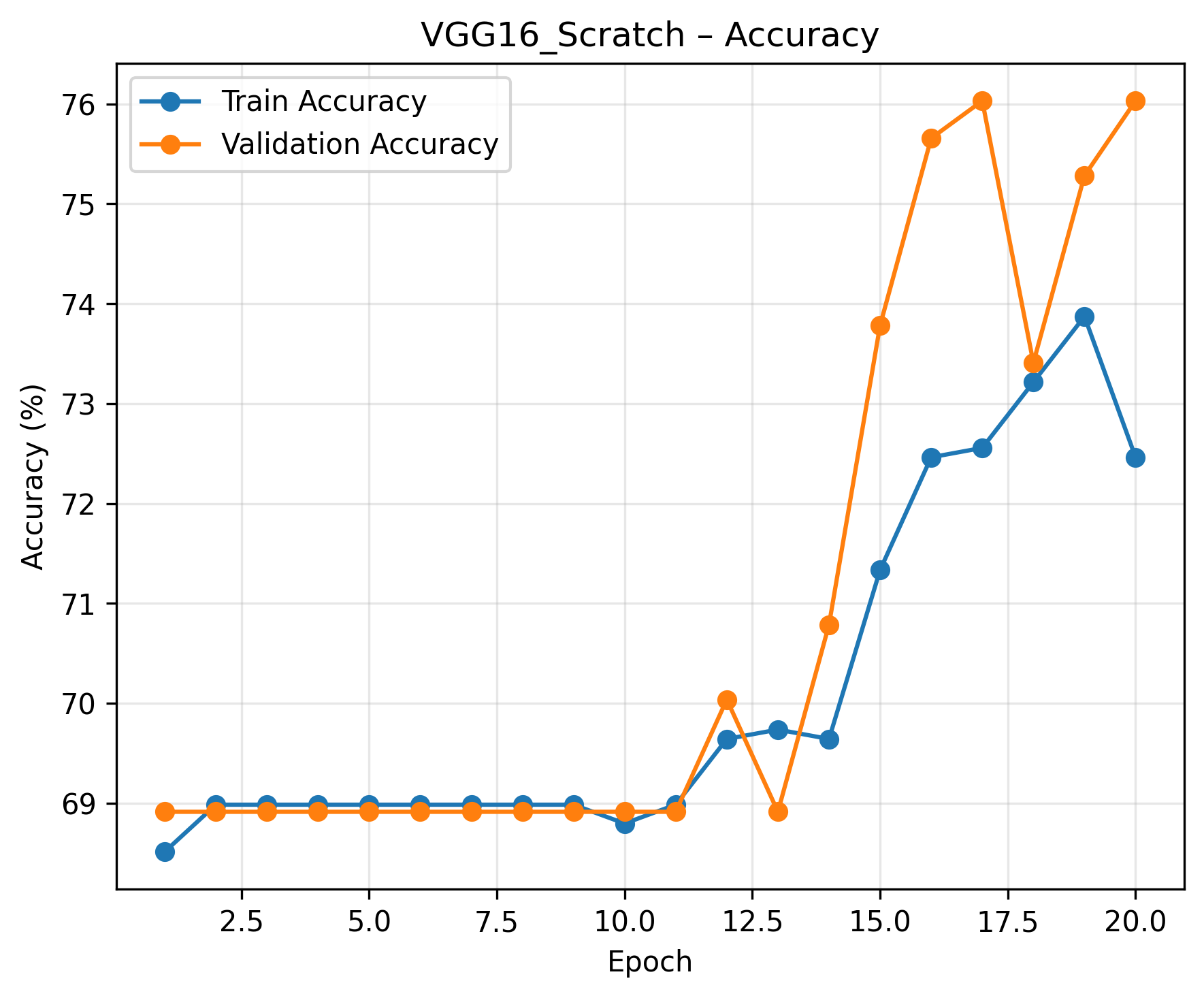}
    \end{minipage}
    \caption{Training vs Validation Accuracy showing (a) ResNet50 with Transfer Learning and (b) VGG16 from scratch.}
    \label{fig:auto_resnet-tl-vgg-scratch}
\end{figure}

\begin{figure}[ht]
    \centering
    \includegraphics[width=0.48\textwidth]{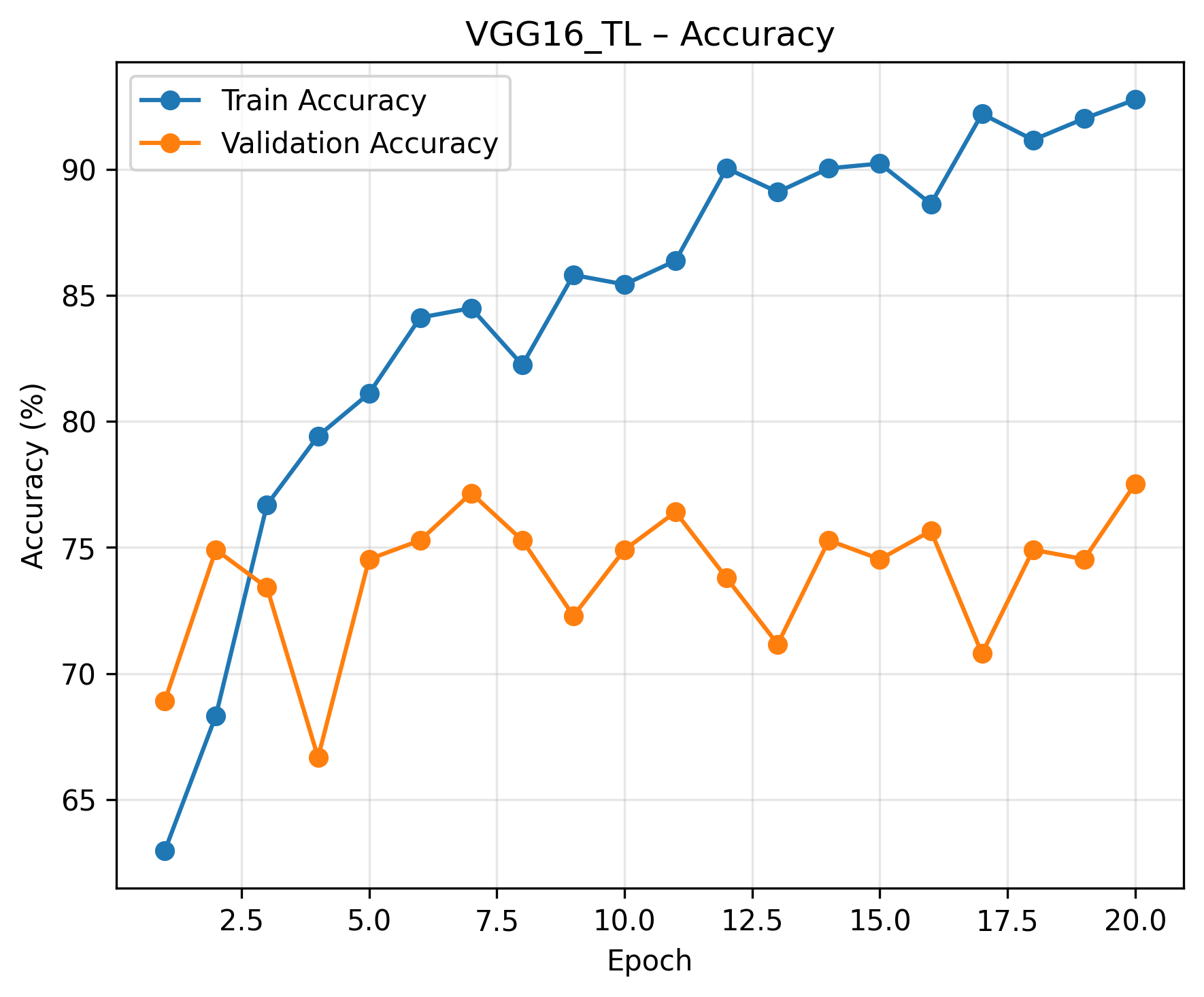}
    \caption{Training vs Validation Accuracy for VGG16 with Transfer Learning.}
    \label{fig:auto_vgg16-tl}
\end{figure}

Figures~\ref{fig:auto_custom-cnn-1-2} to \ref{fig:auto_vgg16-tl} illustrate the training and validation accuracy curves for all evaluated models on the autorickshaw versus non-autorickshaw dataset across 20 training epochs. The figures present a comparison of different architectures and training strategies, including custom CNN configurations, ResNet50, and VGG16, trained both from scratch and using transfer learning. For the Custom CNN models, all configurations show a steady increase in training accuracy over epochs, with validation accuracy following a similar trend but exhibiting minor fluctuations. Among the tested configurations, the model with a learning rate of 0.001 and dropout of 0.5 achieves higher validation accuracy compared to the other Custom CNN variants. The configuration without dropout shows slightly lower and more variable validation accuracy across epochs.

\vspace{10pt}
The ResNet50 model trained from scratch demonstrates gradual improvement in both training and validation accuracy, with validation accuracy stabilizing toward the later epochs. In contrast, the ResNet50 model with transfer learning shows faster convergence in training accuracy, while validation accuracy remains relatively stable after the initial epochs. For the VGG16 architecture, the model trained from scratch exhibits slower improvement in training accuracy, with validation accuracy increasing mainly in later epochs. The VGG16 model with transfer learning shows a rapid increase in training accuracy, reaching high values early in training, while validation accuracy remains comparatively stable with moderate variations across epochs.

\vspace{5pt}
Overall, the accuracy curves highlight differences in convergence behavior and stability across model architectures and training strategies for the autorickshaw classification task.

\newpage

\newpage
\subsubsection{ROC and Precision-Recall Analysis}

\begin{figure}[ht]
\centering
\includegraphics[width=0.48\textwidth]{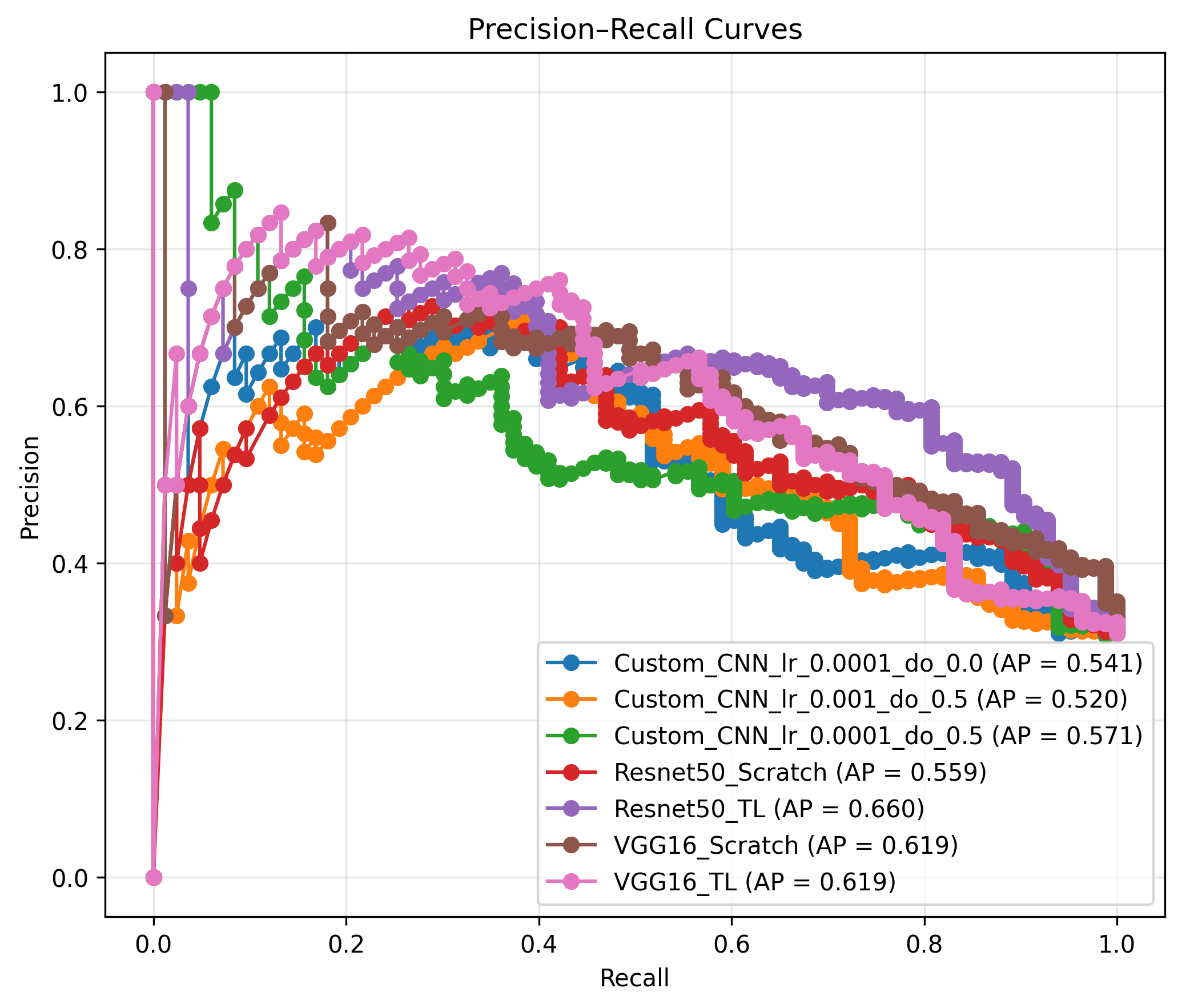}
\includegraphics[width=0.48\textwidth]{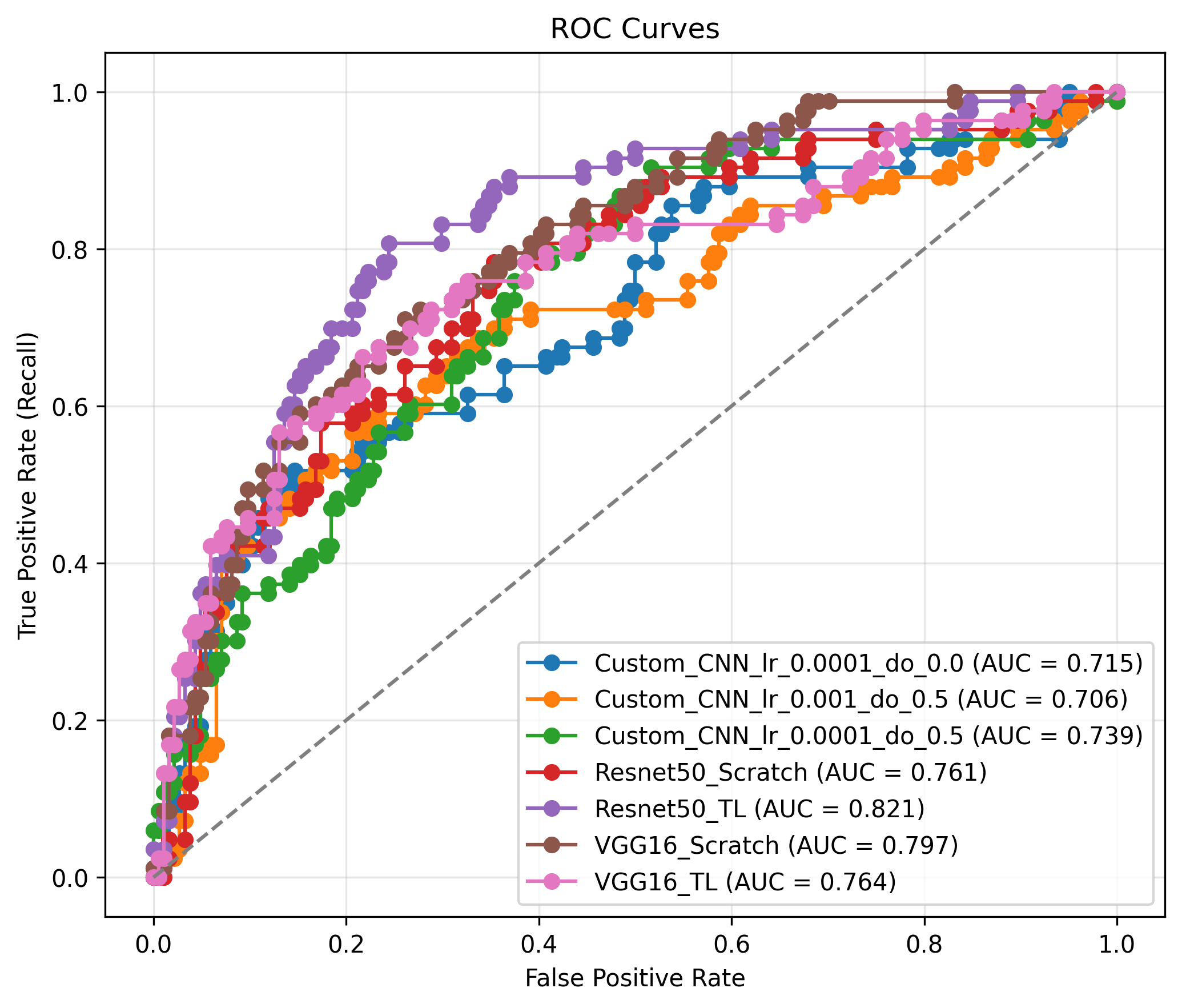}
\caption{Precision--recall curve (left) and ROC curve (right) for Dataset~1.}
\label{fig:ds1-roc-pr}
\end{figure}

Figure~\ref{fig:ds1-roc-pr} shows the Receiver Operating Characteristic (ROC) curves and Precision–Recall (PR) curves \cite{miao2022precision} for all evaluated models on Dataset~1, which consists of two classes: autorickshaw and non-autorickshaw. The ROC curves illustrate the trade-off between the true positive rate and false positive rate, while the PR curves show the relationship between precision and recall across different decision thresholds.

\vspace{5pt}
From the ROC curves, all models achieve performance above the random baseline, as indicated by curves lying above the diagonal reference line. Among the evaluated models, ResNet50 with transfer learning achieves the highest ROC-AUC value, followed by VGG16 trained from scratch and VGG16 with transfer learning. The Custom CNN configurations exhibit moderate ROC-AUC values, with the configuration using a learning rate of 0.0001 and dropout of 0.5 achieving the highest ROC-AUC among the custom models.

\vspace{5pt}
The Precision–Recall curves further highlight differences in performance across models, particularly at higher recall values. ResNet50 with transfer learning achieves the highest average precision (AP), maintaining higher precision over a wider range of recall levels. The VGG16 models achieve comparable AP values, while the Custom CNN variants show lower average precision compared to the pretrained architectures.

\vspace{5pt}
Overall, the ROC and PR curves provide a threshold-independent comparison of model performance for the autorickshaw classification task and highlight clear differences between model architectures and training strategies.

\subsection{Binary Classification: Footpath Dataset}

The second dataset, \textit{FootpathVision} \cite{LubainaFootpathVision}, captures diverse scenes of urban
footpaths in Bangladesh and presents a more challenging context due to varying
light conditions, mixed pedestrian–vehicle interactions, and structural
irregularities. Using this dataset, CNN models demonstrate
strong discriminative ability in identifying encroached versus non-encroached
footpath conditions. The following subsections summarize the dataset-level
performance and the
observed training dynamics.

\subsubsection{Overall Performance}
Table~\ref{tab:overall-performance} presents the overall performance of all evaluated models on the Footpath dataset. The results include classification metrics along with training time, model size, and parameter counts. Among all evaluated models, ResNet50 with transfer learning achieves the highest performance, recording an accuracy of 88.31\% and the highest F1-score. The VGG16 model with transfer learning also demonstrates strong performance, achieving an accuracy of 87.9\%. In contrast, the corresponding models trained from scratch achieve lower performance across all reported metrics.

\vspace{10pt}
The Custom CNN models show consistent performance across different hyperparameter configurations, with accuracy values ranging from 78.63\% to 82.66\%. The configuration using a learning rate of 0.001 and dropout of 0.5 achieves the highest performance among the Custom CNN variants. In terms of computational characteristics, the Custom CNN models have significantly smaller model sizes and fewer parameters compared to ResNet50 and VGG16. Transfer learning substantially reduces the number of trainable parameters for both ResNet50 and VGG16 while maintaining strong classification performance. Overall, the table highlights clear differences in classification performance, model complexity, and computational requirements across the evaluated architectures for the Footpath classification task.
\begin{table}[ht]
\centering
\caption{Overall model performance comparison across all Models for Footpath Dataset}
\label{tab:overall-performance}
\footnotesize
\begin{tabular}{@{}p{2.2cm}cccccccc@{}}
\toprule
Model & Acc. & Prec. & Recall & F1 & Time (s) & Total Par. & Train. Par. & Size (MB) \\
\midrule
Custom CNN (lr: 0.0001, do: 0.0) & 78.63 & 0.8003 & 0.7863 & 0.7877 & 3927.24 & 641304 & 641304 & 2.45 \\
Custom CNN (lr: 0.001, do: 0.5) & 82.66 & 0.8269 & 0.8266 & 0.8267 & 3953.29 & 641304 & 641304 & 2.45 \\
Custom CNN (lr: 0.0001, do: 0.5) & 81.85 & 0.8219 & 0.8185 & 0.815 & 3949.77 & 641304 & 641304 & 2.45 \\
ResNet50 (Scratch) & 80.24 & 0.8064 & 0.8024 & 0.8033 & 3967.34 & 23512130 & 23512130 & 89.69 \\
ResNet50 (TL) & 88.31 & 0.891 & 0.8831 & 0.8838 & 3613.37 & 23512130 & 4098 & 89.69 \\
VGG16 (Scratch) & 77.82 & 0.7823 & 0.7782 & 0.7792 & 4113.22 & 134268738 & 134268738 & 512.19 \\
VGG16 (TL) & 87.9 & 0.8899 & 0.879 & 0.8798 & 4001.34 & 134268738 & 8194 & 512.19 \\
\bottomrule
\end{tabular}
\end{table}


\subsubsection{Model Comparison and Accuracy Curve}

\begin{figure}[ht]
    \centering
    \begin{minipage}{0.48\textwidth}
        \centering
        \includegraphics[width=\textwidth]{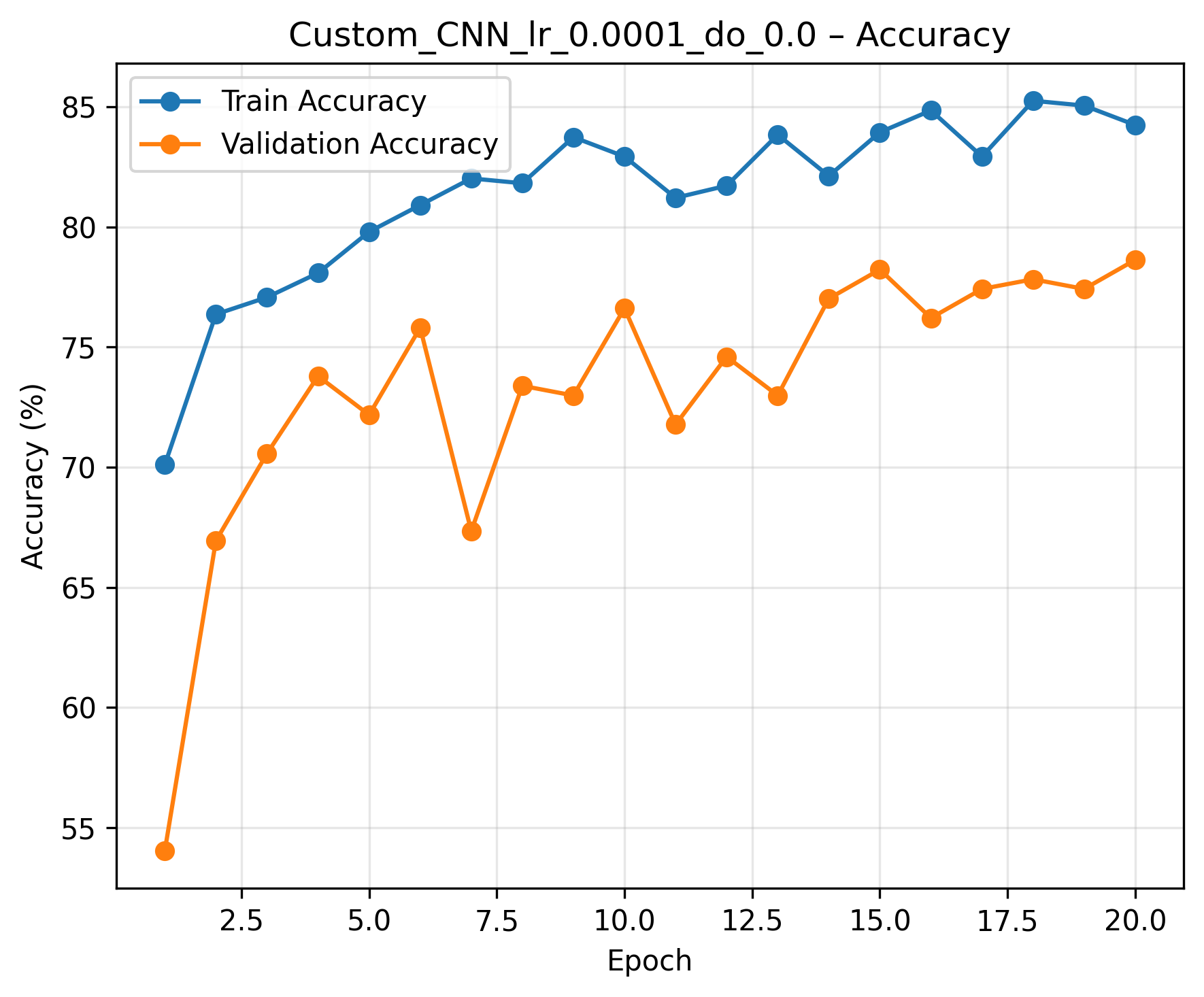}
    \end{minipage}
    \hfill
    \begin{minipage}{0.48\textwidth}
        \centering
        \includegraphics[width=\textwidth]{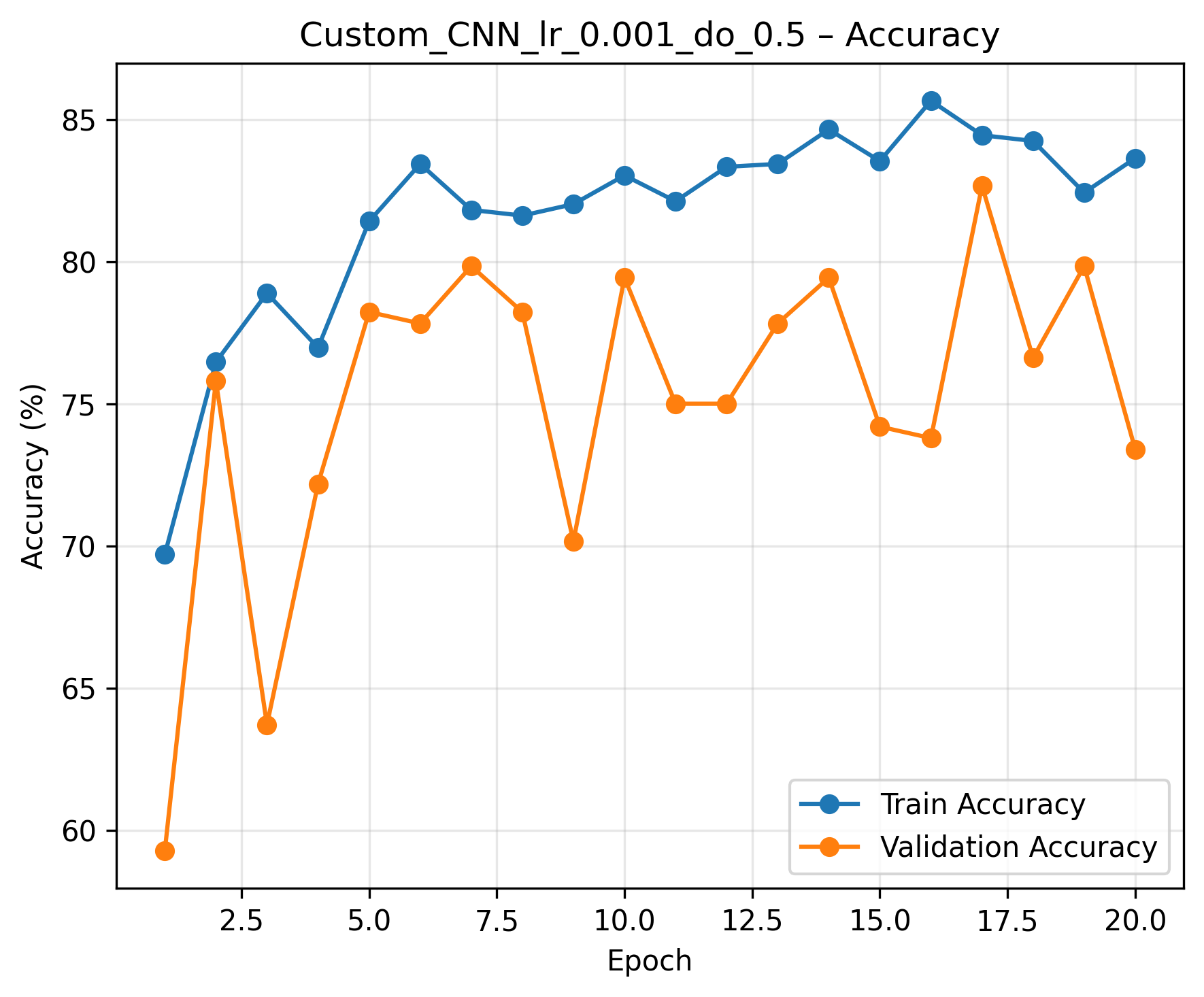}
    \end{minipage}
    \caption{Training vs Validation Accuracy for Custom CNN showing (a) Learning Rate: 0.0001, Dropout: 0.0 and (b) Learning Rate: 0.001, Dropout: 0.5.}
    \label{fig:Footpath_custom-cnn-1-2}
\end{figure}

\begin{figure}[ht]
    \centering
    \begin{minipage}{0.48\textwidth}
        \centering
        \includegraphics[width=\textwidth]{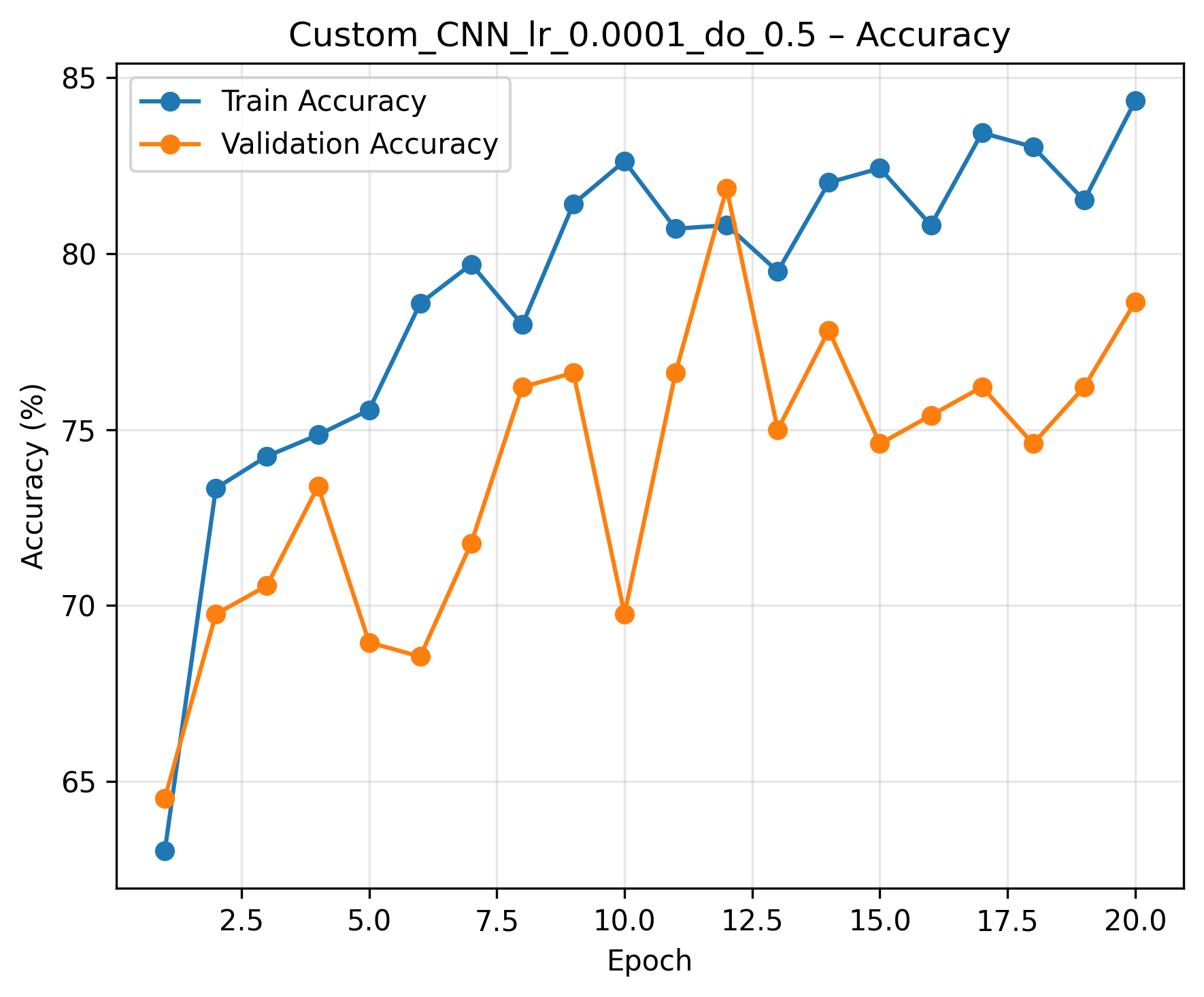}
    \end{minipage}
    \hfill
    \begin{minipage}{0.48\textwidth}
        \centering
        \includegraphics[width=\textwidth]{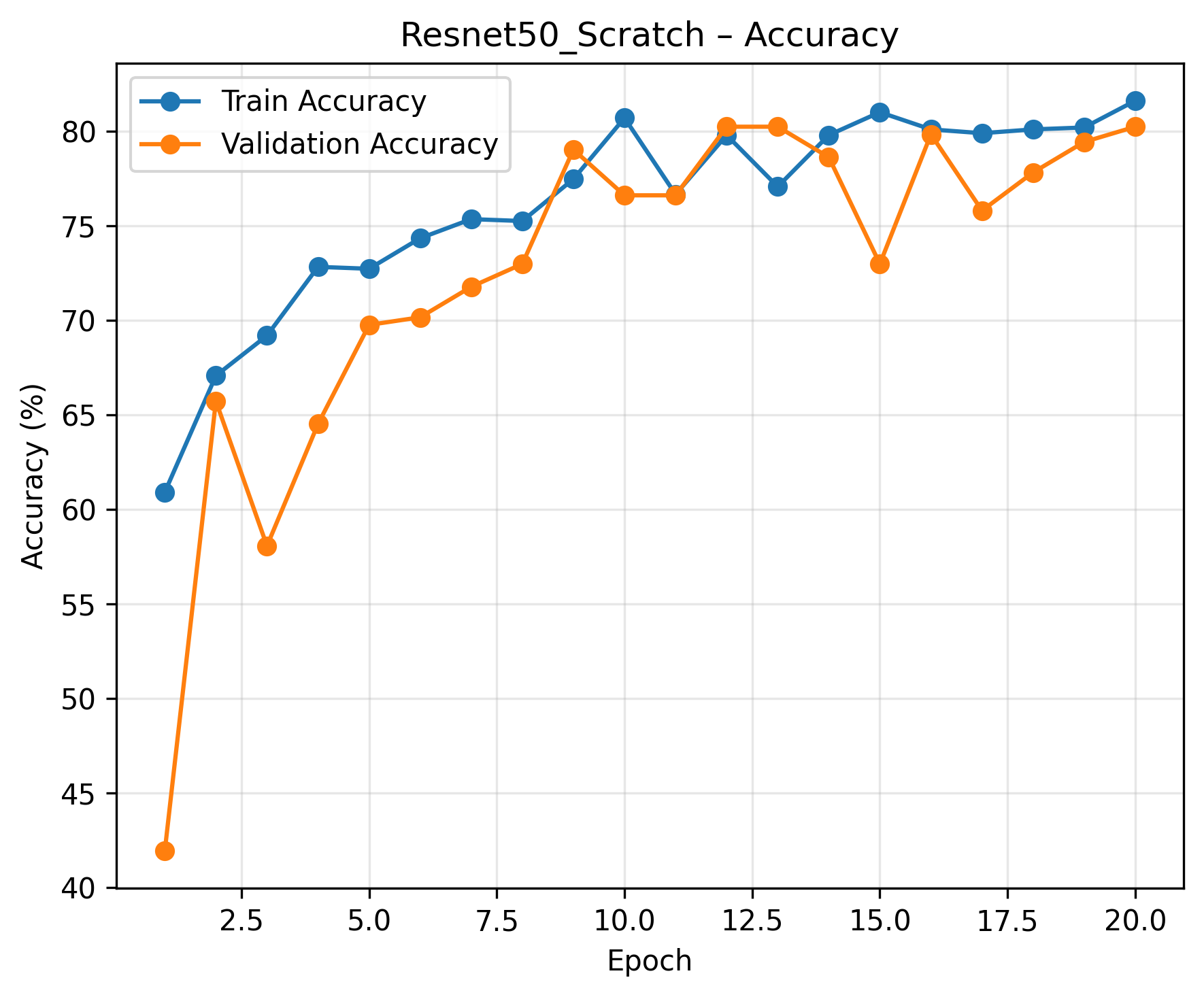}
    \end{minipage}
    \caption{Training vs Validation Accuracy showing (a) Custom CNN (Learning Rate: 0.0001, Dropout: 0.5) and (b) ResNet50 from scratch.}
    \label{fig:footpath_custom-cnn-3-resnet-scratch}
\end{figure}

\begin{figure}[ht]
    \centering
    \begin{minipage}{0.48\textwidth}
        \centering
        \includegraphics[width=\textwidth]{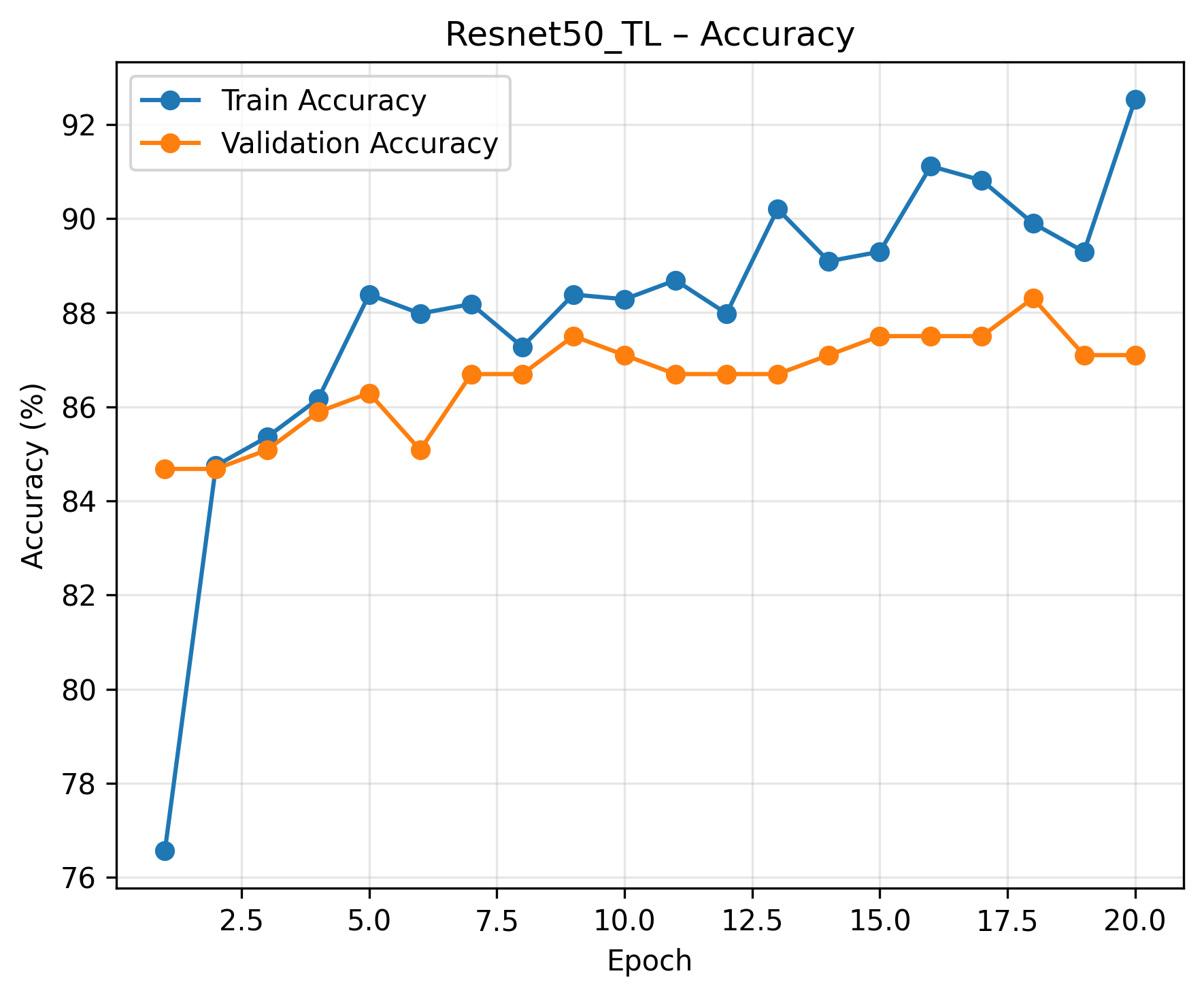}
    \end{minipage}
    \hfill
    \begin{minipage}{0.48\textwidth}
        \centering
        \includegraphics[width=\textwidth]{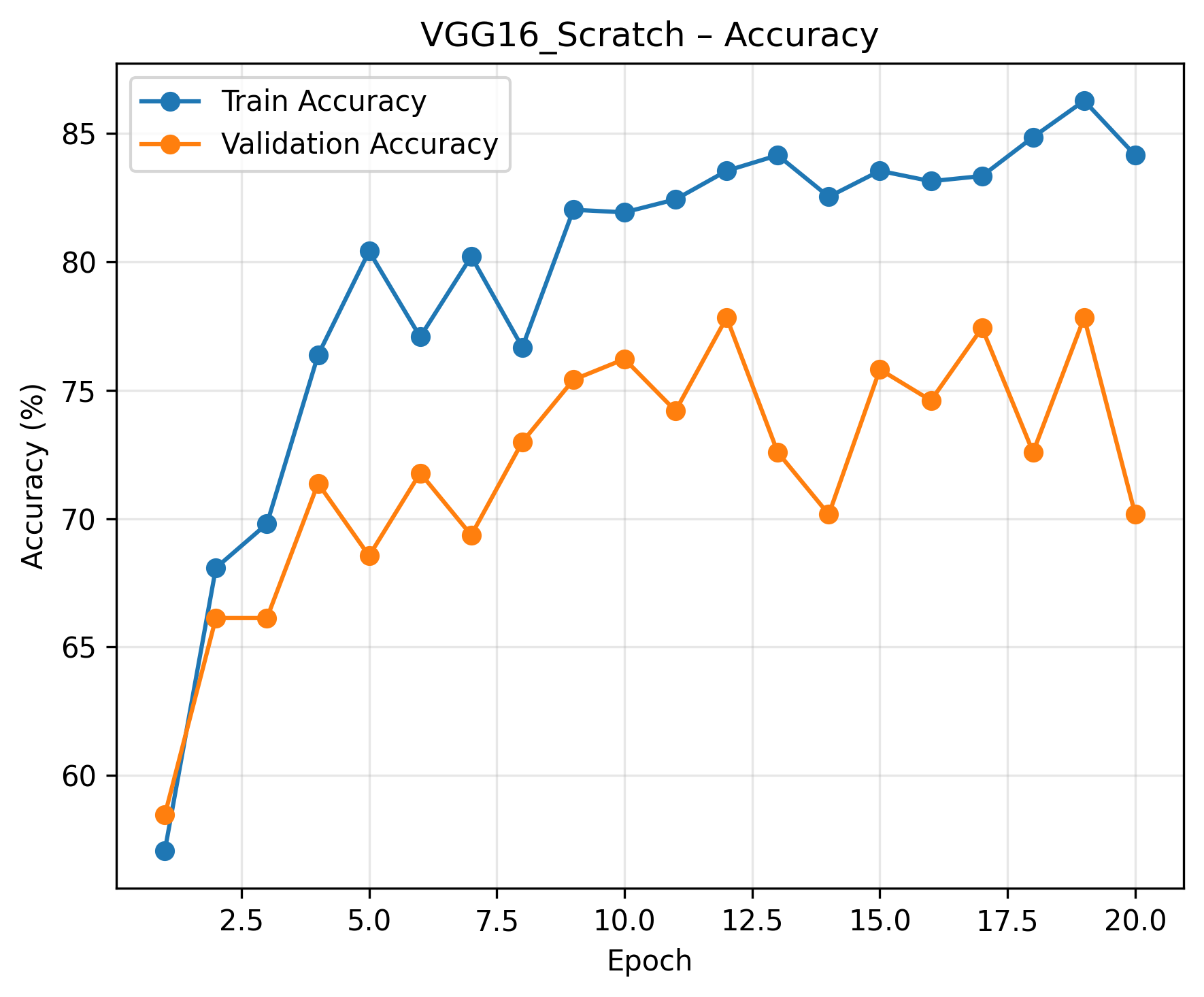}
    \end{minipage}
    \caption{Training vs Validation Accuracy showing (a) ResNet50 with Transfer Learning and (b) VGG16 from scratch.}
    \label{fig:footpath_resnet-tl-vgg-scratch}
\end{figure}

\begin{figure}[ht]
    \centering
    \includegraphics[width=0.48\textwidth]{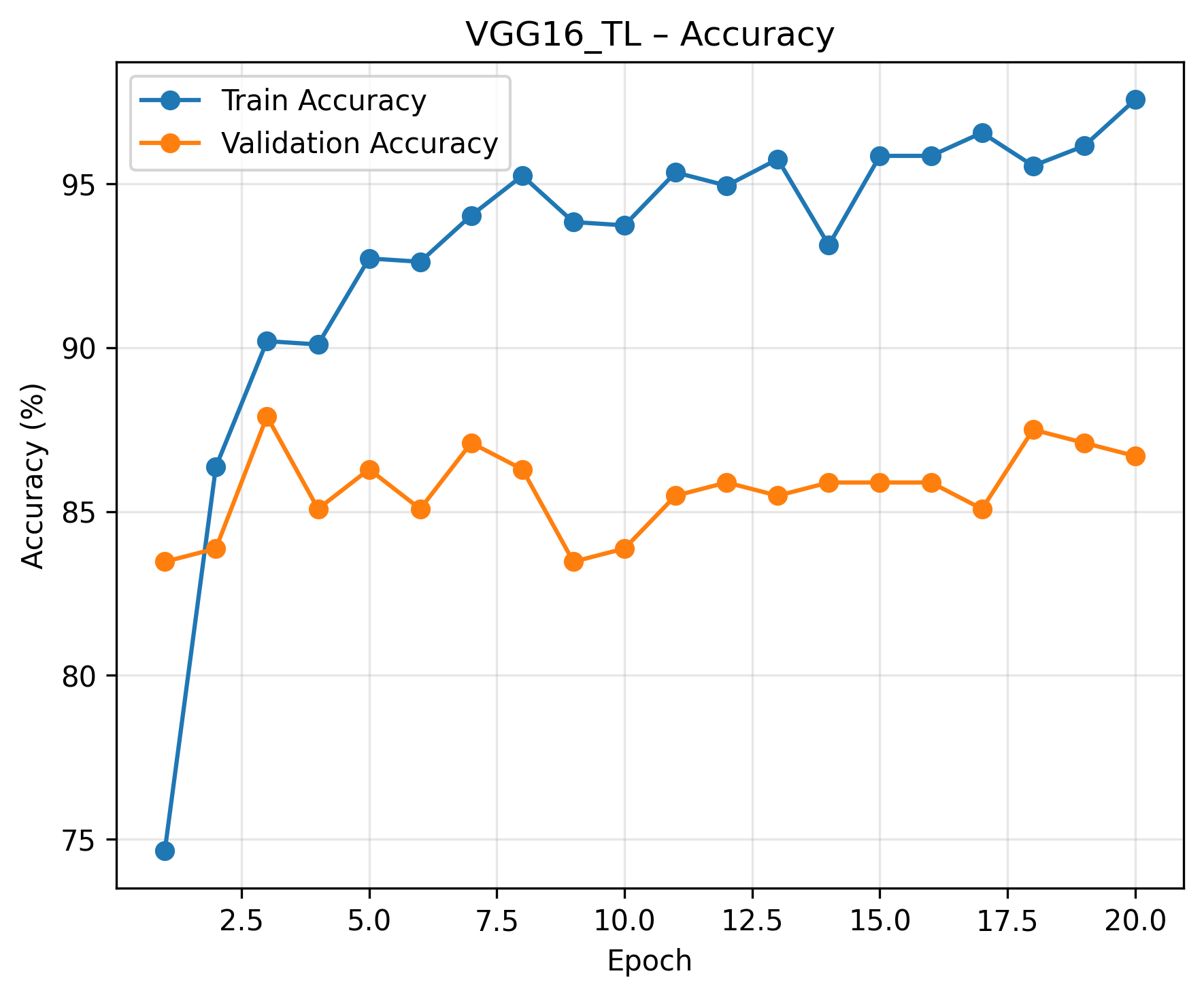 }
    \caption{Training vs Validation Accuracy for VGG16 with Transfer Learning.}
    \label{fig:footpath_vgg16-tl}
\end{figure}

Figures~\ref{fig:Footpath_custom-cnn-1-2} to \ref{fig:footpath_vgg16-tl} present the training and validation accuracy curves for all evaluated models on the Footpath Vision dataset, which consists of two classes: encroached and non-encroached. The figures show model performance over 20 training epochs for different architectures and training configurations.

\vspace{5pt}
For the Custom CNN models, all configurations demonstrate a consistent increase in training accuracy over epochs. Validation accuracy follows a similar trend, with noticeable fluctuations across epochs. Among the Custom CNN variants, the configuration using a learning rate of 0.001 and dropout of 0.5 achieves higher validation accuracy compared to the other configurations. The model without dropout shows relatively lower and more variable validation accuracy during training.

\vspace{5pt}
The ResNet50 model trained from scratch exhibits steady improvement in both training and validation accuracy, with validation accuracy gradually increasing and stabilizing toward later epochs. In comparison, the ResNet50 model with transfer learning shows rapid improvement in training accuracy within the initial epochs, while validation accuracy remains relatively stable across the training process.

\vspace{5pt}
For the VGG16 architecture, the model trained from scratch shows gradual growth in training accuracy, with validation accuracy increasing at a slower rate. The VGG16 model with transfer learning achieves high training accuracy early in training, while validation accuracy remains consistent with moderate variations across epochs. The accuracy curves illustrate differences in convergence behavior and stability across model architectures and training strategies for the encroached versus non-encroached footpath classification task.

\newpage
\subsubsection{ROC and Precision--Recall Analysis}
\begin{figure}[ht]
\centering
\includegraphics[width=0.48\textwidth]{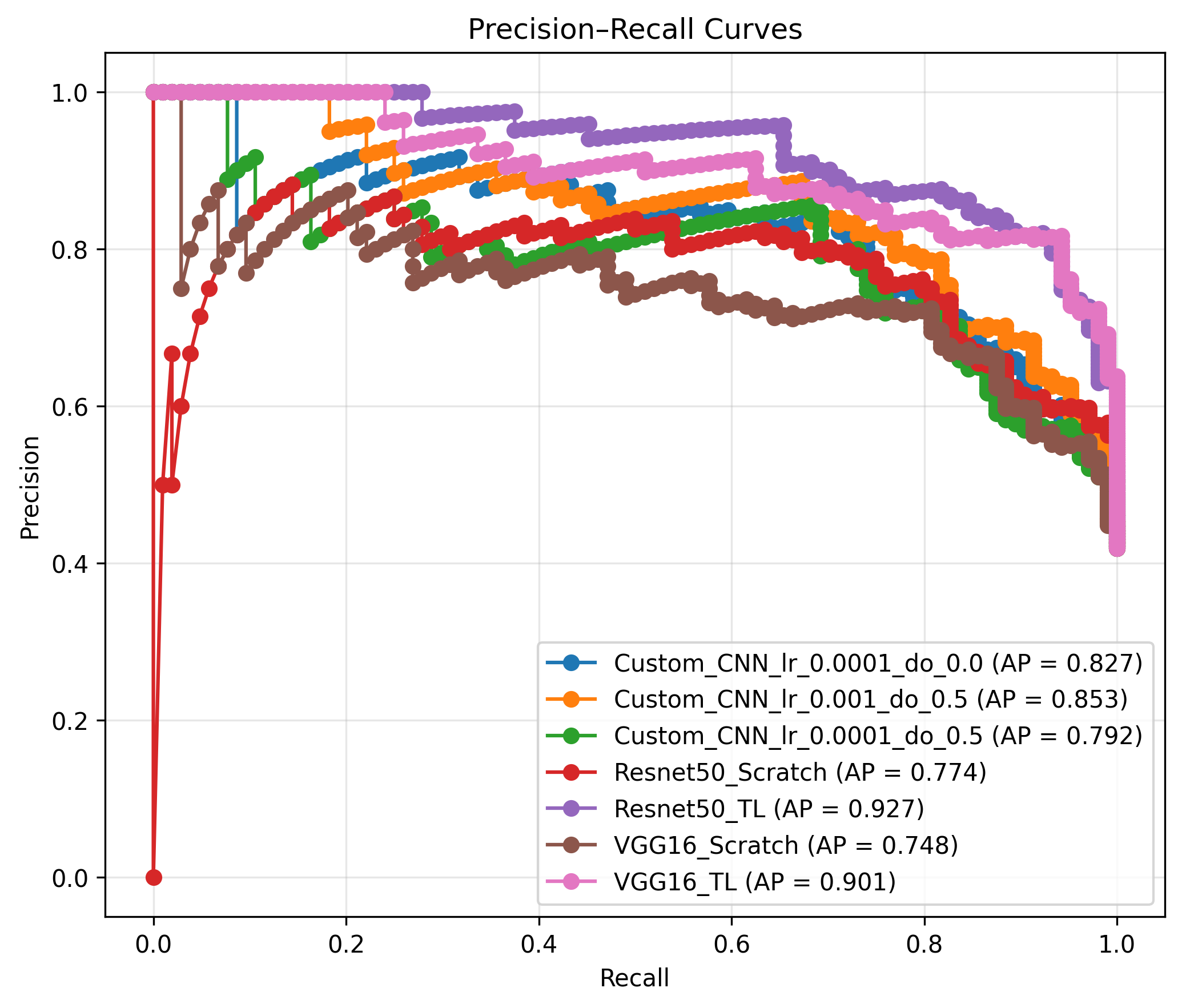}
\includegraphics[width=0.48\textwidth]{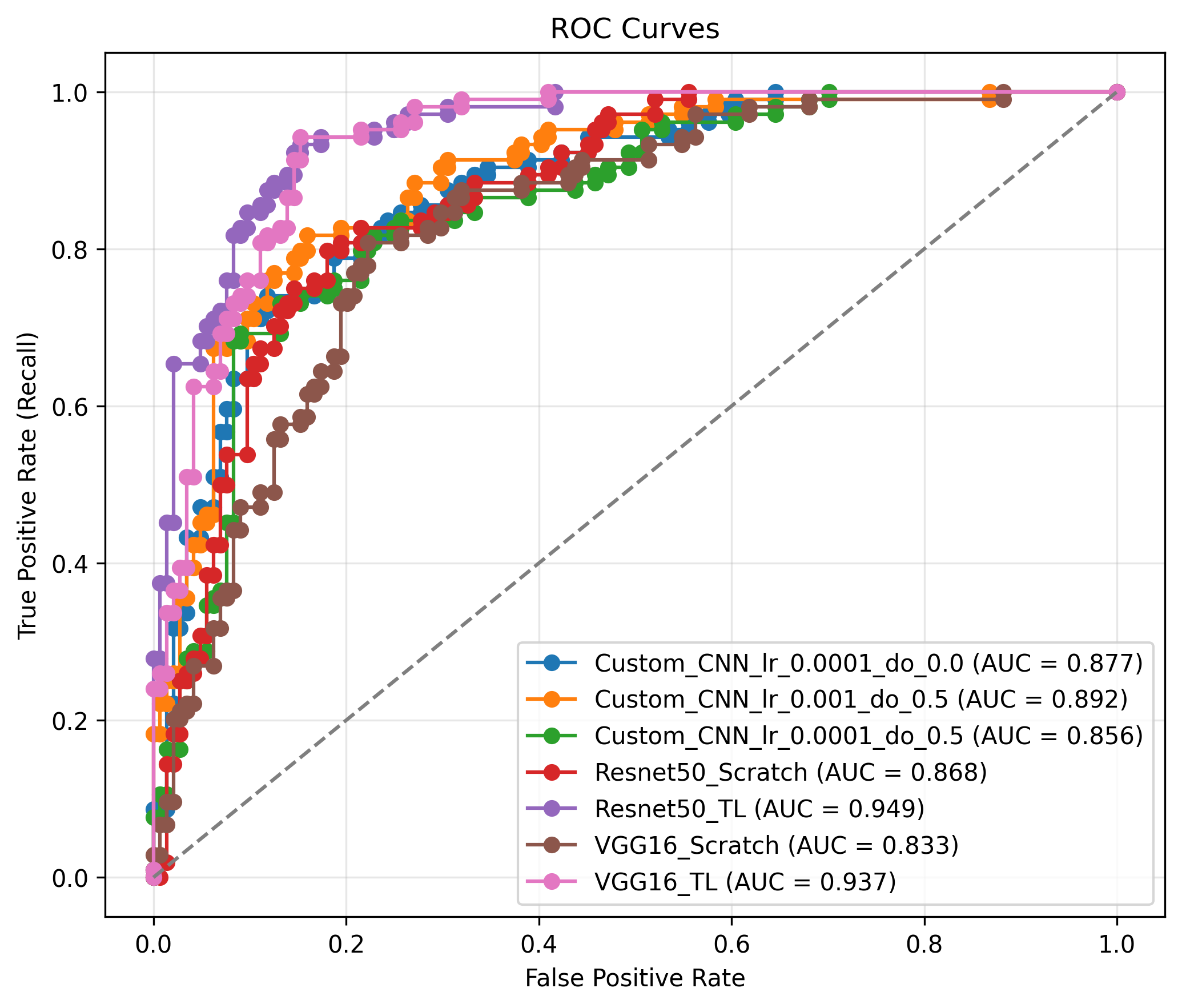}
\caption{\centering Precision--recall curve (left) and ROC curve (right) for Dataset~2.}
\label{fig:ds2-roc-pr}
\end{figure}

Figure~\ref{fig:ds2-roc-pr} presents the Receiver Operating Characteristic (ROC) curves and Precision–Recall (PR) curves for all evaluated models on Dataset~2, which consists of two classes: encroached and non-encroached. The ROC curves illustrate the relationship between the true positive rate and false positive rate, while the PR curves show the trade-off between precision and recall across different classification thresholds.

\vspace{5pt}
From the ROC curves, all models achieve performance above the random baseline, as indicated by curves lying well above the diagonal reference line. ResNet50 with transfer learning achieves the highest ROC-AUC value, followed by VGG16 with transfer learning. The Custom CNN configurations obtain slightly lower ROC-AUC values, with the configuration using a learning rate of 0.001 and dropout of 0.5 achieving the highest ROC-AUC among the custom models.

\vspace{5pt}
The Precision–Recall curves show a similar trend across models. ResNet50 with transfer learning achieves the highest average precision (AP), maintaining higher precision across a wide range of recall values. The VGG16 model with transfer learning also shows strong PR performance. The Custom CNN variants demonstrate moderate average precision, with differences observed across the tested hyperparameter configurations.

\vspace{10pt}
Overall, the ROC and PR curves provide a threshold-independent evaluation of model performance for the footpath encroachment classification task and highlight clear differences among model architectures and training strategies.

\subsection{Binary Classification: Road Damage Dataset}
This section presents a comprehensive evaluation of the \textit{Custom\_CNN}
model on Dataset~3 \cite{hossen2025roaddamagemanholedetection}. The dataset consists of road surface images collected and organized into two categories: GoodRoads and DamagedRoads. Each class contains images representing different road conditions. The GoodRoads class includes images of smooth, well-maintained road segments, while the DamagedRoads class contains images showing visible defects such as cracks, potholes, and surface deterioration. The images are stored in a folder-based structure, allowing automatic label assignment during training.
This dataset is designed for binary image classification, enabling a model to distinguish between normal and damaged road conditions based solely on visual features.
\subsubsection{Overall Performance}

Table~\ref{tab:overall-performance} summarizes the overall performance of all evaluated models on the Road dataset. The results are reported using classification metrics along with training time, model size, and parameter counts. Among all evaluated models, VGG16 with transfer learning achieves the highest performance, recording the highest accuracy of 97.78\%, precision, recall, and F1-score. ResNet50 with transfer learning also demonstrates strong performance, achieving an accuracy of 96.67\% and comparable precision and recall values. Models trained from scratch show slightly lower performance across all metrics.

\vspace{5pt}
The Custom CNN models achieve competitive results, with accuracy values ranging from 92.22\% to 95.56\%. The configuration using a learning rate of 0.001 and dropout of 0.5 yields the best performance among the Custom CNN variants, achieving an accuracy of 95.56\% and a high F1-score. In terms of computational efficiency, the Custom CNN models have significantly smaller model sizes (2.45 MB) and fewer parameters compared to ResNet50 (89.69 MB) and VGG16 (512.19 MB). Training times across all models are comparable, with transfer learning models generally requiring slightly less training time than models trained from scratch.
\begin{table}[ht]
\centering
\caption{Overall model performance comparison across all Models for Road Dataset}
\label{tab:overall-performance}
\footnotesize
\begin{tabular}{@{}p{2.2cm}cccccccc@{}}
\toprule
Model & Acc. & Prec. & Recall & F1 & Time (s) & Total Par. & Train. Par. & Size (MB) \\
\midrule
Custom CNN (lr: 0.0001, do: 0.0) & 93.33 & 0.939 & 0.9333 & 0.9303 & 1620.23 & 641304 & 641304 & 2.45 \\
Custom CNN (lr: 0.001, do: 0.5) & 95.56 & 0.9581 & 0.9556 & 0.9543 & 1583.81 & 641304 & 641304 & 2.45 \\
Custom CNN (lr: 0.0001, do: 0.5) & 92.22 & 0.9298 & 0.9222 & 0.9179 & 1589.91 & 641304 & 641304 & 2.45 \\
ResNet50 (Scratch) & 92.22 & 0.9216 & 0.9222 & 0.9206 & 1585.61 & 23512130 & 23512130 & 89.69 \\
ResNet50 (TL) & 96.67 & 0.9665 & 0.9667 & 0.9665 & 1398.76 & 23512130 & 4098 & 89.69 \\
VGG16 (Scratch) & 94.44 & 0.9484 & 0.9444 & 0.9424 & 1475.68 & 134268738 & 134268738 & 512.19 \\
VGG16 (TL) & 97.78 & 0.9778 & 0.9778 & 0.9778 & 1619.87 & 134268738 & 8194 & 512.19 \\
\bottomrule
\end{tabular}
\end{table}

\subsubsection{Model Comparison and Accuracy Curves}

\begin{figure}[ht]
    \centering
    \begin{minipage}{0.48\textwidth}
        \centering
        \includegraphics[width=\textwidth]{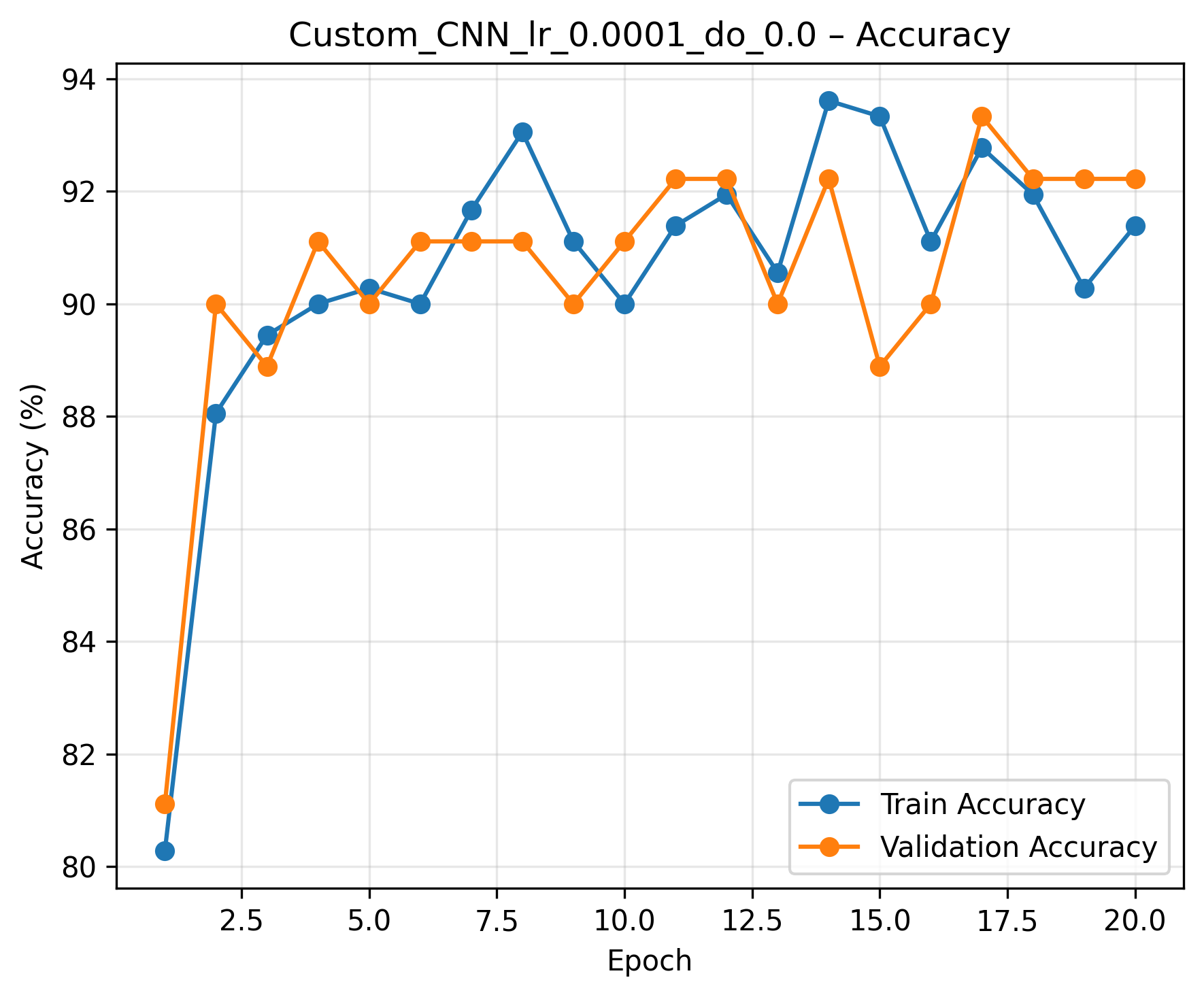}
    \end{minipage}
    \hfill
    \begin{minipage}{0.48\textwidth}
        \centering
        \includegraphics[width=\textwidth]{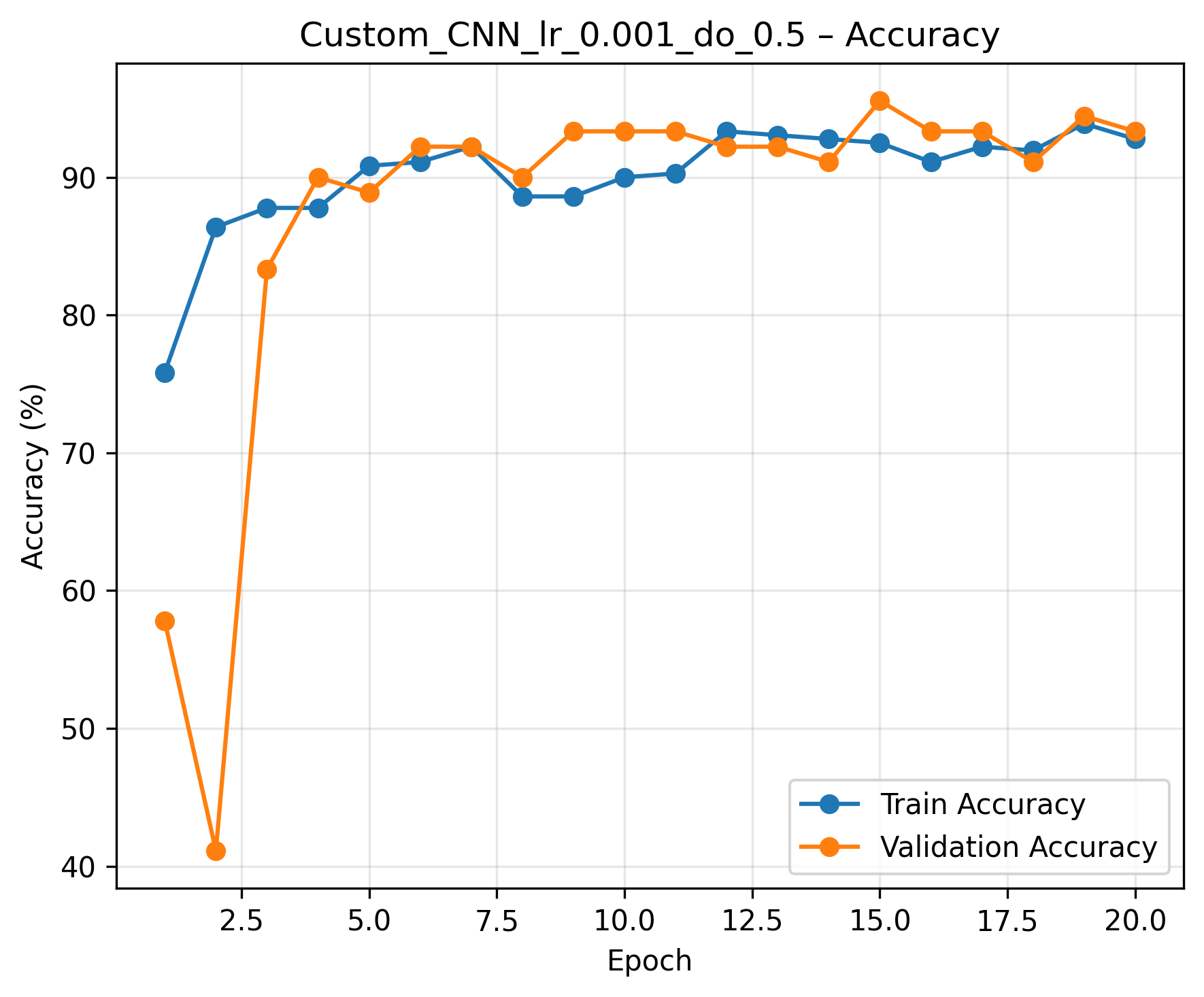}
    \end{minipage}
    \caption{Training vs Validation Accuracy for Custom CNN showing (a) Learning Rate: 0.0001, Dropout: 0.0 and (b) Learning Rate: 0.001, Dropout: 0.5.}
    \label{fig:road_custom-cnn-1-2}
\end{figure}

Figures~\ref{fig:road_custom-cnn-1-2} present the training and validation accuracy curves for the Custom CNN under different learning rate and dropout configurations. In both cases, training accuracy increases steadily over epochs, indicating effective learning. The validation accuracy follows a similar trend with minor fluctuations, suggesting stable generalization. The configuration with dropout shows slightly smoother validation behavior in later epochs.

\begin{figure}[ht]
    \centering
    \begin{minipage}{0.48\textwidth}
        \centering
        \includegraphics[width=\textwidth]{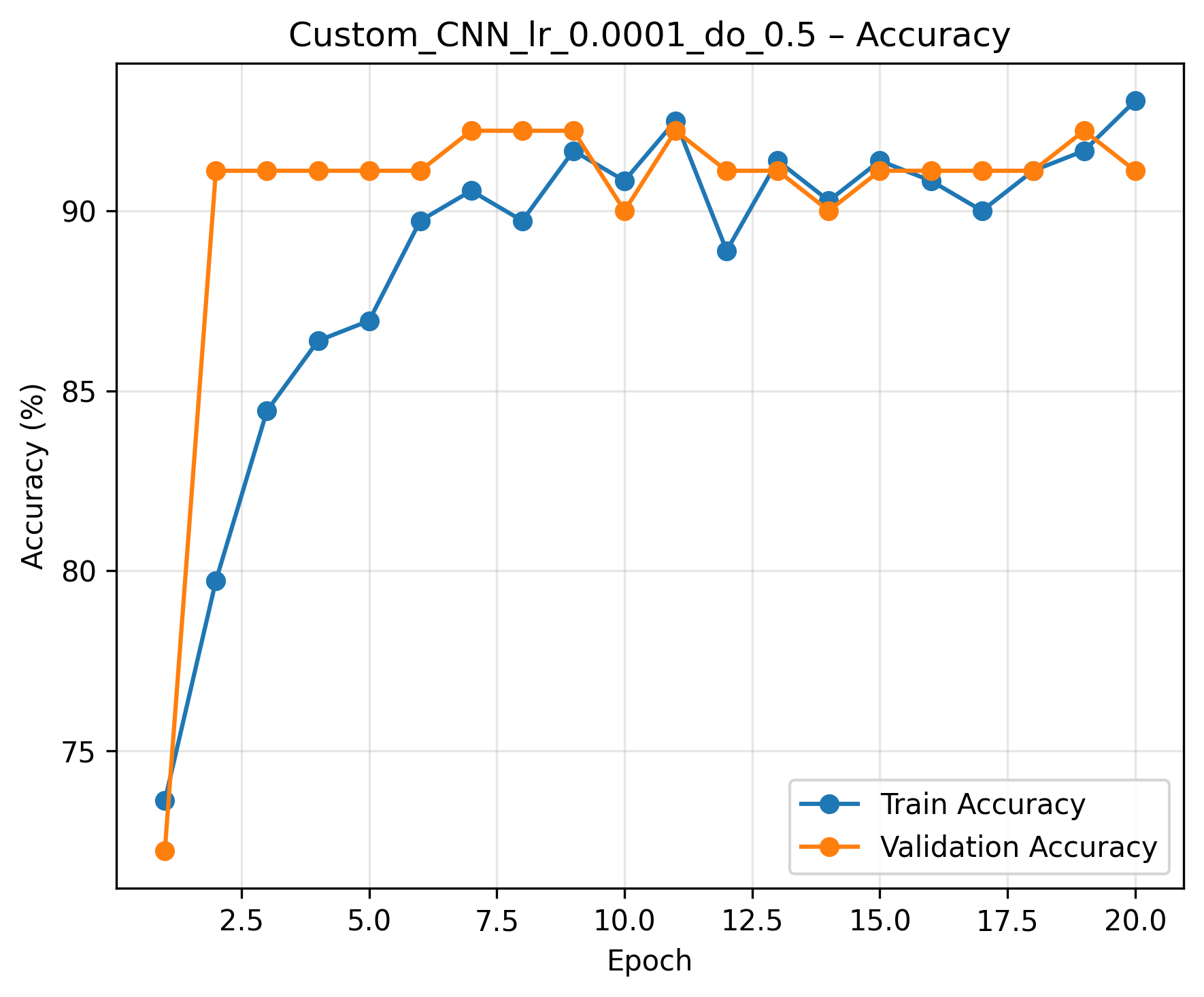}
    \end{minipage}
    \hfill
    \begin{minipage}{0.48\textwidth}
        \centering
        \includegraphics[width=\textwidth]{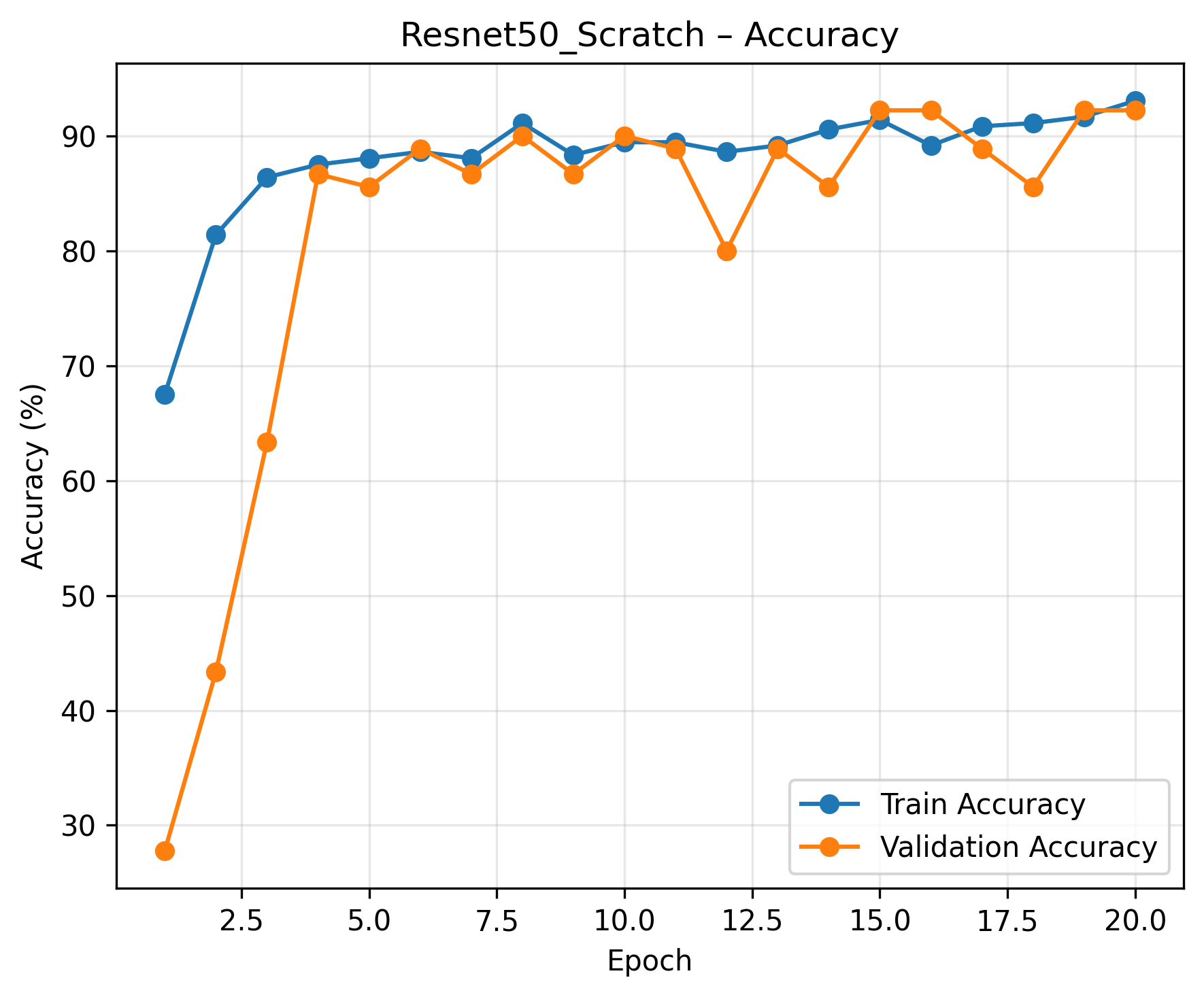}
    \end{minipage}
    \caption{Training vs Validation Accuracy showing (a) Custom CNN (Learning Rate: 0.0001, Dropout: 0.5) and (b) ResNet50 from scratch.}
    \label{fig:road_custom-cnn-3-resnet-scratch}
\end{figure}

\vspace{5pt}
Figure~\ref{fig:road_custom-cnn-3-resnet-scratch} compares the Custom CNN with ResNet50 trained from scratch. Both models demonstrate consistent growth in training accuracy across epochs. However, the Custom CNN exhibits more stable validation accuracy, while ResNet50 from scratch shows noticeable oscillations, particularly in the early training stages.

\begin{figure}[ht]
    \centering
    \begin{minipage}{0.48\textwidth}
        \centering
        \includegraphics[width=\textwidth]{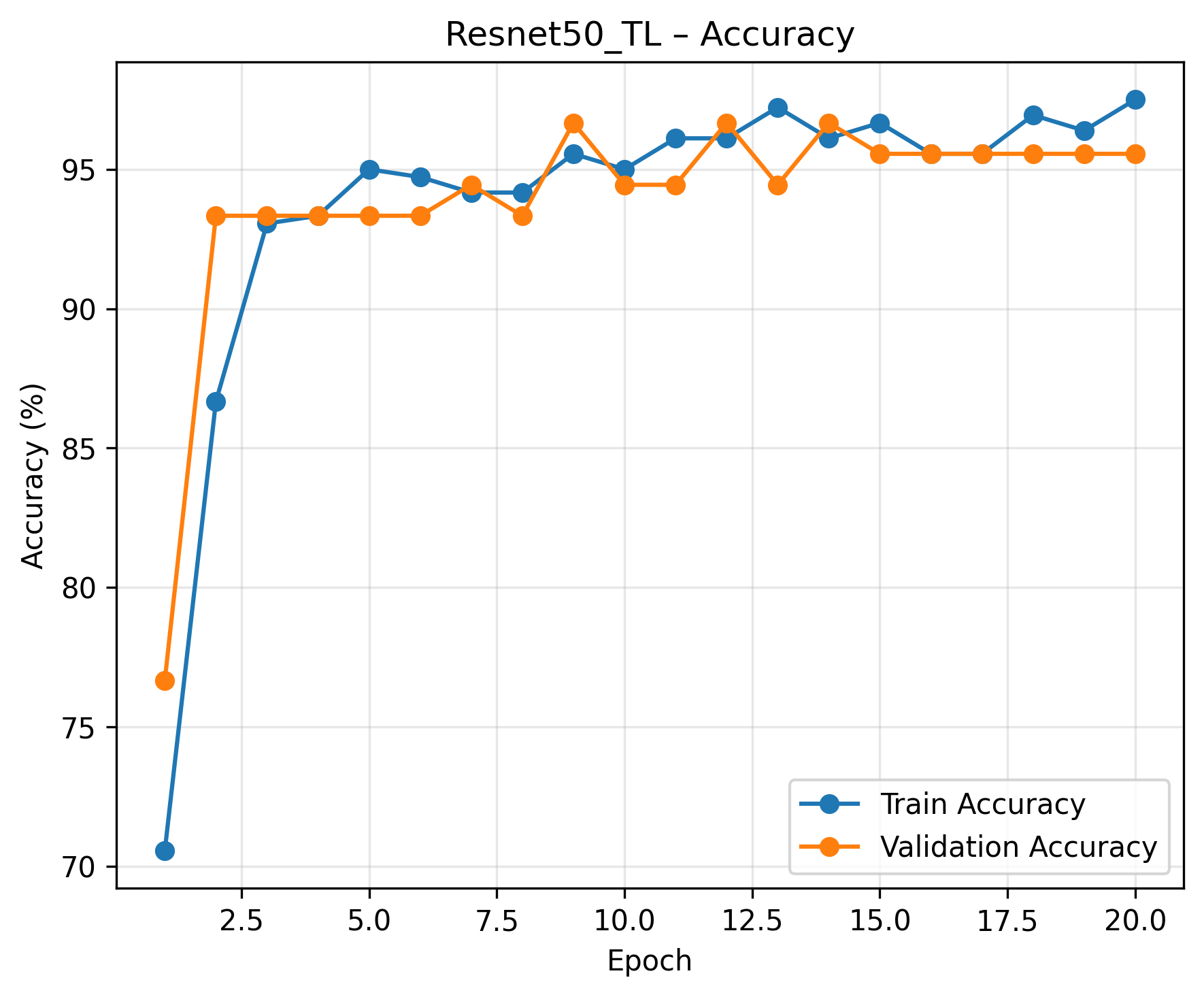}
    \end{minipage}
    \hfill
    \begin{minipage}{0.48\textwidth}
        \centering
        \includegraphics[width=\textwidth]{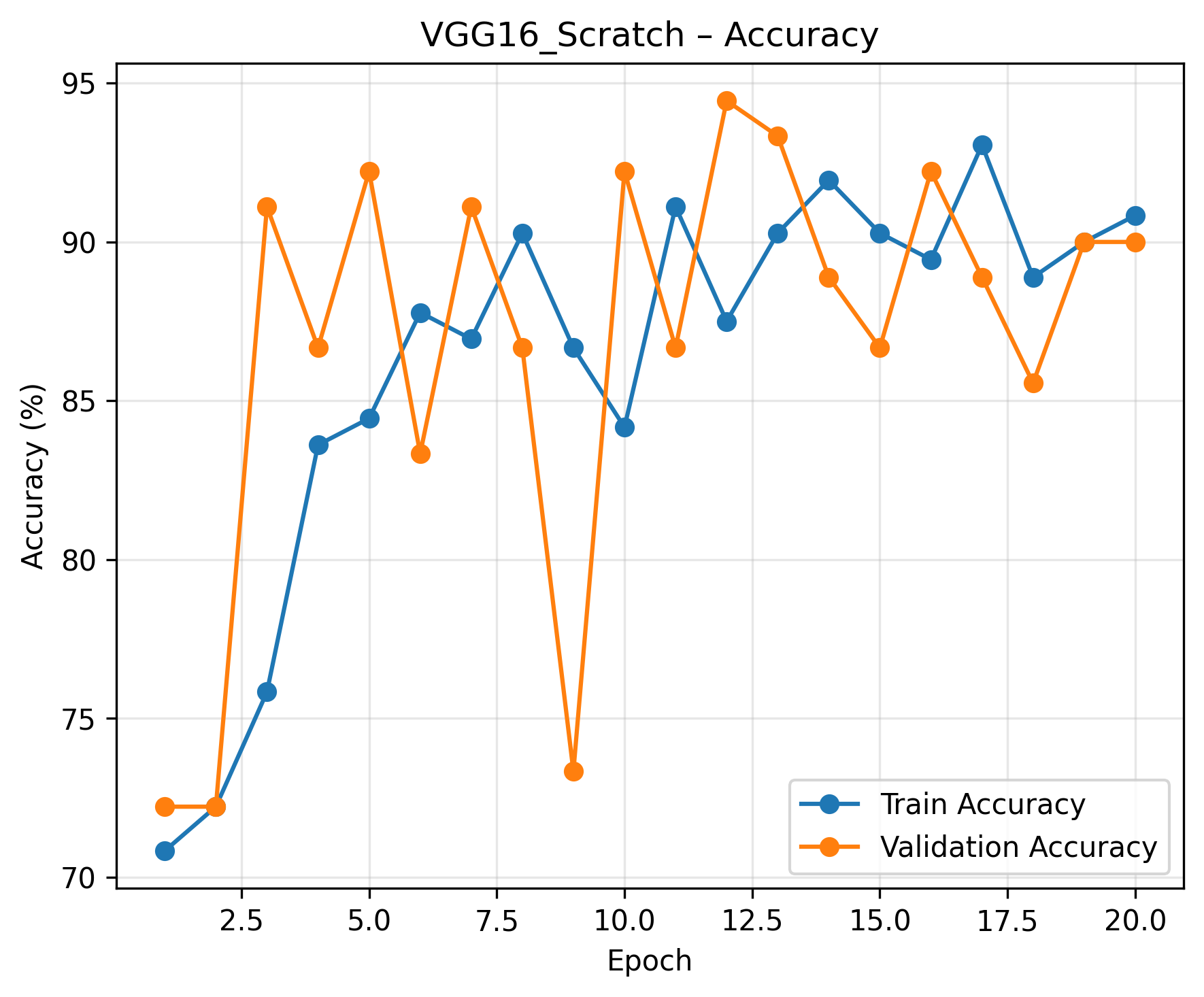}
    \end{minipage}
    \caption{Training vs Validation Accuracy showing (a) ResNet50 with Transfer Learning and (b) VGG16 from scratch.}
    \label{fig:road_resnet-tl-vgg-scratch}
\end{figure}

\vspace{5pt}
The comparison between transfer learning and training from scratch is shown in Figure~\ref{fig:road_resnet-tl-vgg-scratch}. ResNet50 with transfer learning achieves rapid convergence and maintains high validation accuracy throughout training. In contrast, VGG16 trained from scratch shows slower convergence and higher variability in validation accuracy across epochs.

\begin{figure}[ht]
    \centering
    \includegraphics[width=0.7\textwidth]{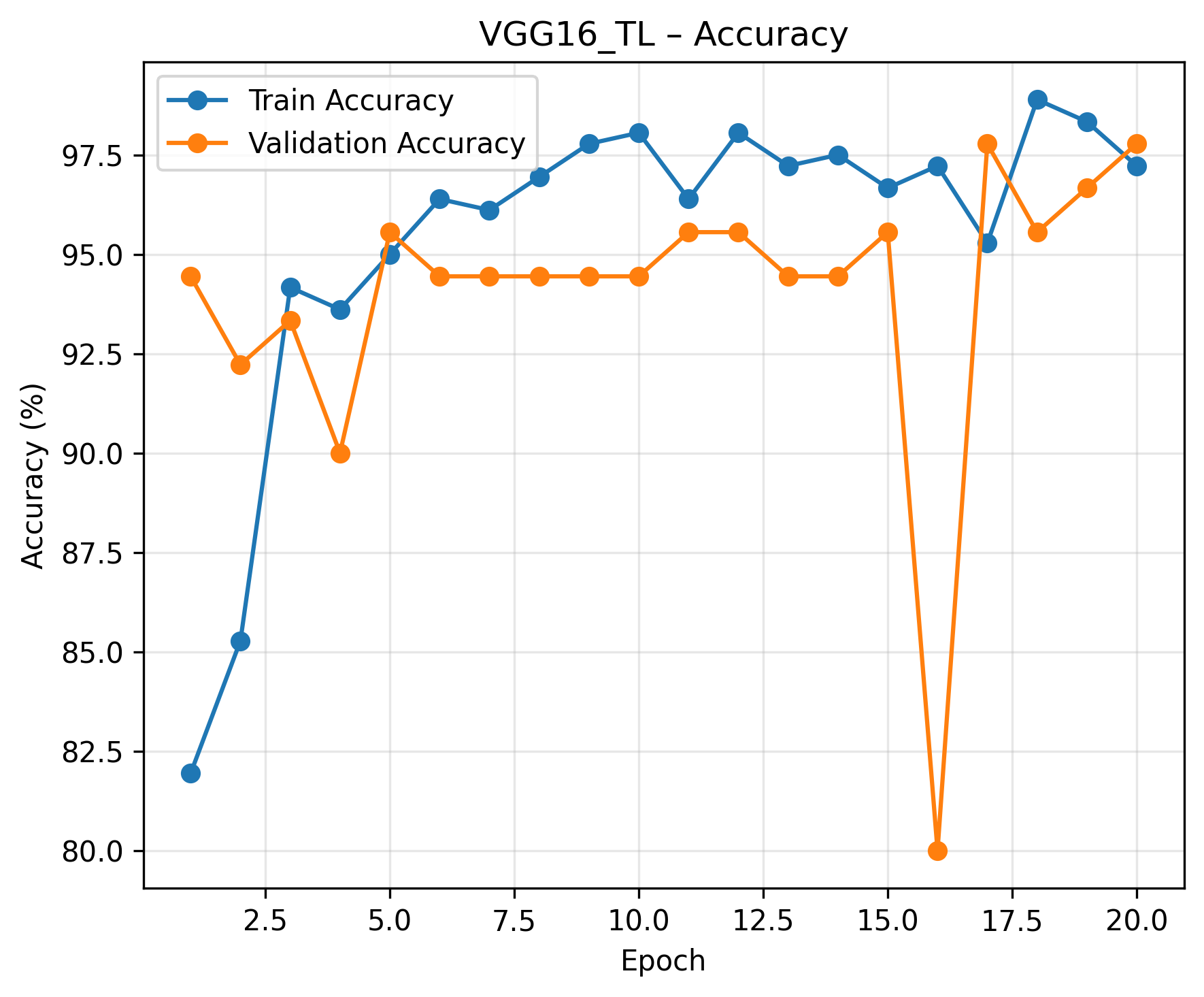}
    \caption{Training vs Validation Accuracy for VGG16 with Transfer Learning.}
    \label{fig:road_vggtl}
\end{figure}

\vspace{5pt}
Figure~\ref{fig:road_vggtl} illustrates the performance of VGG16 with transfer learning. The model demonstrates strong alignment between training and validation accuracy curves after the initial epochs, with minimal divergence, indicating consistent performance during training.
\vspace{5pt}
The figures show that all models are able to learn effectively on the road dataset, with transfer learning models exhibiting faster convergence and more stable validation behavior. The Custom CNN maintains smooth training dynamics, while deeper models trained from scratch display higher variability during training.


\newpage
\subsubsection{ROC and Precision--Recall Analysis}
\begin{figure}[ht]
    \centering
    \begin{minipage}{0.48\textwidth}
        \centering
        \includegraphics[width=\textwidth]{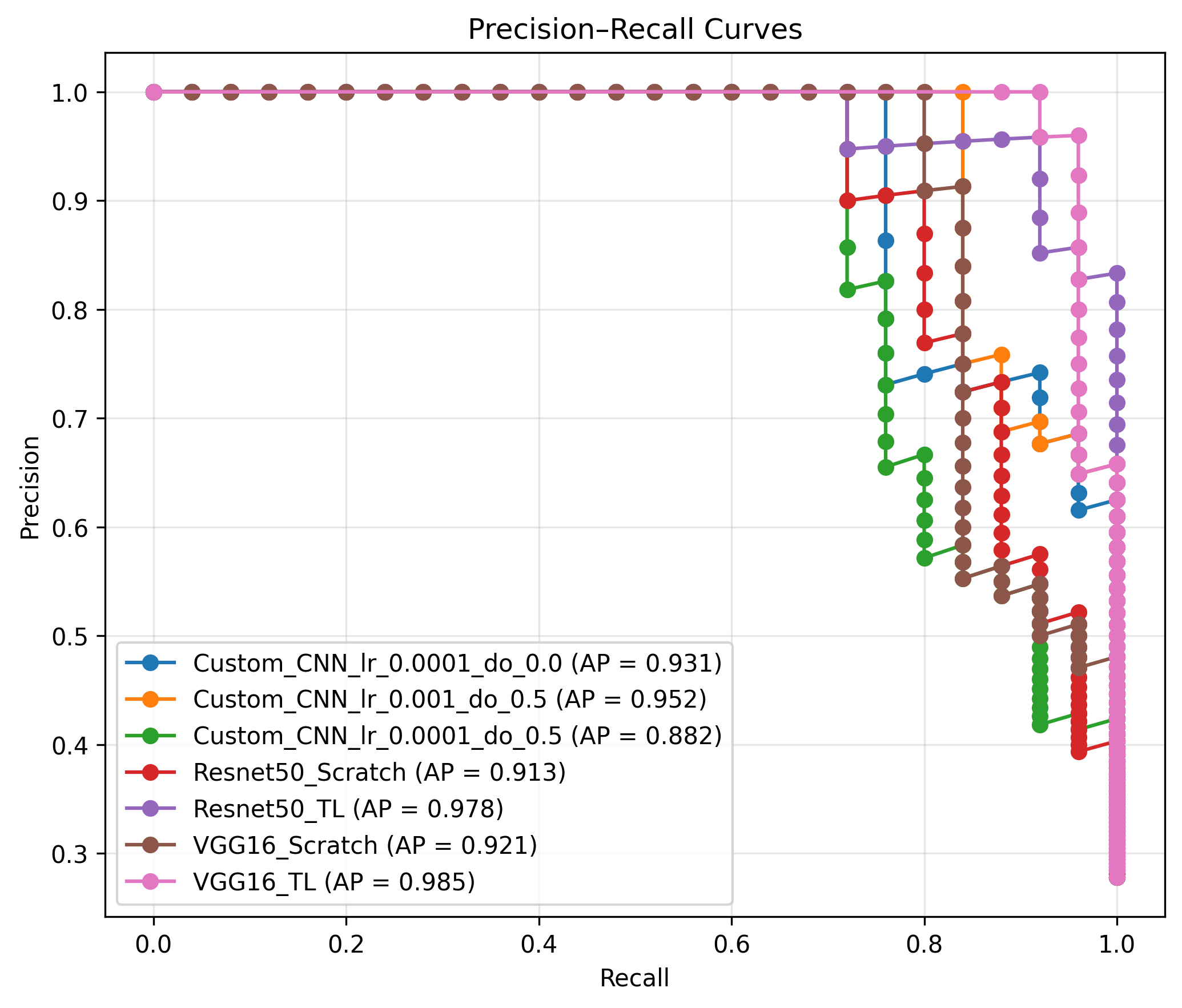}
    \end{minipage}
    \hfill
    \begin{minipage}{0.48\textwidth}
        \centering
        \includegraphics[width=\textwidth]{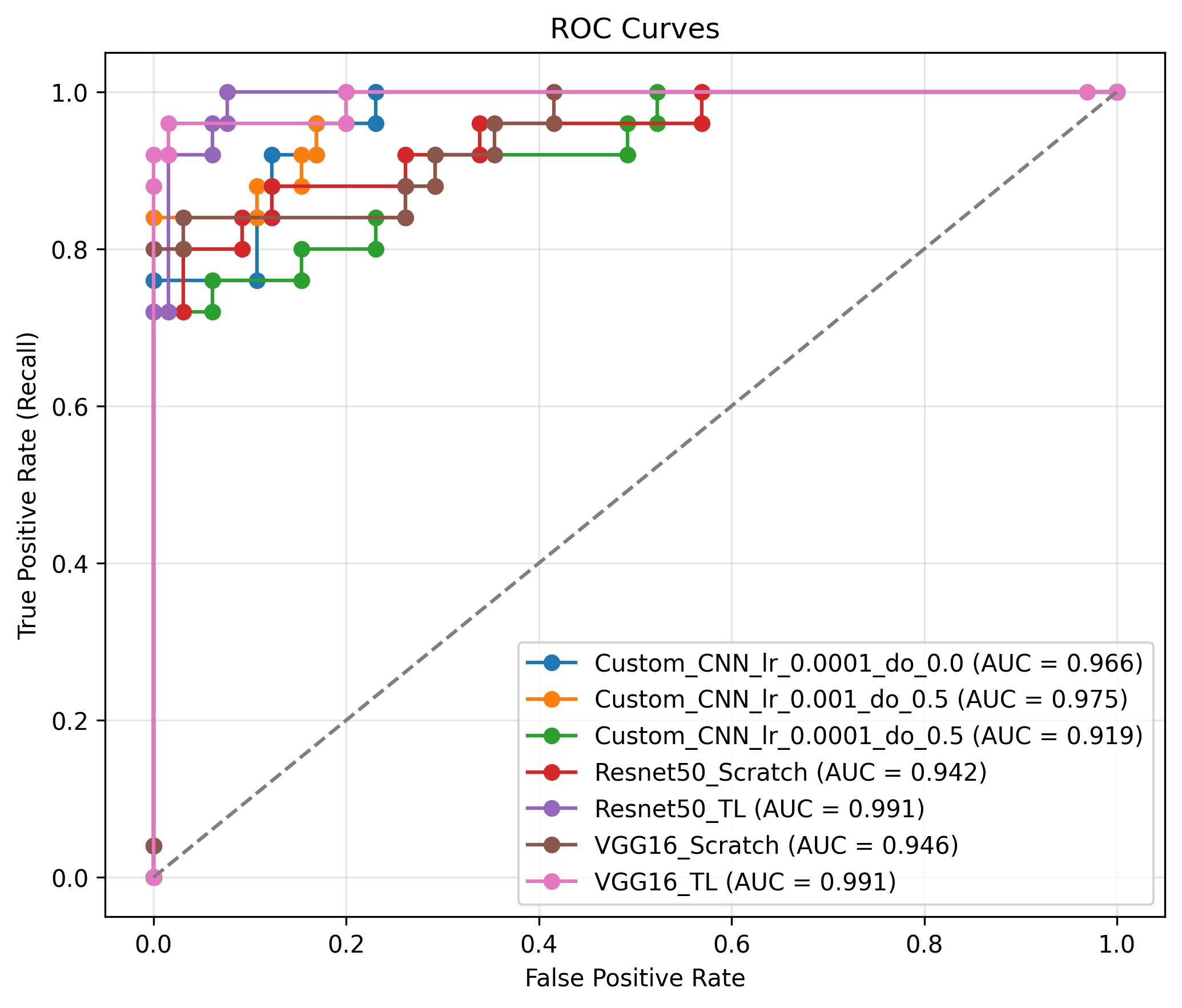}
    \end{minipage}
    \caption{(a) Precision-Recall Curves for various models on the road condition classification dataset, and (b) ROC curves comparing all models.}
    \label{fig:road_pr_roc_curves}
\end{figure}

Figure~\ref{fig:road_pr_roc_curves}(a) shows the precision–recall curves for all evaluated models on the road condition classification dataset. The curves indicate that most models maintain high precision across a wide range of recall values, with transfer learning–based models exhibiting more consistent precision at higher recall levels. The Custom CNN configurations also demonstrate strong precision–recall behavior, particularly in the mid-to-high recall region, with minor variations across different hyperparameter settings.

\vspace{5pt}
Figure~\ref{fig:road_pr_roc_curves}(b) presents the corresponding ROC curves. All models achieve curves that lie well above the diagonal baseline, indicating effective discrimination between the two classes. The transfer learning models produce curves that remain closer to the top-left corner, reflecting strong true positive rates at low false positive rates. The Custom CNN models also achieve high true positive rates, with slightly more variation in the lower false positive region compared to pretrained architectures.

Overall, the Precision–Recall and ROC curves confirm that all evaluated models perform well on the road dataset, with transfer learning models showing more consistent curve shapes, while the Custom CNN maintains competitive classification behavior across both evaluation plots.

\subsection{Multiclass Classification: MangoImageBD Dataset}
The MangoImageBD dataset \cite{FERDAUS2025111908} is a large-scale, high-quality image collection containing 28,515 images of 15 popular mango varieties from Bangladesh, including Amrapali, Ashshina Classic, Ashshina Zhinuk, Banana Mango, Bari-4, Bari-11, Fazli Classic, Fazli Shurmai, Gourmoti, Harivanga, Himsagor, Katimon, Langra, Rupali, and Shada. The mango samples were sourced from major mango-producing districts and photographed in a controlled environment using a high-resolution smartphone camera to ensure consistent lighting and background. The dataset consists of three parts: 5703 original images, 5703 processed images with cleaned virtual backgrounds, and 17,109 augmented images created using transformations such as rotation, flipping, shearing, brightness and exposure variation, blurring, and noise addition. All images are standardized to a resolution of 504×1120 px, making the dataset suitable for computer vision tasks such as classification, detection, and segmentation.

\subsubsection{Overall Performance}
Table~\ref{tab:mango-class} presents the performance comparison of all evaluated models on the MangoImageBD dataset. All models achieve high accuracy and balanced precision–recall values, indicating strong separability between classes. Among the evaluated approaches, the VGG16 model trained from scratch attains the highest overall performance across accuracy and F1-score, followed closely by ResNet50 trained from scratch.

\vspace{5pt}
The Custom CNN models consistently deliver strong results with only a marginal reduction in performance compared to deeper architectures, while maintaining significantly smaller model sizes and lower parameter counts. Across different hyperparameter settings, the Custom CNN shows stable behavior in precision, recall, and F1-score, with comparable training times.

\vspace{5pt}
Transfer learning variants of ResNet50 and VGG16 reduce training time relative to their scratch-trained counterparts but exhibit slightly lower performance metrics on this dataset. Overall, the results indicate that while deeper networks achieve the strongest classification performance, the Custom CNN provides an efficient alternative with competitive accuracy and substantially lower computational overhead.

\begin{table}[ht]
\centering
\caption{Model wise performance metrics for MangoImageBD}
\label{tab:mango-class}
\footnotesize
\begin{tabular}{@{}p{2.2cm}cccccccc@{}}
\toprule
Model & Acc. & Prec. & Recall & F1 & Time (s) & Total Par. & Train. Par. & Size (MB) \\
\midrule
Custom CNN (lr: 0.0001, do: 0.0) & 95.38 & 0.9544 & 0.9538 & 0.9532 & 8558.01 & 642981 & 642981 & 2.45 \\
Custom CNN (lr: 0.001, do: 0.5) & 94.66 & 0.9493 & 0.9466 & 0.9442 & 8703.03 & 642981 & 642981 & 2.45 \\
Custom CNN (lr: 0.0001, do: 0.5) & 93.68 & 0.9377 & 0.9368 & 0.9360 & 8767.69 & 642981 & 642981 & 2.45 \\
ResNet50 (Scratch) & 97.39 & 0.9739 & 0.9739 & 0.9735 & 9297.13 & 23538767 & 23538767 & 89.79 \\
ResNet50 (TL) & 96.00 & 0.9606 & 0.9600 & 0.9595 & 6267.95 & 23538767 & 30735 & 89.79 \\
VGG16 (Scratch) & 97.60 & 0.9760 & 0.9760 & 0.9759 & 11877.57 & 134321999 & 134321999 & 512.40 \\
VGG16 (TL) & 96.82 & 0.9691 & 0.9682 & 0.9682 & 7839.66 & 134321999 & 61455 & 512.40 \\
\bottomrule
\end{tabular}
\end{table}

\subsubsection{Model Comparison and Accuracy Curves}

\begin{figure}[ht]
    \centering
    \begin{minipage}{0.48\textwidth}
        \centering
        \includegraphics[width=\textwidth]{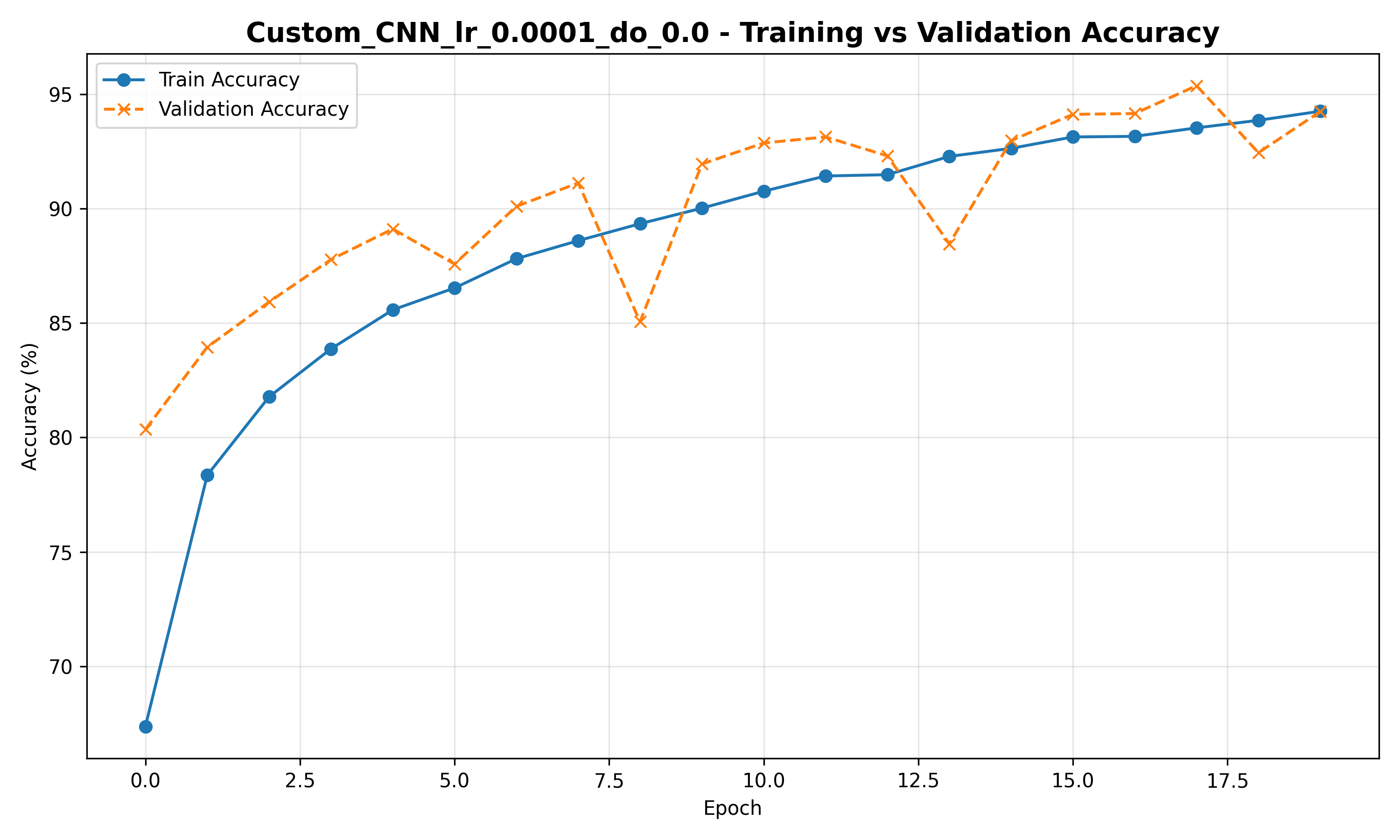}
    \end{minipage}
    \hfill
    \begin{minipage}{0.48\textwidth}
        \centering
        \includegraphics[width=\textwidth]{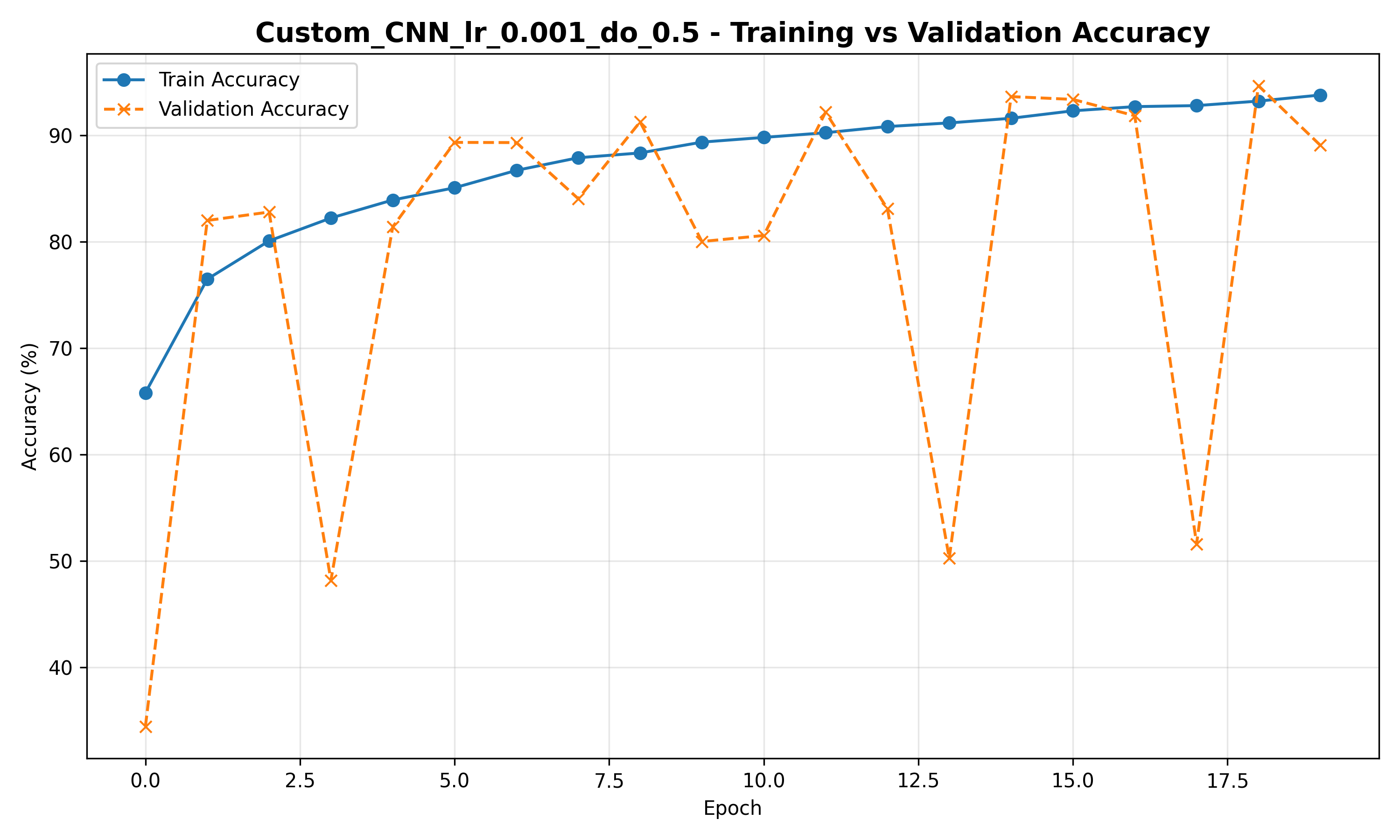}
    \end{minipage}
    \caption{Training vs Validation Accuracy for Custom CNN showing (a) Learning Rate: 0.0001, Dropout: 0.0 and (b) Learning Rate: 0.001, Dropout: 0.5.}
    \label{fig:mango_custom-cnn-1-2}
\end{figure}

Figures~\ref{fig:mango_custom-cnn-1-2} illustrate the training and validation accuracy trends of the Custom CNN under two different hyperparameter settings. Both configurations show a steady improvement in training accuracy over epochs, with validation accuracy closely tracking the training curve after the initial epochs. Minor fluctuations are visible in early stages, after which the curves stabilize, indicating consistent learning behavior across both settings.

\begin{figure}[ht]
    \centering
    \begin{minipage}{0.48\textwidth}
        \centering
        \includegraphics[width=\textwidth]{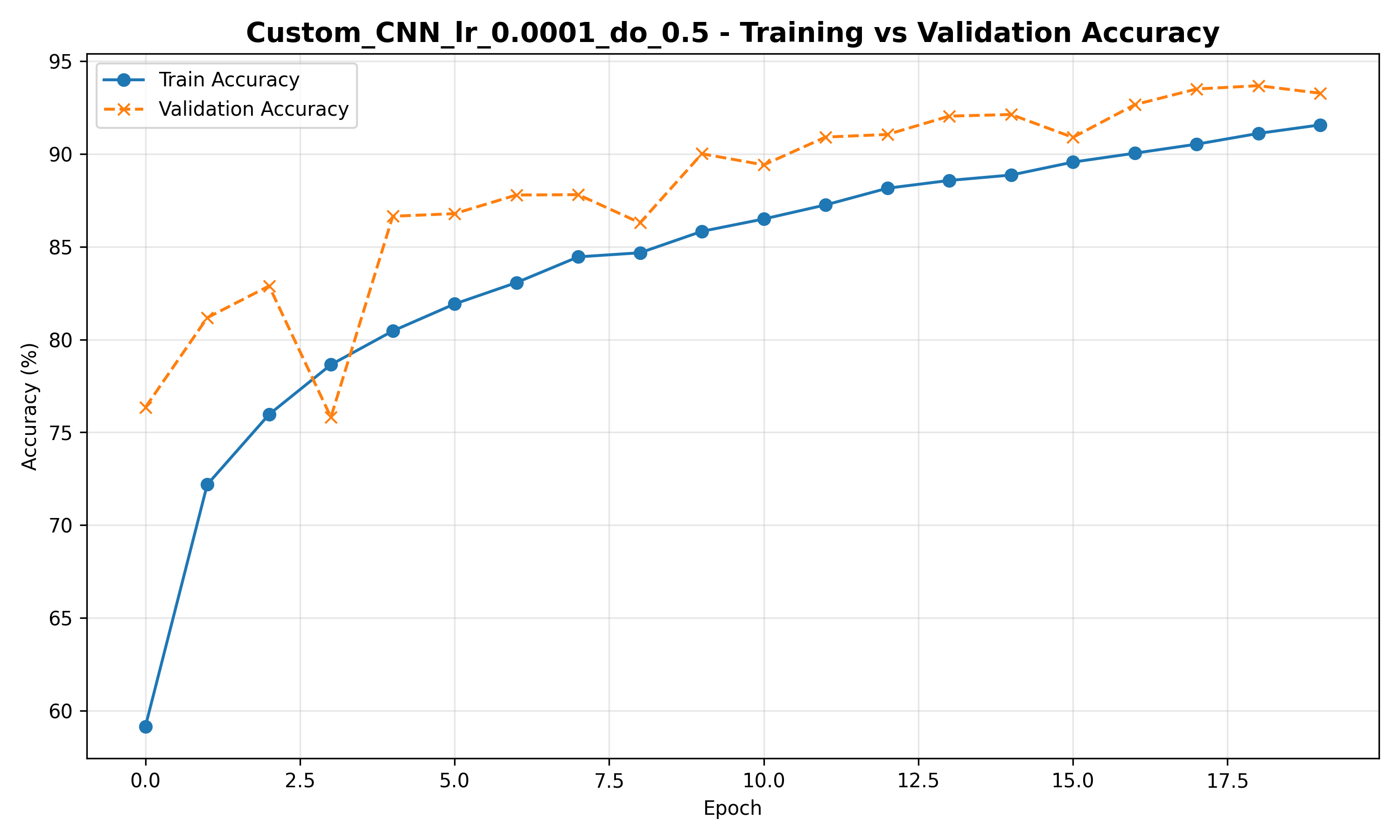}
    \end{minipage}
    \hfill
    \begin{minipage}{0.48\textwidth}
        \centering
        \includegraphics[width=\textwidth]{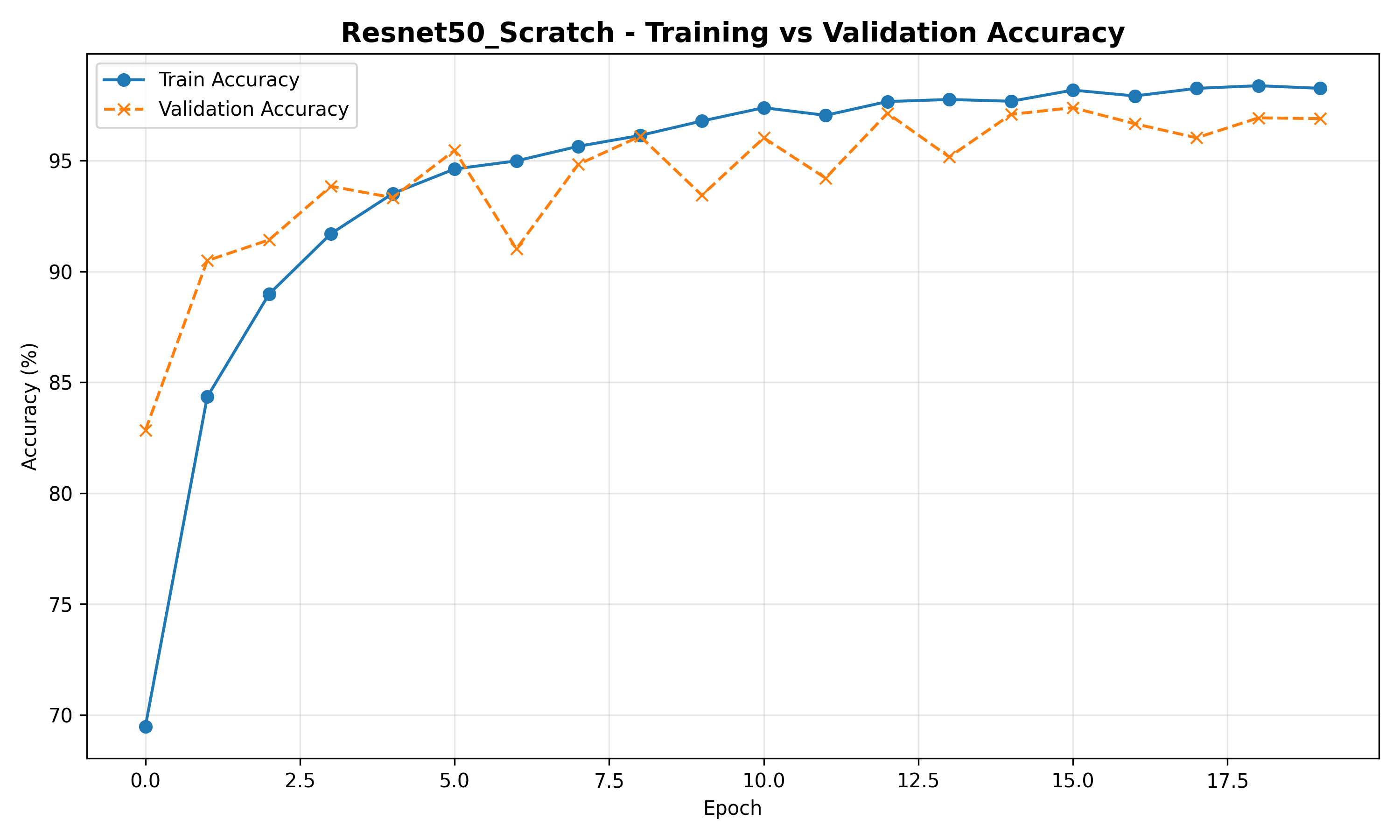}
    \end{minipage}
    \caption{Training vs Validation Accuracy showing (a) Custom CNN (Learning Rate: 0.0001, Dropout: 0.5) and (b) ResNet50 from scratch.}
    \label{fig:mango_custom-cnn-3-resnet-scratch}
\end{figure}

\vspace{5pt}
Figure~\ref{fig:mango_custom-cnn-3-resnet-scratch} presents a comparison between the Custom CNN (with learning rate 0.0001 and dropout 0.5) and ResNet50 trained from scratch. The Custom CNN exhibits smoother convergence with relatively stable validation accuracy, while the ResNet50 from-scratch model shows a faster rise in training accuracy accompanied by noticeable oscillations in validation performance across epochs.

\begin{figure}[ht]
    \centering
    \begin{minipage}{0.48\textwidth}
        \centering
        \includegraphics[width=\textwidth]{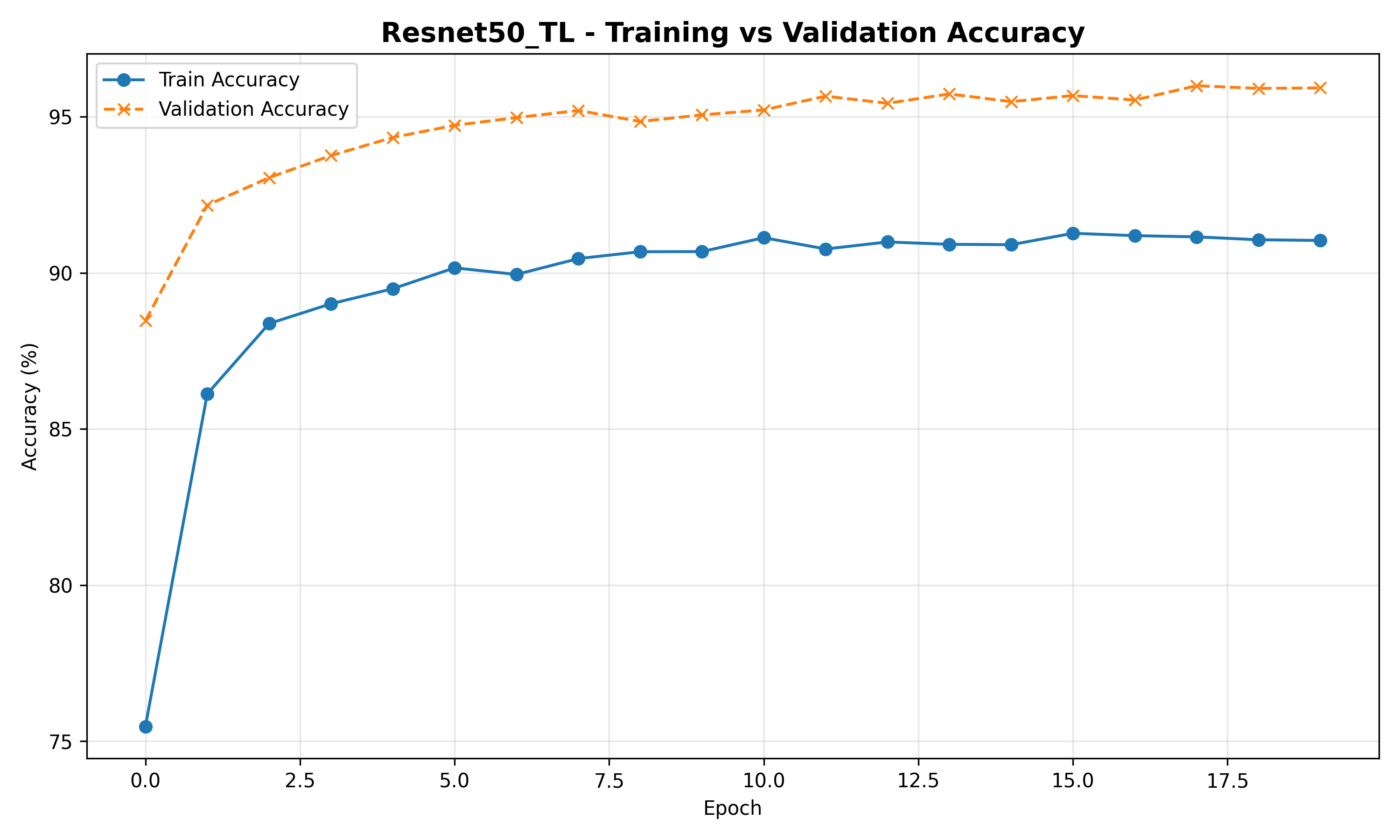}
    \end{minipage}
    \hfill
    \begin{minipage}{0.48\textwidth}
        \centering
        \includegraphics[width=\textwidth]{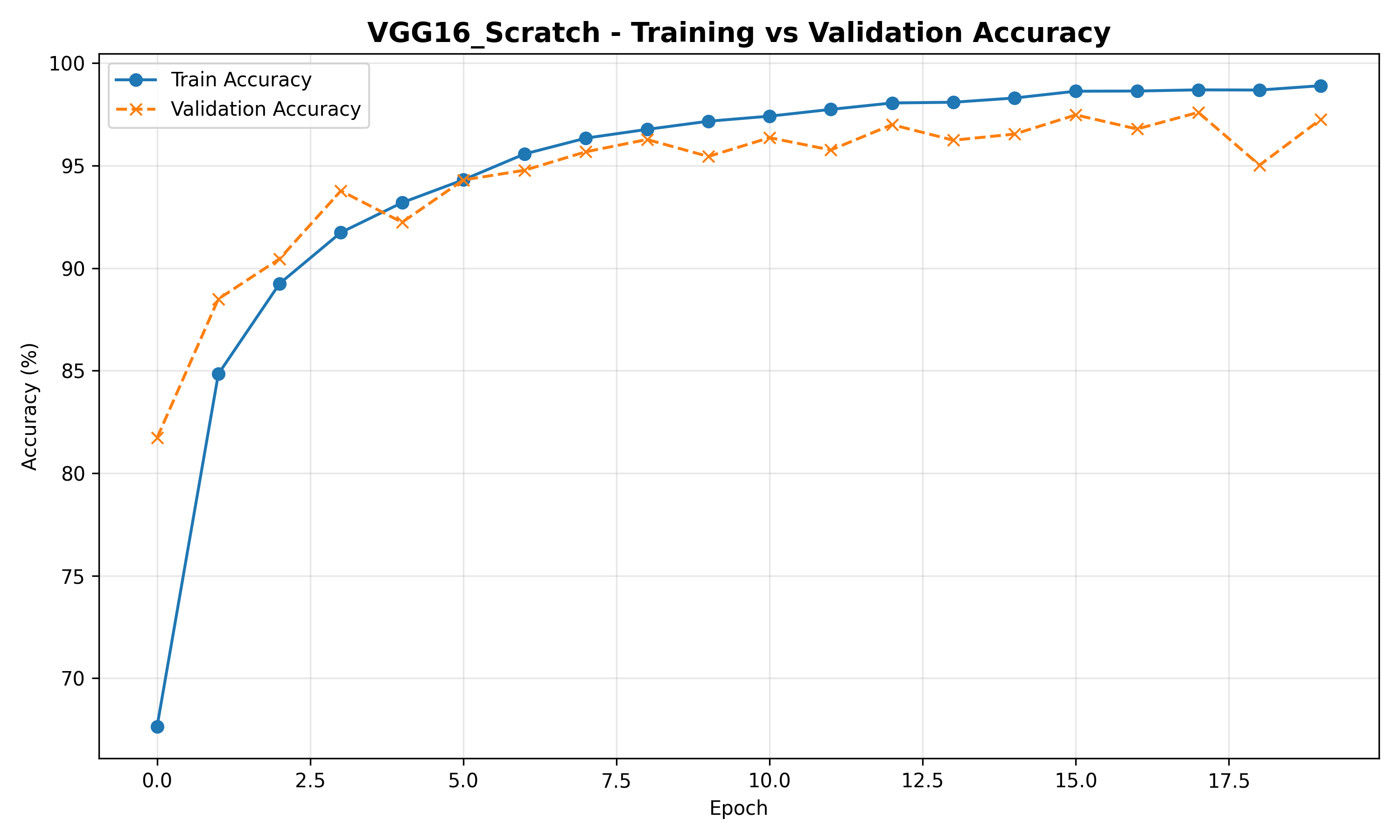}
    \end{minipage}
    \caption{Training vs Validation Accuracy showing (a) ResNet50 with Transfer Learning and (b) VGG16 from scratch.}
    \label{fig:mango_resnet-tl-vgg-scratch}
\end{figure}

\vspace{5pt}
In Figure~\ref{fig:mango_resnet-tl-vgg-scratch}, the accuracy curves for ResNet50 with transfer learning and VGG16 trained from scratch are shown. The transfer learning model demonstrates rapid convergence and maintains a narrow gap between training and validation accuracy throughout training. In contrast, the VGG16 scratch model shows a sharp increase in training accuracy, with validation accuracy following a similar trend but with slightly higher variability.

\begin{figure}[ht]
    \centering
    \includegraphics[width=0.48\textwidth]{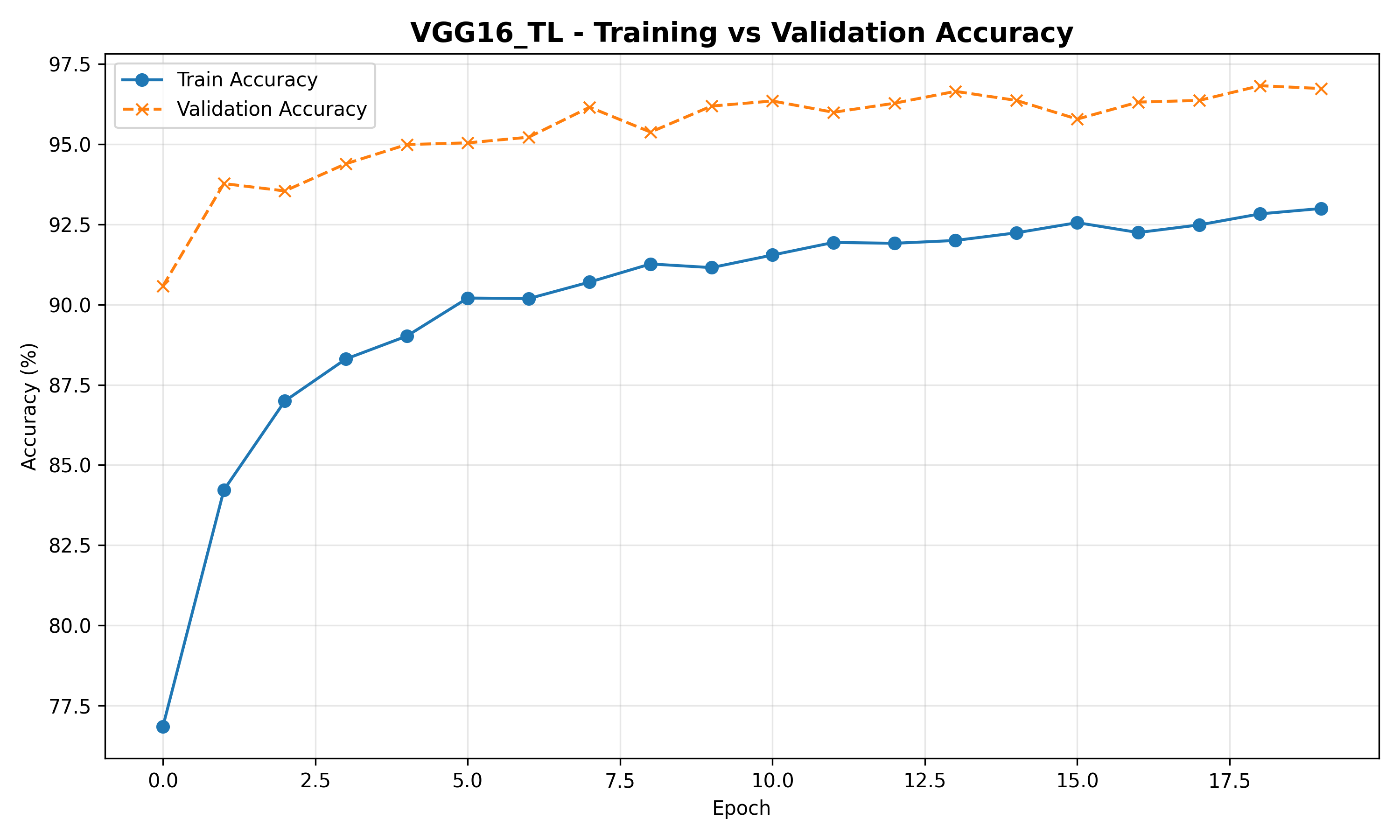}
    \caption{Training vs Validation Accuracy for VGG16 with Transfer Learning.}
    \label{fig:mango_vgg16-tl}
\end{figure}

\vspace{5pt}
Finally, Figure~\ref{fig:mango_vgg16-tl} depicts the training and validation accuracy of VGG16 with transfer learning. The validation accuracy remains consistently high across epochs, with a clear separation from the training curve. This figure highlights stable convergence behavior and sustained performance during training.

\newpage
\subsection{Multiclass Classification: PaddyVarietyBD Dataset}

The PaddyVarietyBD dataset \cite{TAHSIN2025111514} contains images from multiple paddy varieties cultivated across Bangladesh, making it one of the most diverse datasets in this study. The Custom\_CNN model was evaluated on this dataset to assess its capability to discriminate between visually similar crop types under varying lighting, texture, and morphological conditions. The following results highlight the model’s learning dynamics, classification behaviour, and overall performance.

\subsubsection{Overall Performance}

Table~\ref{tab:paddy-class} summarizes the performance of all evaluated models on the PaddyVarietyBD dataset across multiple evaluation metrics. Among the Custom CNN configurations, the model trained with a learning rate of 0.001 and dropout of 0.5 achieves the strongest overall performance, showing noticeable improvements in accuracy, precision, recall, and F1-score compared to the other custom variants. The remaining Custom CNN settings yield comparable but lower results, indicating sensitivity to hyperparameter choices for this dataset.

\vspace{5pt}
The models trained from scratch demonstrate a clear performance advantage. ResNet50 trained from scratch achieves the highest accuracy and balanced metric values, closely followed by VGG16 from scratch. These models consistently outperform their transfer learning counterparts, reflecting stronger discriminative capability when fully trained on the paddy-specific data.

\vspace{5pt}
In contrast, both ResNet50 and VGG16 with transfer learning exhibit substantially lower performance across all metrics. Despite reduced training time, their accuracy, recall, and F1-scores lag behind not only the scratch-trained models but also the best Custom CNN configuration. This contrast highlights a distinct divergence in behavior compared to other datasets.

\vspace{5pt}
From a resource perspective, the Custom CNN remains the most lightweight model with minimal parameters and model size, while VGG16-based models incur significantly higher computational and memory costs. Overall, the table highlights a clear distinction between scratch-trained and transfer learning models for this dataset, with strong performance gains achieved at the cost of increased model complexity and training time.

\begin{table}[ht]
\centering
\caption{Model performance metrics for PaddyVarietyBD}
\label{tab:paddy-class}
\footnotesize
\begin{tabular}{@{}p{2.2cm}cccccccc@{}}
\toprule
Model & Acc. & Prec. & Recall & F1 & Time (s) & Total Par. & Train. Par. & Size (MB) \\
\midrule
Custom CNN (lr: 0.0001, do: 0.0) & 81.64 & 0.8395 & 0.8164 & 0.8151 & 10602.07 & 645561 & 645561 & 2.46 \\
Custom CNN (lr: 0.001, do: 0.5) & 86.82 & 0.8828 & 0.8682 & 0.8689 & 10501.09 & 645561 & 645561 & 2.46 \\
Custom CNN (lr: 0.0001, do: 0.5) & 81.51 & 0.8300 & 0.8151 & 0.8142 & 11588.65 & 645561 & 645561 & 2.46 \\
ResNet50 (Scratch) & 92.23 & 0.9263 & 0.9223 & 0.9219 & 12959.42 & 23579747 & 23579747 & 89.95 \\
ResNet50 (TL) & 69.74 & 0.7029 & 0.6974 & 0.6954 & 5635.47 & 23579747 & 71715 & 89.95 \\
VGG16 (Scratch) & 89.40 & 0.8970 & 0.8940 & 0.8938 & 18622.88 & 134403939 & 134403939 & 512.71 \\
VGG16 (TL) & 67.67 & 0.6967 & 0.6767 & 0.6777 & 8559.95 & 134403939 & 143395 & 512.71 \\
\bottomrule
\end{tabular}
\end{table}

\subsubsection{Model Comparison and Accuracy Curves}

\begin{figure}[ht]
    \centering
    \begin{minipage}{0.48\textwidth}
        \centering
        \includegraphics[width=\textwidth]{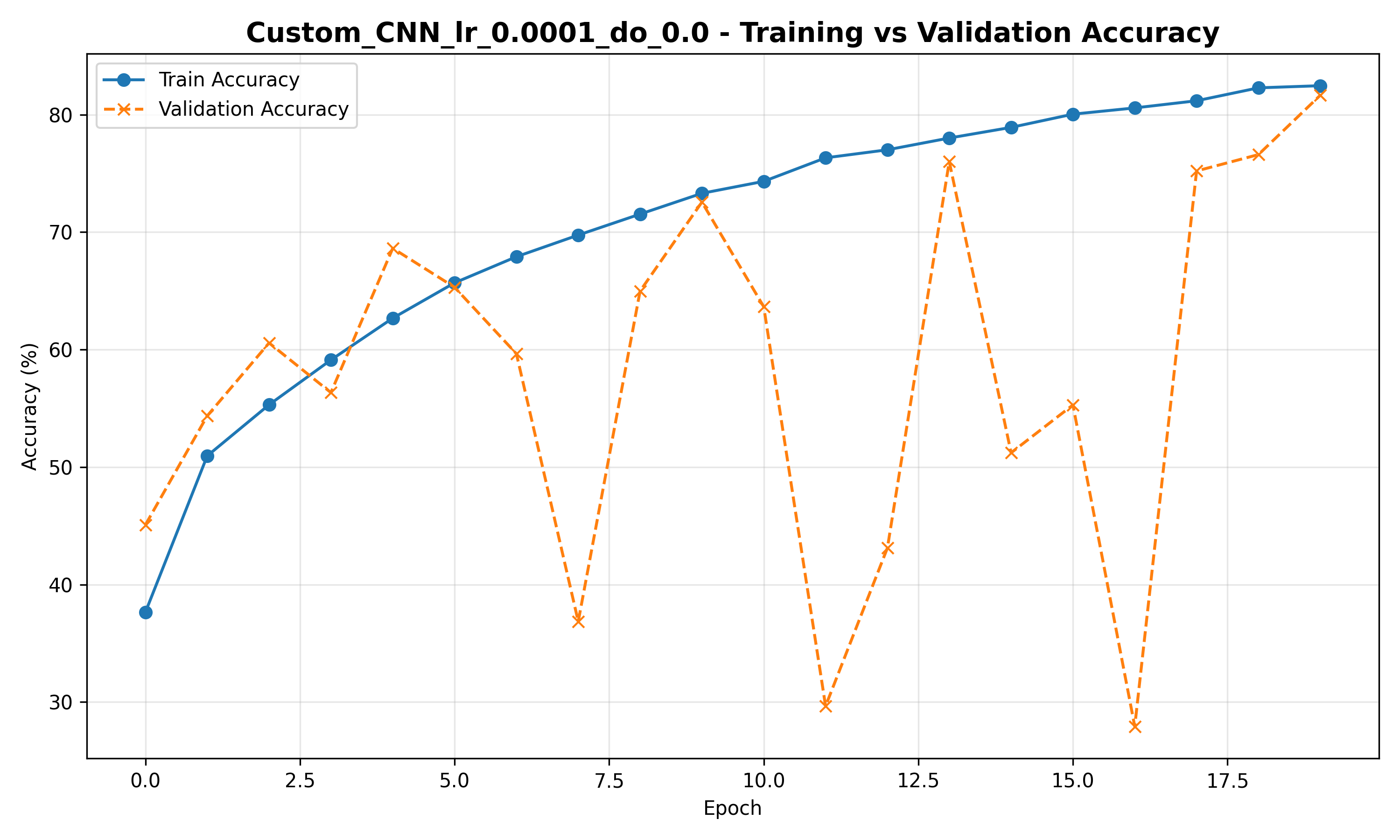}
    \end{minipage}
    \hfill
    \begin{minipage}{0.48\textwidth}
        \centering
        \includegraphics[width=\textwidth]{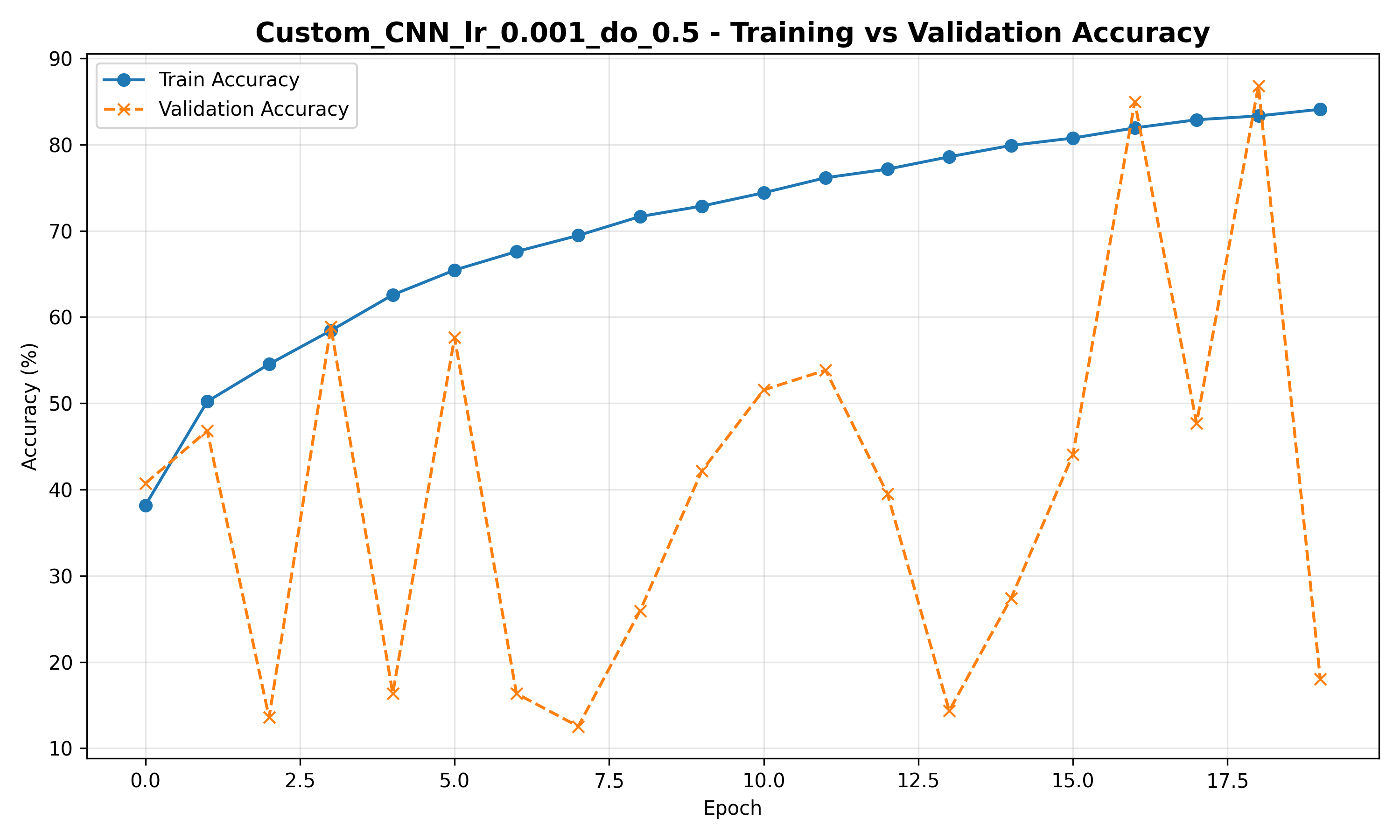}
    \end{minipage}
    \caption{Training vs Validation Accuracy for Custom CNN showing (a) Learning Rate: 0.0001, Dropout: 0.0 and (b) Learning Rate: 0.001, Dropout: 0.5.}
    \label{fig:paddy_custom-cnn-1-2}
\end{figure}

Figures \ref{fig:paddy_custom-cnn-1-2} illustrate the training and validation accuracy trends of the Custom CNN under two different hyperparameter settings. In both cases, training accuracy increases steadily across epochs, indicating stable learning. However, the validation accuracy exhibits noticeable fluctuations, particularly for the configuration with higher learning rate and dropout, reflecting unstable generalization across epochs. The configuration with a lower learning rate and no dropout shows comparatively smoother validation trends, although the overall validation performance remains moderate.

\begin{figure}[ht]
    \centering
    \begin{minipage}{0.48\textwidth}
        \centering
        \includegraphics[width=\textwidth]{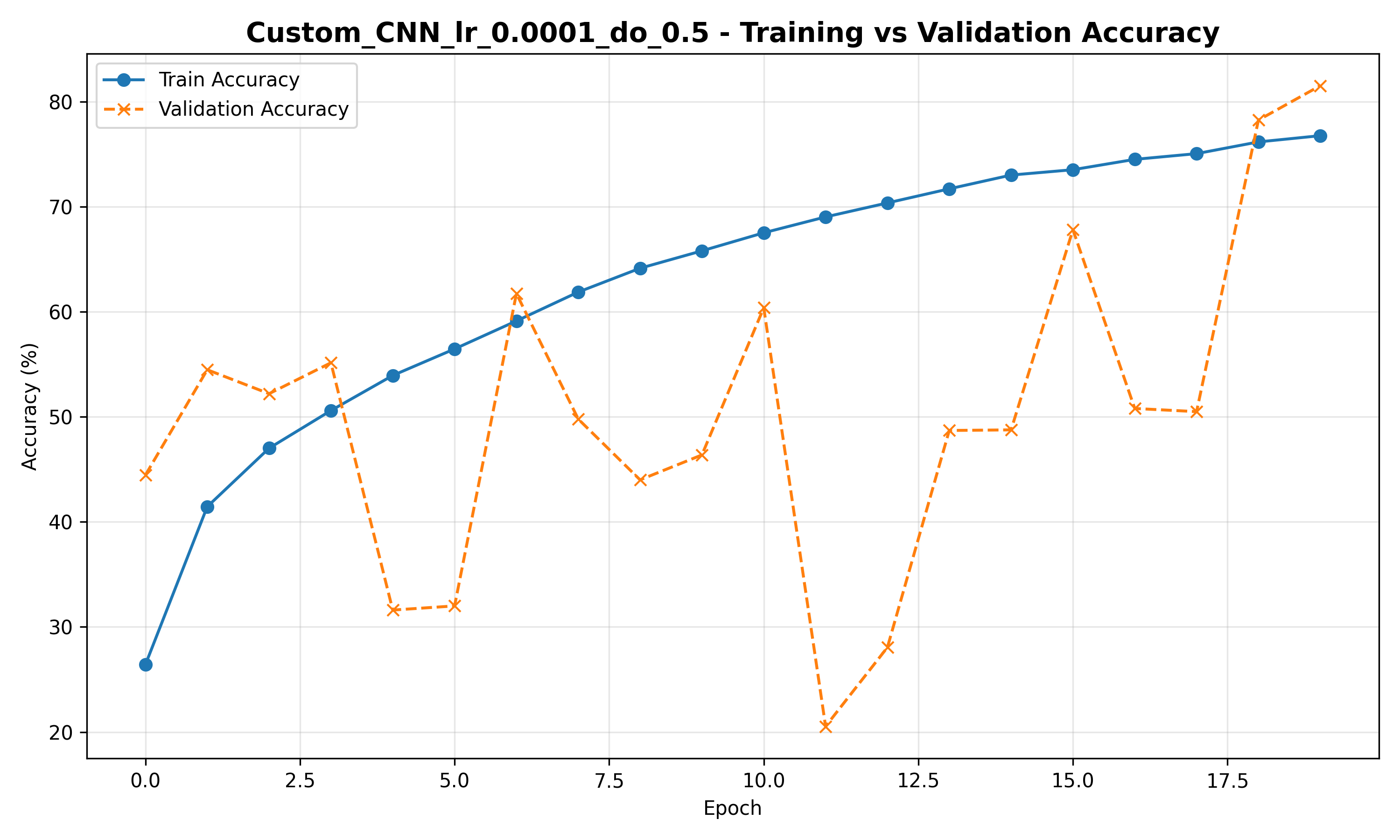}
    \end{minipage}
    \hfill
    \begin{minipage}{0.48\textwidth}
        \centering
        \includegraphics[width=\textwidth]{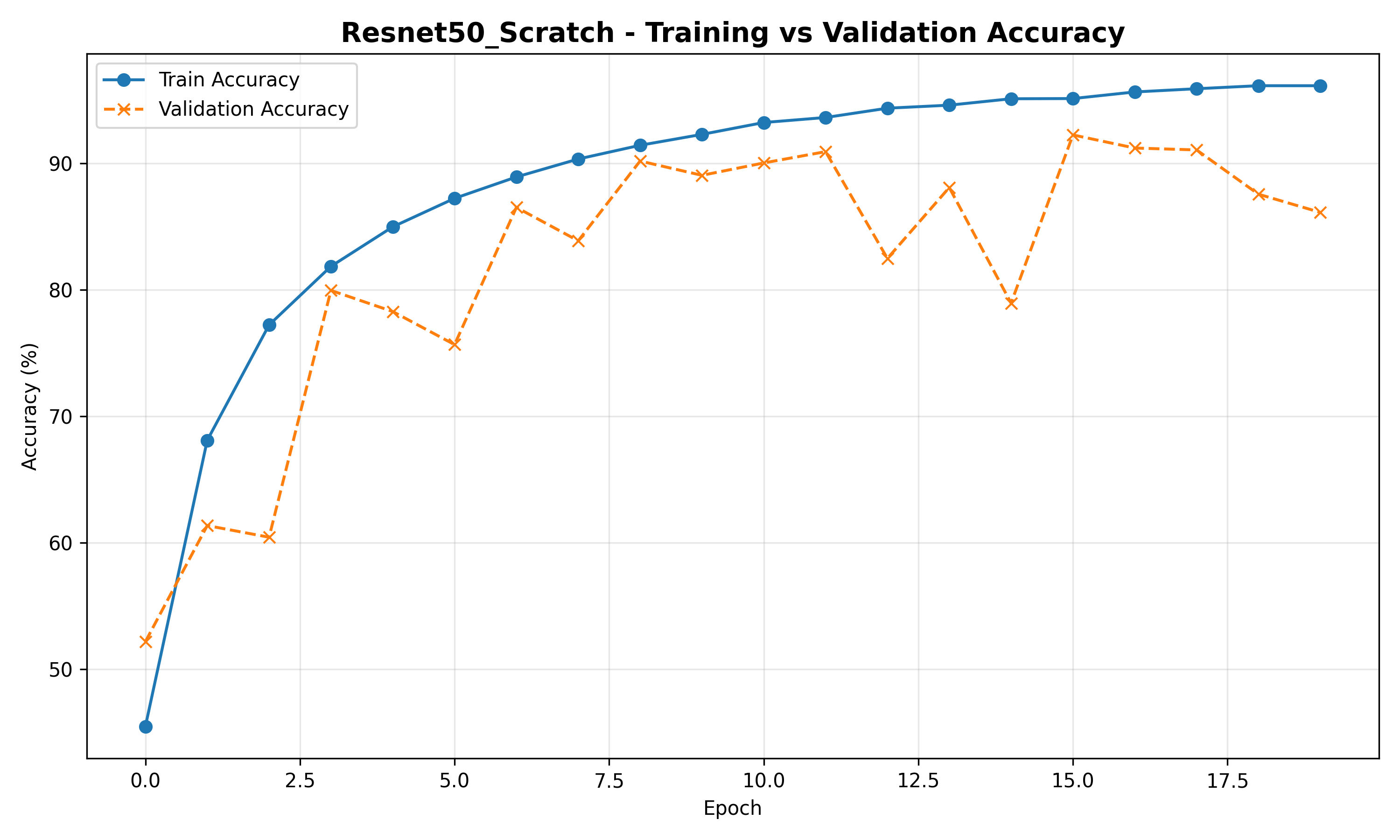}
    \end{minipage}
    \caption{Training vs Validation Accuracy showing (a) Custom CNN (Learning Rate: 0.0001, Dropout: 0.5) and (b) ResNet50 from scratch.}
    \label{fig:paddy_custom-cnn-3-resnet-scratch}
\end{figure}

\vspace{5pt}
Figure \ref{fig:paddy_custom-cnn-3-resnet-scratch} compares the Custom CNN with ResNet50 trained from scratch. While the Custom CNN demonstrates gradual improvement with fluctuating validation accuracy, ResNet50 from scratch achieves a rapid increase in training accuracy and maintains relatively higher and more consistent validation accuracy. This contrast highlights the stronger fitting capability of the deeper architecture when trained directly on the dataset.

\begin{figure}[ht]
    \centering
    \begin{minipage}{0.48\textwidth}
        \centering
        \includegraphics[width=\textwidth]{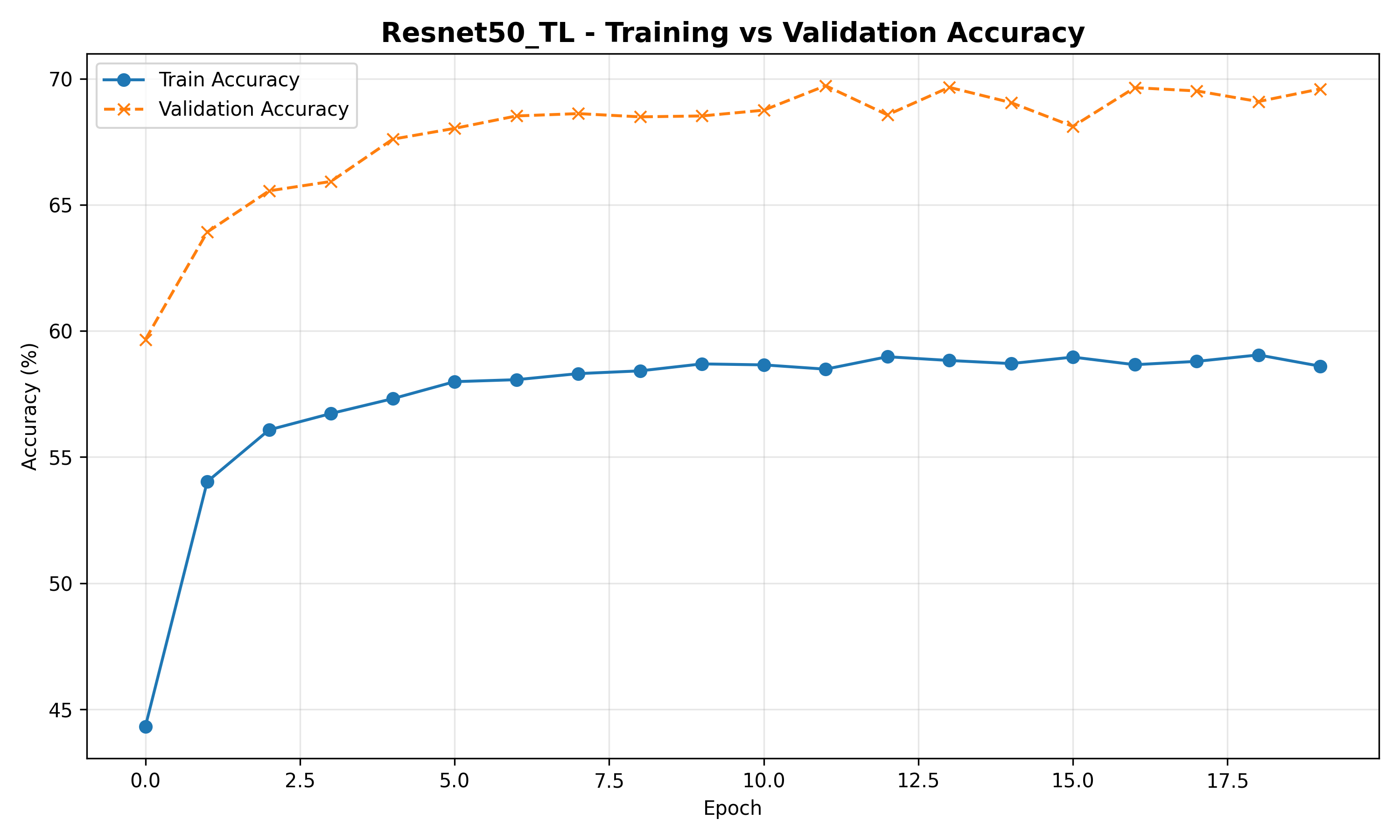}
    \end{minipage}
    \hfill
    \begin{minipage}{0.48\textwidth}
        \centering
        \includegraphics[width=\textwidth]{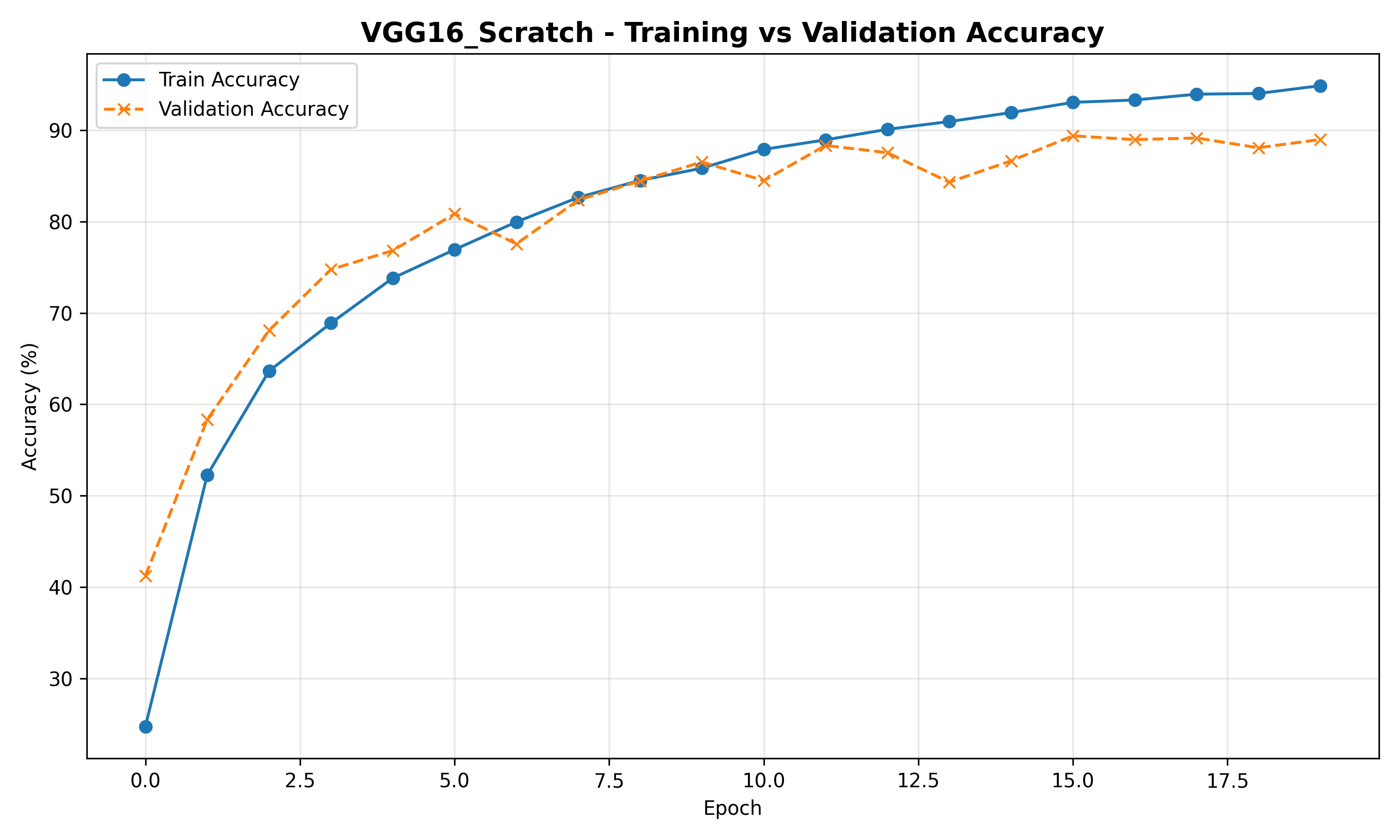}
    \end{minipage}
    \caption{Training vs Validation Accuracy showing (a) ResNet50 with Transfer Learning and (b) VGG16 from scratch.}
    \label{fig:paddy_resnet-tl-vgg-scratch}
\end{figure}

\vspace{5pt}
Figure \ref{fig:paddy_resnet-tl-vgg-scratch} presents the accuracy curves for ResNet50 with transfer learning and VGG16 trained from scratch. The ResNet50 transfer learning model shows limited improvement in training accuracy and plateaus early, with validation accuracy remaining consistently low. In contrast, VGG16 trained from scratch exhibits a strong upward trend in both training and validation accuracy, with a smaller gap between the two, indicating more stable learning behavior.

\begin{figure}[ht]
    \centering
    \includegraphics[width=0.48\textwidth]{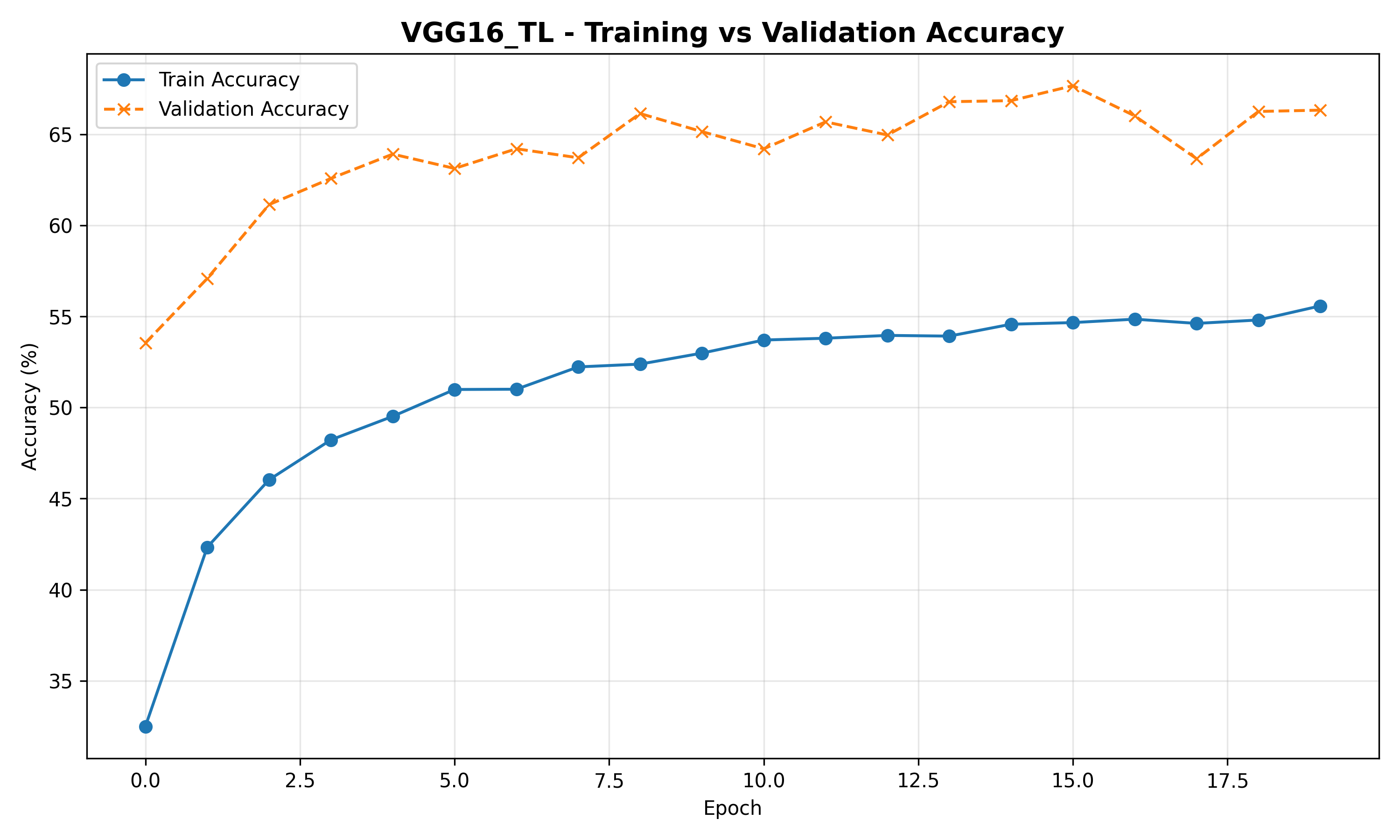}
    \caption{Training vs Validation Accuracy for VGG16 with Transfer Learning.}
    \label{fig:paddy_vgg16-tl}
\end{figure}

\vspace{5pt}
Finally, Figure \ref{fig:paddy_vgg16-tl} shows the training and validation accuracy for VGG16 with transfer learning. The training accuracy increases slowly and remains significantly lower than the scratch-trained counterpart, while validation accuracy fluctuates within a narrow range and does not show sustained improvement across epochs. Overall, the figures demonstrate clear differences in convergence behavior and stability across models and training strategies on the PaddyVarietyBD dataset, with scratch-trained deep models showing stronger and more consistent accuracy trends compared to their transfer learning counterparts.

\newpage
\section{Discussion}\label{sec12}

This section discusses the experimental findings across the five datasets, focusing on (i) the best-performing model for each dataset, (ii) the underlying reasons for its performance, (iii) the behavior of the proposed Custom CNN, and (iv) the associated performance–efficiency trade-offs. The discussion emphasizes architectural suitability, dataset characteristics, and practical deployment considerations.

\subsection{Binary Classification: Auto-RickshawImageBD Dataset}

For the Autorickshaw dataset, ResNet50 with transfer learning is the best-performing model, primarily due to its ability to leverage pretrained ImageNet features that generalize well to vehicle-centric visual patterns. These pretrained representations enable stronger class separation, which is reflected in higher accuracy, F1-score, and AUC values. The model’s strong performance across both threshold-dependent and threshold-independent metrics indicates robust generalization rather than overfitting.

\vspace{5pt}
The Custom CNN demonstrates competitive performance when compared to the deep models trained from scratch, particularly in recall. This suggests that the Custom CNN is effective at identifying autorickshaw instances, even if it occasionally misclassifies non-autorickshaw samples. The relatively higher recall can be attributed to its focused architecture, which is optimized for learning dataset-specific patterns rather than generic visual abstractions. Residual connections improve gradient flow, while squeeze-and-excitation blocks help emphasize informative feature channels, contributing to stable training and consistent performance.

\vspace{5pt}
However, the Custom CNN exhibits lower precision and overall accuracy compared to the pretrained ResNet50. This indicates a reduced ability to suppress false positives, which is likely a consequence of its limited depth and representational capacity. While the architecture captures dominant visual cues effectively, it struggles to model finer inter-class variations, which are better handled by deeper pretrained networks.

\vspace{5pt}
In comparison, ResNet50 and VGG16 trained from scratch underperform relative to both the pretrained models and, in some cases, the Custom CNN. Their lower F1-scores and AUC values suggest weaker generalization, which can be attributed to random initialization and the limited size of the dataset. Deep architectures with large parameter counts require substantially more data to learn robust feature hierarchies; without pretrained weights, they tend to overfit early or converge to suboptimal solutions. This is further reflected in longer training times without corresponding gains in performance.
\subsection{Binary Classification: Footpath Dataset}

For the Footpath dataset, ResNet50 with transfer learning is the strongest overall model, achieving the best balance across accuracy, F1-score, ROC–AUC, and precision–recall performance. The task involves recognizing subtle spatial patterns such as object placement, boundary violations, and scene-level context. Pretrained ResNet50 features are particularly effective in capturing these mid- to high-level spatial cues, allowing the model to distinguish encroached and non-encroached regions more reliably. The strong ROC–AUC and average precision values indicate robust class separation across decision thresholds.

\vspace{5pt}
The Custom CNN performs competitively, especially in precision and recall balance, showing that it is capable of learning the dominant structural patterns present in footpath scenes. Its relatively strong recall suggests effective detection of encroached areas, which is critical for urban monitoring applications. However, fluctuations in validation accuracy across epochs indicate sensitivity to hyperparameter choices, reflecting the model’s limited capacity to consistently capture complex background variations and occlusions. This limitation becomes more apparent when compared to pretrained models that have learned richer spatial hierarchies from large-scale datasets.

\vspace{5pt}
Models trained from scratch, particularly VGG16 and ResNet50 without transfer learning, show weaker generalization. While they eventually reach moderate accuracy, their lower AUC and average precision suggest difficulty in maintaining consistent performance across thresholds. This can be attributed to the high intra-class variability of footpath scenes combined with insufficient data to train deep networks effectively from random initialization. Overall, this dataset emphasizes the importance of pretrained spatial representations, while also demonstrating that the Custom CNN remains a viable option when computational efficiency is prioritized.

\subsection{Binary Classification: Road Damage Dataset}

In Dataset 3, VGG16 with transfer learning emerges as the best-performing model, particularly excelling in precision, F1-score, and average precision. This dataset benefits from VGG16’s deep but uniform convolutional structure, which is effective at capturing texture-rich and scene-level visual patterns common in road and urban environments. Transfer learning enables the model to reuse low-level texture detectors and mid-level compositional features, resulting in strong precision and stable validation performance.

\vspace{5pt}
The Custom CNN shows stable but comparatively lower performance, particularly in precision-sensitive metrics. While recall remains reasonably high, indicating effective detection of positive instances, the reduced precision suggests increased false positives. This behavior indicates that while the Custom CNN captures prominent visual cues, it lacks the depth required to disambiguate visually similar background elements. Its performance remains consistent across epochs, highlighting good training stability, but the architectural simplicity limits its discriminative power for complex scene understanding.

\vspace{5pt}
From-scratch models again lag behind, with VGG16 and ResNet50 showing slower convergence and lower F1-scores. The absence of pretrained weights forces these models to learn both low-level and high-level features simultaneously, which is challenging given dataset complexity. Consequently, they exhibit weaker generalization and less reliable performance across thresholds. This dataset reinforces that transfer learning is especially beneficial for texture- and context-driven classification tasks, while the Custom CNN remains best suited for scenarios where interpretability and efficiency outweigh marginal gains in precision.

\subsection{Multiclass Classification: MangoImageBD Dataset}

For Dataset 4, ResNet50 with transfer learning achieves the best overall performance, particularly in recall, F1-score, and ROC–AUC. The dataset contains complex object arrangements and overlapping visual features, which benefit from the deep residual architecture of ResNet50. Transfer learning enables the model to reuse hierarchical representations that are effective at modeling object interactions and contextual dependencies, resulting in superior generalization and stable performance across all evaluation metrics.

\vspace{5pt}
The Custom CNN performs adequately but shows noticeable limitations in precision and AUC, indicating difficulty in achieving clean class separation. While recall remains relatively strong—suggesting sensitivity to positive instances—the reduced precision points to confusion between visually similar classes. This behavior reflects the trade-off inherent in compact architectures: while efficient and stable, they may struggle with datasets that require deep semantic understanding and multi-level feature abstraction.

\vspace{5pt}
ResNet50 and VGG16 trained from scratch again underperform due to insufficient data to support their depth and parameter count. Their lower AUC and inconsistent validation behavior indicate overfitting and suboptimal feature learning. Compared to these models, the Custom CNN demonstrates better efficiency and comparable performance in certain metrics, but transfer learning clearly provides the most reliable solution for this dataset.
\subsection{Multiclass Classification: PaddyVarietyBD Dataset}

For the Paddy dataset, ResNet50 trained from scratch is the best-performing model, showing the most reliable balance across accuracy, precision, recall, and F1-score. This dataset is fundamentally a fine-grained recognition problem, where different paddy varieties often share very similar global shapes and differ mainly in subtle cues such as grain texture, color tone, edge sharpness, and small pattern variations. A deep model like ResNet50 has enough representational capacity to learn these subtle discriminative features when trained directly on the target domain, and its residual connections support stable optimization even with many layers. As a result, the from-scratch ResNet50 is better able to build a hierarchy of features that are specifically tuned to paddy variety differences, which improves both recall (finding the correct variety) and precision (reducing confusion between similar classes).

\vspace{5pt}
A key observation in this dataset is that transfer learning performs noticeably worse than expected, even though it helped in earlier datasets. The main reason is domain mismatch. ImageNet pretraining is optimized for everyday objects (animals, tools, vehicles, etc.), where the most useful features in deeper layers tend to represent high-level semantic parts (e.g., wheels, faces, object contours). In contrast, paddy variety classification relies heavily on micro-textures and fine surface-level details, which are not strongly represented in the high-level ImageNet feature space. When the backbone is pretrained, the convolutional filters—especially in deeper layers—are biased toward patterns that may not align with the visual structure of paddy grains. This can lead to negative transfer, where the pretrained representations restrict adaptation instead of helping it.

\vspace{5pt}
This issue is amplified by how transfer learning was applied in the setup: only the final classification head was trained for ResNet50 in TL, meaning the convolutional feature extractor remains largely fixed. That strategy works well when the target dataset shares feature similarity with ImageNet (as in vehicles or street scenes), but it becomes a disadvantage in highly domain-specific data like paddy varieties. Because most convolutional layers are frozen, the model cannot sufficiently adjust its mid-level and high-level feature maps to capture paddy-specific discriminative signals. As a result, transfer learning may produce a model that appears to learn quickly but does not improve meaningful class separation, which shows up as weaker F1-score and poorer recall/precision balance compared to the from-scratch version.

\vspace{5pt}
Compared with pretrained models, the Custom CNN shows more consistent and interpretable behavior, but it does not match the best from-scratch deep architecture. The Custom CNN is lightweight and can learn useful low-level patterns, which helps it achieve reasonable recall and stable convergence. However, its limited depth and smaller number of feature channels restrict its ability to form complex hierarchical representations needed for fine-grained variety discrimination. In practice, this means the model can recognize obvious cases but struggles with borderline samples where varieties differ only by subtle texture cues—leading to more confusion between classes and a lower F1-score relative to ResNet50 trained from scratch. Another important point is that the Custom CNN’s performance depends strongly on optimization settings: if the learning rate and dropout are not well balanced, the model may either underfit (missing subtle cues) or overfit (memorizing dataset-specific noise instead of generalizable texture patterns).

\vspace{5pt}
The other large model, VGG16, performs differently because of its architecture. VGG-style networks rely on sequential convolutions without residual shortcuts, and although they can extract strong local textures, they are harder to optimize deeply from scratch and often require careful regularization. In a fine-grained domain, VGG16 may produce good texture features, but without residual connections it can be less robust than ResNet50 when learning subtle class boundaries across many similar categories.

\vspace{5pt}
Overall, the Paddy dataset demonstrates an important conclusion: transfer learning is not automatically beneficial, especially when (1) the domain is texture-heavy and fine-grained, (2) the pretrained feature hierarchy does not align with target cues, and (3) most convolutional layers are frozen so the network cannot adapt. In this setting, training a deep model from scratch becomes the better choice, while the Custom CNN remains valuable as a lightweight alternative when deployment constraints are strict, even though its limited capacity makes it less suitable for maximum-accuracy fine-grained classification.
\subsection{Overall Insight}

Across all datasets, the experiments reveal a clear trade-off between classification performance and computational efficiency. Large pretrained models such as ResNet50 and VGG16 generally achieve higher accuracy, F1-score, and AUC-based metrics due to their deep architectures and the reuse of rich representations learned from ImageNet. These advantages are most evident in datasets with limited training samples or complex scene-level variations, where pretrained features improve generalization and stabilize training. However, this performance gain comes at the cost of significantly higher memory usage, larger model sizes, and increased computational demands, which may limit their practicality in real-world, resource-constrained deployments.

\vspace{5pt}
In contrast, the Custom CNN consistently offers a lightweight and efficient alternative. Although it typically underperforms pretrained models in absolute accuracy and discriminative power, it demonstrates stable training behavior, reasonable recall, and competitive F1-scores across multiple datasets. Its small parameter count and reduced training and inference cost make it particularly suitable for applications such as edge-based monitoring, embedded systems, or large-scale deployments where efficiency and scalability are critical. In several datasets, the performance gap between the Custom CNN and larger models is relatively small, suggesting that carefully designed compact architectures can capture task-relevant features effectively.

\vspace{5pt}
The suitability of each approach depends strongly on the dataset characteristics and application requirements. For object-centric or scene-based datasets with strong similarity to ImageNet, transfer learning provides the most reliable performance. For highly domain-specific or fine-grained datasets, training deep models from scratch can outperform transfer learning when sufficient data is available. In scenarios where computational resources are limited or real-time processing is required, the Custom CNN presents a practical compromise between performance and efficiency.

\vspace{5pt}
Despite its strengths, the Custom CNN has notable limitations. Its limited depth restricts its ability to model subtle inter-class variations, particularly in fine-grained classification tasks. Performance is also more sensitive to hyperparameter selection, which can affect generalization if not carefully tuned. Potential enhancements include incorporating deeper residual blocks, multi-scale feature extraction, or lightweight attention mechanisms to improve representational capacity without significantly increasing model size. Additionally, domain-specific pretraining or self-supervised learning could help bridge the performance gap with large pretrained models while preserving efficiency.

\vspace{5pt}
In summary, the results highlight that no single model is universally optimal. Instead, effective model selection requires balancing performance metrics with computational constraints and application goals, with the Custom CNN serving as a strong candidate when efficiency and deployability are prioritized over maximum accuracy.
\section{Conclusion}\label{sec13}

This work systematically compared a custom lightweight CNN with deeper architectures trained from scratch and via transfer learning across multiple datasets, revealing clear performance–efficiency trade-offs. Transfer learning consistently achieved strong accuracy and stable convergence on object-centric and urban datasets with fewer trainable parameters, while scratch-trained deep models performed well when sufficient domain-specific features were learnable from the data. The Custom CNN demonstrated competitive and stable performance with significantly lower computational and memory costs, making it suitable for resource-constrained applications, although it generally lagged behind deeper pretrained models in peak accuracy. Importantly, the results also showed that transfer learning is not universally optimal, as domain-specific datasets such as PaddyVarietyBD favored scratch-trained models, highlighting the influence of dataset characteristics on model effectiveness. The study emphasizes that optimal model selection depends on balancing accuracy requirements with computational constraints and domain relevance.
\backmatter



\newpage
\bibliography{sn-bibliography}

\end{document}